\pdfoutput=1
\documentclass[lettersize,journal]{IEEEtran}

\usepackage{times}
\usepackage{graphics} 
\usepackage{graphicx}
\usepackage{subfigure}
\usepackage{amsmath,amssymb,amsopn,amstext,amsfonts}
\usepackage{cancel}
\usepackage[space]{cite}
\usepackage{pdfsync}
\usepackage{balance}
\usepackage{color}
\usepackage{mathtools}
\usepackage{bm}

\usepackage{diagbox}
\usepackage{float}
\usepackage{epstopdf}
\usepackage{pifont}
\usepackage{fixltx2e}
\usepackage{amsmath}
\usepackage{multirow}
\usepackage{url}
\usepackage{verbatim}
\usepackage{caption}
\usepackage{adjustbox}
\usepackage{booktabs}
\usepackage{threeparttable}
\usepackage{makecell}
\usepackage{tabularx}
\usepackage{arydshln}
\usepackage[normalem]{ulem}

\usepackage{algorithm}
\usepackage{ifthen}

\usepackage{algorithmic}

\usepackage{array}
\usepackage{textcomp}
\usepackage{stfloats}

\usepackage{siunitx}
\pdfpxdimen=\dimexpr 1 in/72\relax  

\newcommand{\etal}{\emph{et al}.}
\newcommand{\ie}{\emph{i}.\emph{e}.}
\newcommand{\eg}{\emph{e}.\emph{g}.}

\def\MYTITLE{ESVO2: Direct Visual-Inertial Odometry\\ with Stereo Event Cameras}

\definecolor{eccvblue}{rgb}{0.12,0.49,0.85}

\makeatletter
\let\NAT@parse\undefined
\makeatother
\usepackage[pagebackref=false,breaklinks=true,colorlinks,citecolor=eccvblue,bookmarks=true,bookmarksnumbered=true]{hyperref}
\hypersetup{
  pdftitle={\MYTITLE},
  pdfsubject={Robotics, Computer Vision},
  pdfauthor={Junkai Niu, Sheng Zhong, Xiuyuan Lu, Shaojie Shen, Guillermo Gallego, Yi Zhou},
  pdfkeywords={Event Cameras, Stereo, Ego-motion, Asynchronous Sensor, High Temporal Resolution}
}

\usepackage{cleveref}

\title{\MYTITLE}
\author{Junkai Niu$^{\ast}$, Sheng Zhong$^{\ast}$, Xiuyuan Lu, Shaojie Shen, Guillermo Gallego, Yi Zhou$^{\dagger}$
\thanks{Junkai Niu, Sheng Zhong, and Yi Zhou are with the Neuromorphic Automation and Intelligence Lab (NAIL) at School of Robotics, Hunan University, Changsha, China.
Xiuyuan Lu and Shaojie Shen are with the Department of Electronic and Computer Engineering at the Hong Kong University of Science and Technology, Hong Kong, China.
Guillermo Gallego is with TU Berlin, the Science of Intelligence Excellence Cluster, 
the Robotics Institute Germany and the Einstein Center Digital Future, Berlin, Germany.}
\thanks{$\ast$ denotes equal contribution.}
\thanks{Corresponding author ($\dagger$): Yi Zhou. Email: {\small eeyzhou@hnu.edu.cn}.}
}

\def\bc{\mathbf{c}}

\newcommand{\bnum}[1]{\bfseries #1}
\newcommand{\novalue}{{\textendash}}

\definecolor{light-gray}{gray}{0.5}
\newcommand\gframe[1]{{\color{light-gray}\frame{#1}}}

\begin{document}

\setcounter{figure}{-2} %
\makeatletter

\g@addto@macro\@maketitle{
\vspace{4ex}
  \centering
  \includegraphics[width=0.98\linewidth]{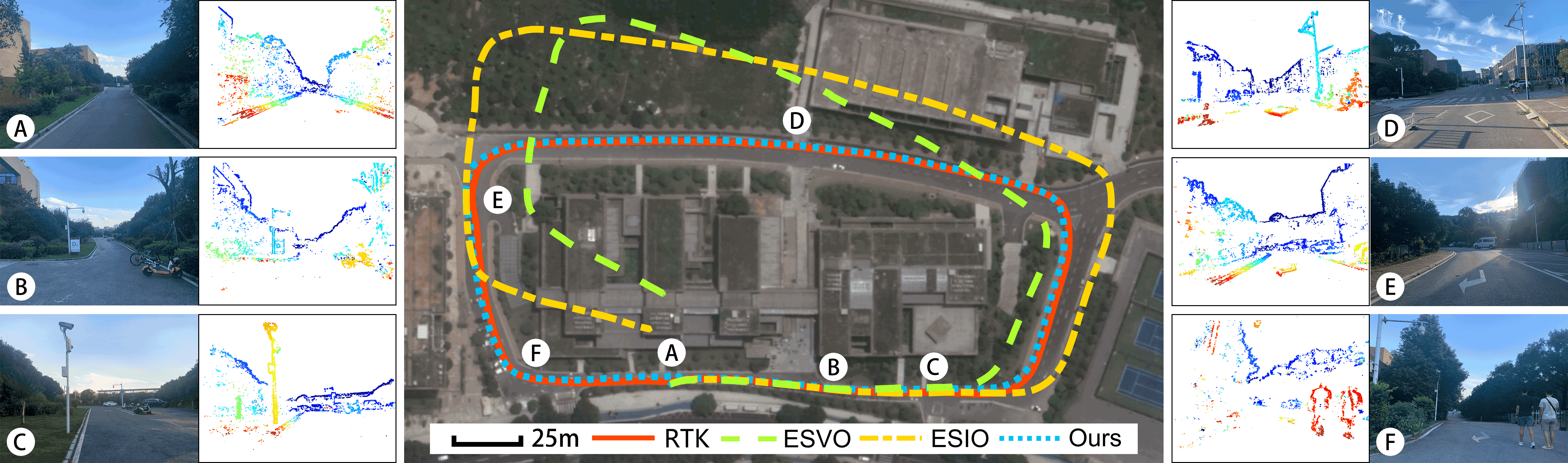}
   \captionof{figure}{
   The proposed event-based stereo visual-inertial odometry system (ESVO2) operates in real time on large-scale VGA-resolution sequences like \emph{hnu\_campus} utilizing a standard CPU.
   Middle: The ground truth of the trajectory is provided by real-time kinematic (RTK) positioning (red) and is aligned with Google Maps. 
   Our method achieves the best performance in terms of global trajectory consistency, while significant drift is witnessed in the results of ESVO~\cite{zhou2021event} (green) and ESIO~\cite{chen2023esvio} (yellow).
   Left and Right: Depth maps reconstructed by our system and corresponding scene images (for visualization purposes only).
   The locations where visual measurements are captured are associated to the map using letters with a circle around them.
   }
\label{fig:eye catcher}

\vspace{-2ex}
}
\makeatother
\maketitle

\begin{abstract}
\label{sec:abstract}
Event-based visual odometry is a specific branch of visual Simultaneous Localization and Mapping (SLAM) techniques, which aims at solving tracking and mapping sub-problems (typically in parallel), by exploiting the special working principles of neuromorphic (\ie, event-based) cameras.
Due to the motion-dependent nature of event data, explicit data association (\ie, feature matching) under large-baseline viewpoint changes is difficult to establish, making direct methods a more rational choice.
However, state-of-the-art direct methods are limited by the high computational complexity of the mapping sub-problem and the degeneracy of camera pose tracking in certain degrees of freedom (DoF) in rotation.
In this paper, we tackle these issues by building an event-based stereo visual-inertial odometry system on top of a direct pipeline~\cite{zhou2021event}.
Specifically, to speed up the mapping operation, we propose an efficient strategy for sampling contour points according to the local dynamics of events. 
The mapping performance is also improved in terms of structure completeness and local smoothness by merging the temporal stereo and static stereo results. 
To circumvent the degeneracy of camera pose tracking in recovering the pitch and yaw components of general 6-DoF motion, we introduce IMU measurements as motion priors via pre-integration. 
To this end, a compact back-end is proposed for continuously updating the IMU bias and predicting the linear velocity, enabling an accurate motion prediction for camera pose tracking.
The resulting system scales well with modern high-resolution event cameras 
and leads to better global positioning accuracy in large-scale outdoor environments.
Extensive evaluations on five publicly available datasets featuring different resolutions and scenarios justify the superior performance of the proposed system against five state-of-the-art methods.
Compared to ESVO~\cite{zhou2021event}, our new pipeline significantly reduces the camera pose tracking error by $40\%$--$80\%$ and $20\%$–-$80\%$ in terms of absolute trajectory error and relative pose error, respectively; at the same time, the mapping efficiency is improved by a factor of five.
We release our pipeline as an open-source software for future research in this field.
\end{abstract}

ov\section*{Multimedia Material}
\noindent Video: {\small \url{https://youtu.be/gmAU32Oeiv8}}\\
Code: {\small\url{https://github.com/NAIL-HNU/ESVO2.git}}

\section{Introduction}
\label{sec: introduction}

Drawing inspiration from biological systems, event cameras represent a revolutionary advancement in sensor technology. 
These sensors transmit pixel-wise intensity changes, called ``events'', asynchronously as they occur \cite{Lichtsteiner08ssc,Gallego20pami,chakravarthi2024recent}.
Consequently, they do not produce grayscale images, nor do they adhere to an external clock of fixed capture rate typical of traditional cameras.
This asynchronous and differential principle of operation minimizes temporal redundancy, thereby reducing power consumption and bandwidth requirements. 
Endowed with high temporal resolution and high dynamic range (HDR) capabilities \cite{Gallego20pami, chakravarthi2024recent}, event cameras are qualified to deal with challenging scenarios that are inaccessible to traditional cameras, such as high-speed motion and/or HDR illumination tracking~\cite{Mueggler14iros,Lagorce15tnnls,Zhu17icra,Gallego17pami,Gallego17ral,Mueggler18tro,Gehrig18eccv,Wang24tro}, control~\cite{Conradt09iscas,Delbruck13fns,Paredes24scirob,forrai2023event, wang2022ev,falanga2020dynamic,sanket2020evdodgenet,bhattacharya2024monocular}, image synthesis~\cite{Pan19cvpr,ZhangX22cvpr,Zhang22pami,Tulyakov22cvpr,rudnev2023eventnerf,bhattacharya2024evdnerf,klenk2023nerf,hwang2023ev,xiong2024event3dgs,deguchi2024e2gs}
and Simultaneous Localization and Mapping (SLAM)~\cite{Rebecq16bmvc,Kim16eccv,Rebecq17ijcv,Rebecq17ral,Rebecq17bmvc,Rosinol18ral,Hidalgo20threedv,Guo24tro,Guo24eccv}.

Although the potential of event cameras is promising, their unconventional output is incompatible with existing modules in visual odometry (VO) / SLAM systems designed for traditional cameras.
Event-based VO, on the one hand, inherits the theoretical formulation and parallel design from modern VO solutions.
On the other hand, the distinctiveness of event-based methods is witnessed in several aspects, including data processing and representation.
Early works using a monocular event camera (\ie,~\cite{Kim16eccv,Rebecq17ral}) require a very gentle motion (typically a small-loopy behavior) for the initialization of a local 3D map, based on which the camera pose can be tracked using a 3D-2D registration pipeline.
The dependence on dedicated hardware (\eg, a GPU) \cite{Kim16eccv} and the inefficient map expansion \cite{Kim16eccv, Rebecq17ral} restrict their applications on mobile platforms equipped with a limited computation resource and energy supply.
More recently, Zhou \etal~\cite{zhou2021event} have proposed an event-based stereo visual odometry (ESVO) pipeline that achieves better accuracy and demonstrates a real-time performance on data collected by a moving agent, such as a flying drone or a ground vehicle.
However, this method does not scale well with the streaming rate of event cameras.
Real-time performance is only witnessed on event cameras with a small spatial resolution (Q-VGA), \eg, a DAVIS346 (Dynamic and Active-Pixel Vision Sensor) \cite{Taverni18tcsii}.
Also, the failure of tracking in the presence of a sudden variation in angular velocity casts the necessity of introducing an inertial measurement unit (IMU) into the system.

The goal of this paper is to lift the above-mentioned limitations of the original ESVO framework.
To this end, we extend ESVO and present a method for visual-inertial odometry (VIO) with a stereo event camera and an IMU.
As demonstrated in Fig.~\ref{fig:eye catcher}, the proposed system achieves better mapping and tracking performance than the event-based state of the art (\eg,~\cite{zhou2021event} and \cite{chen2023esvio}) in terms of accuracy and efficiency due to the following efforts.

\textbf{\emph{Contributions:}}
\begin{enumerate}
    \item A novel image-like representation of events, which utilizes their local dynamics for an adaptive accumulation (AA). 
    It is used for efficiently determining pixel locations associated wth instantaneous edges~(Sec.~\ref{subsec:Adaptive Accumulation of Events}).
    \item An improved solution to the mapping subproblem by leveraging both the temporal-stereo and static-stereo configurations and, additionally, a fast block-matching scheme in the static-stereo operation~(Sec.~\ref{subsec:mapping}). 
    
    \item Incorporating IMU pre-integration as an initial value for event camera pose estimation (Sec.~\ref{subsec:camera pose tracking}), and further constructing a tightly-coupled visual-inertial back-end that keeps suppressing the drift in the bias of the gyroscope and the accelerometer (Sec.~\ref{sec:backend}).
    
    \item An extensive experimental evaluation on five publicly available datasets and comparing against five methods, 
    reporting state-of-the-art performance in accuracy and efficiency (Sec.~\ref{sec:evaluation}).
    It also shows that the system scales well with event cameras of different spatial resolutions and is able to run stably in real time on a standard CPU with data of up to VGA resolution. 
    Besides, our dataset and implementation are open-sourced.
\end{enumerate}

This work has evolved from our ICRA paper~\cite{niu2024imu}, and the technical improvements are clarified at the end of Sec.~\ref{sec:related work}.

\emph{Outline}: The rest of the paper is organized as follows.
First, a literature review of relevant work is provided in Sec.~\ref{sec:related work}.
Second, the proposed framework is then overviewed from a systematic perspective (Sec.~\ref{sec:system overview}).
Then, we discuss our method by detailing each item listed in the contribution (Secs.~\ref{sec:method} and \ref{sec:backend}).
Finally, experimental results are provided in Sec.~\ref{sec:evaluation}, 
followed by the conclusion (Sec.~\ref{sec:conclusion}).
\section{Related Work}
\label{sec:related work}
Like its standard-vision counterparts, event-based VO/SLAM also aims to solve simultaneously the mapping and tracking sub-problems in a recursive manner.
The main challenge therein is to answer the following question: 
\emph{How to process raw events in a way that photometric and geometric constraints can be established to infer the camera pose and scene depth information}?
To this end, several data processing pipelines that utilize different sensor configurations and exploit various data representations have been proposed, as summarized in Tab.~\ref{tab:method comparision}.
From the perspective of how event data are processed, existing methods (also including event-based VIO) can be mainly divided into two categories (2nd column of Tab.~\ref{tab:method comparision}): indirect methods and direct methods.

\textbf{Indirect Methods:}
To build on top of existing indirect methods (\eg, \cite{klein2007parallel, MurArtal15tro}) using standard cameras, researchers have resorted to developing hand-crafted features from event data, such as event corners~\cite{Vasco16iros, Mueggler17bmvc,Alzugaray18ral,Li19iros}, which are typically adapted from the original Harris~\cite{Harris88} and FAST~\cite{Rosten06eccv} methods.
Additionally, strategies for tracking event corners are presented by \cite{Alzugaray18ral, Alzugaray18threedv}.
Such event features enable straightforward application of mature geometric tools, such as epipolar geometry~\cite{hartley1997defense, nister2004efficient} and Perspective-\emph{n}-Point (PnP) methods~\cite{li2012robust, kneip2014upnp}.
Despite the success of these feature-based solutions (\eg,~\cite{hadviger2021feature} and \cite{wang2023event}), 
event features are not as theoretically robust as their standard-vision counterparts.
This is due to the motion-dependent nature of event data, which sometimes leads to incomplete observation of junctions.
Consequently, feature matching can easily fail in a sudden variation of the event camera's motion.
Another mainstream strategy for feature detection and tracking is inspired by the motion compensation method \cite{Gallego18cvpr}, a unified pipeline for event-based model fitting.
This strategy is widely witnessed in event-based VIO pipelines~\cite{Zhu17cvpr,Rebecq17bmvc,chen2023esvio}, which typically build features from motion-compensated event sets~\cite{Zhu17icra} or event histograms~\cite{Gallego19cvpr} and further fuse with inertial measurements via either a Kalman filter~\cite{Mourikis07icra} or keyframe-based nonlinear optimization~\cite{leutenegger2013keyframe}.

\begin{table*}[t]
\centering
\caption{Literature review on representative event-based VO/VIO systems.}
\begin{adjustbox}{max width=\linewidth}
\renewcommand{\arraystretch}{1.15}
\begin{tabular}{llllll}
\toprule
\textbf{Method} & \textbf{(In)Direct} & \textbf{Sensors} & \textbf{Event representation} & \textbf{Real-time performance} & \textbf{Evaluation datasets} \\
\midrule

Kim \etal~\cite{Kim16eccv} & Direct & Monocular & Raw events& 
(128 $\times$ 128 px) on GPU & \novalue \\
Rebecq \etal~\cite{Rebecq17ral} & Direct & Monocular & Raw events \& Naive Accum. & 
(240 $\times$ 180 px) on CPU & ECD \\ 
Zhu \etal~\cite{Zhu17cvpr} & Indirect & Monocular + IMU & Raw events& \ding{55} & ECD \\
Rebecq \etal~\cite{Rebecq17bmvc} & Indirect & Monocular + IMU & Motion-compensated Event Images & 
(240 $\times$ 180 px) on CPU & \novalue \\

ESVO~\cite{zhou2021event} & Direct & Stereo & Time Surfaces & 
(346 $\times$ 260 px) on CPU & RPG, MVSEC, HKUST \\
Hadviger \etal~\cite{hadviger2021feature} & Indirect & Stereo & Time Surfaces & 
(346 $\times$ 260 px) on CPU &  MVSEC, DSEC \\
ESIO~\cite{chen2023esvio} & Indirect & Stereo + IMU & Time Surfaces & 
(346 $\times$ 260 px) on CPU & HKU, MVSEC, VECtor \\
Wang \etal~\cite{wang2023event} & Indirect & Stereo & Time Surfaces \& Naive Accum. & \ding{55} & MVSEC \\
ESVIO~\cite{liu2023esvio} & Direct & Stereo + IMU & Time Surfaces & 
(346 $\times$ 260 px) on CPU & RPG, MVSEC, ESIM, DSEC \\
Elmoudni \etal~\cite{Elmoudni23itsc} & Direct & Stereo & Raw events \& Time Surfaces & \ding{55} & DSEC\\
Shiba \etal~\cite{Shiba24pami} & Direct & Monocular / Stereo & Raw events & \ding{55} & RPG, DSEC\\
ES-PTAM~\cite{Ghosh24eccvw} & Direct & Stereo & Raw events \& Naive Accum. & 
(346 $\times$ 260 px) on CPU & RPG, MVSEC, DSEC, EVIMO2, TUM-VIE\\
Niu \etal~\cite{niu2024imu} & Direct & Stereo + IMU & Time Surfaces \& Adaptive Accum. & 
(346 $\times$ 260 px) on CPU & RPG, DSEC\\
\textbf{This work} & Direct & Stereo + IMU & Time Surfaces \& Adaptive Accum. & 
(640 $\times$ 480 px) on CPU & RPG, MVSEC, DSEC, VECtor, TUM-VIE \\
\bottomrule
\end{tabular}
\label{tab:method comparision}
\end{adjustbox}
\end{table*}

\textbf{Direct Methods:}
Unlike feature-based methods, direct methods refer to those that directly process either events or raw pixel information via some intermediate representation of event data (4th column of Tab.~\ref{tab:method comparision}).
Based on the constant-brightness assumption in logarithmic scale, Kim \etal \cite{Kim16eccv} proposed the first direct method, consisting of three interleaved probabilistic filters to solve the sub-problems of mapping, camera pose tracking and, additionally, recovering the intensity information.
To justify that recovering the intensity information is not needed and that a GPU is not required, Rebecq \etal \cite{Rebecq17ral} proposed a geometric approach.
The mapping module determined the 3D location of scene structures by searching for local maxima of ray densities (represented by a Disparity Space Image -- DSI), 
and the camera pose was estimated by 3D-2D registration, aligning the local 3D map to the 2D event locations on the image plane. 
These two pioneering works are, however, limited by the requirements of gentle motion in the initialization and slow expansion of the local map.

To overcome these limitations, Zhou \etal~\cite{zhou2021event} presented the first event-based VO pipeline (ESVO) using a stereo event camera.
The method exploits spatio-temporal consistency of the events across the image planes of the cameras to solve both localization and mapping sub-problems of VO.
Nevertheless, ESVO does not achieve real-time performance for camera spatial resolutions of $640 \times 480$ pixels (VGA) or higher.
This is mainly due to the large number of redundant operations in the mapping module, which are originally caused by the way contour points are determined.
Besides, we observe that ESVO's tracking module sometimes cannot fully recover the yaw and pitch components in general 6-DoF motion, 
which degrades the accuracy of the recovered trajectory.
To resolve these issues, we recently proposed an IMU-aided version \cite{niu2024imu} of ESVO.
By employing an efficient strategy for sampling contour points, it scales better with modern event cameras of VGA resolution.
The degeneracy issue of camera pose tracking is circumvented by introducing as a prior the gyroscope measurements via pre-integration. 
We also noticed the recent emergence of a direct method using a stereo event camera and an IMU~\cite{liu2023esvio}.
It is built on top of ESVO, with an additional back-end that refines camera poses using IMU pre-integration constraints.
Although it outperforms ESVO in terms of trajectory accuracy, the core limitations of ESVO mentioned above are not well resolved.

Some other follow-up works of ESVO, such as \cite{Elmoudni23itsc}, specifically focus on improving the accuracy of the mapping sub-problem using stereo-fused ray densities \cite{Ghosh22aisy,Ghosh24eccvw}.
The resulting semi-dense depth maps are sharper and more accurate, however, real-time performance on VGA spatial resolution cameras has never been reported.
Finally, Shiba \etal \cite{Shiba24pami} recently extended the Contrast Maximization framework~\cite{Gallego18cvpr,Shiba22sensors} to estimate optical flow and ego-motion from event data.
With special consideration on the space-time properties of event data, it can handle overfitting in the event-alignment problem, thus leading to better performance in geometric-model-fitting tasks, including ego-motion estimation.
Nevertheless, this theoretical work does not demonstrate real-time performance.

The system proposed in this paper is built on top of our recent ICRA work~\cite{niu2024imu} (also in Tab.~\ref{tab:method comparision}) with the following extensions and improvements:
\begin{enumerate}
\item The AA scheme originally proposed in \cite{niu2024imu} is non-trivially improved here. 
Compared to the original design, we introduce the idea of event-dynamic monitoring \cite{nunes2023adaptive} into the termination criteria of event accumulation. 
This modification makes AA responsive to event rate variations without manually adjusting parameters. 
To achieve real-time performance, we also derive an approximate formula to calculate the converged event activity.


\item Inheriting the combination of temporal-stereo and static-stereo configurations from \cite{niu2024imu}, we significantly reduce the computational cost of static-stereo operation in this paper by proposing a fast block-matching scheme, which ensures the mapping module to run stably at 20 Hz on event data of VGA spatial resolution.

\item Using IMU pre-integration as motion priors was naively proposed in \cite{niu2024imu}, however, the IMU bias was not effectively suppressed. 
To solve this problem while ensuring real-time performance, we create a compact and efficient back-end, which considers only linear velocity and IMU biases as optimizing variables. 
Using the back-end's results as the initial values for the spatio-temporal registration leads to more efficient and accurate performance of camera pose tracking.
According to our investigation on various back-end designs (\eg, \cite{liu2023esvio}), our model strikes a good balance between accuracy and efficiency.

\item The proposed pipeline is more extensively evaluated. Compared to \cite{niu2024imu}, ESVO2 reduces the trajectory error by approximately 50\% across all datasets and improves the computational efficiency of the mapping subproblem by more than a factor of two.

\end{enumerate}















\section{System Overview}
\label{sec:system overview}

\begin{figure*}[t]
    \centering 
    {\includegraphics[trim={0 2.2cm 0 0},clip,width=\linewidth]{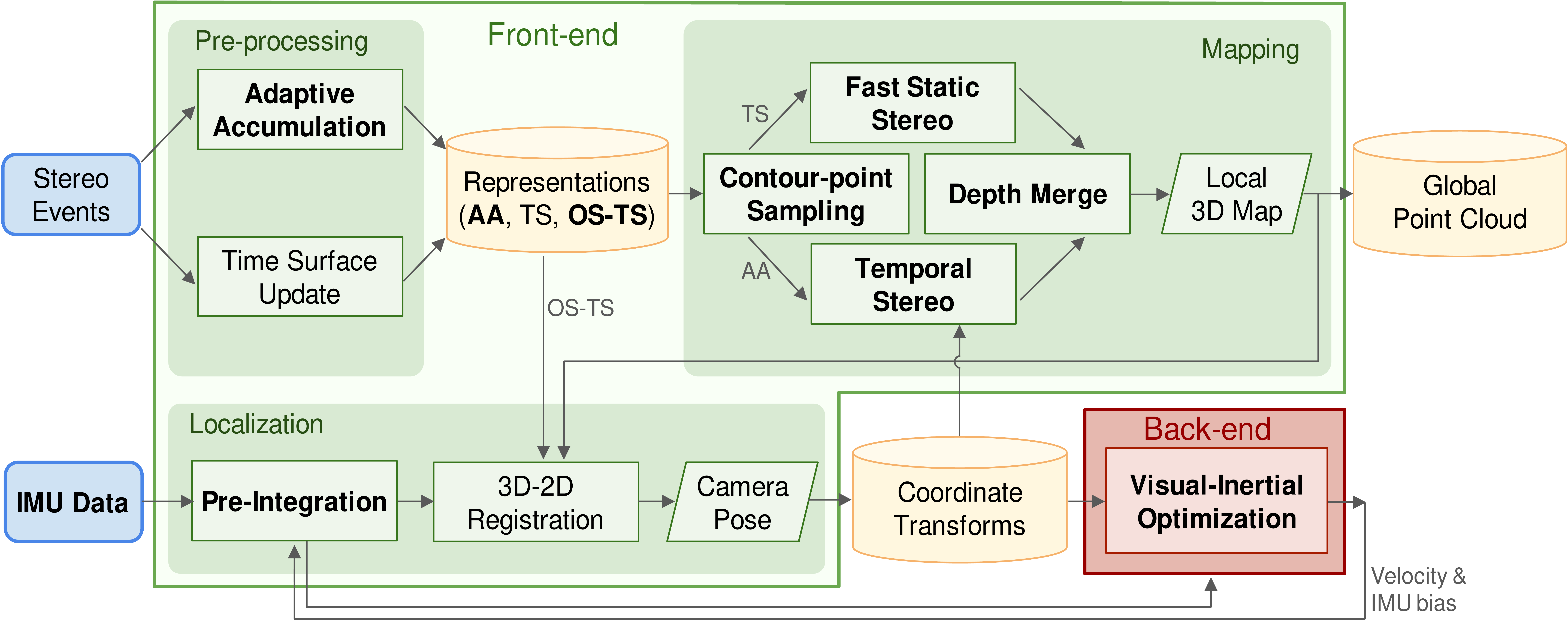}}
    \caption{\emph{Flowchart of the proposed system (ESVO2)}. 
    New functions added to the original ESVO framework are highlighted in bold.
    Each of the four main modules (pre-processing, localization, mapping, and back-end) is executed independently and occupies at least one thread.
    The outputs (rhomboidal shapes) consist of the ``local 3D map'' and the ``camera pose'', 
    which can be stored to produce coordinate transforms and a global point cloud (in yellow).
    \label{fig:system flowchart}
    }
\end{figure*}

Using the original ESVO~\cite{zhou2021event} pipeline as a backbone, we extend it into a stereo VIO system.
Given as input the raw events from a calibrated stereo event camera and the inertial measurements from an IMU, the goal is identical to that of ESVO, namely estimating the stereo event camera's pose in real time and recovering a semi-dense depth map and, furthermore, aiming at a better performance in terms of accuracy and efficiency. 
Figure~\ref{fig:system flowchart} displays the flowchart of the system, with the newly added functions highlighted in bold.
The entire system can be primarily divided into four modules: pre-processing of event data, mapping, camera-pose estimation, and a back-end that keeps updating the IMU bias and the camera's linear velocity. 
The first three modules constitute what is referred to as the front-end.

%

To keep this section self-contained, let us briefly introduce the functions in each module, with a particular emphasis on new components added to the original ESVO pipeline.

First of all, the pre-processing module generates two event representations, including the time surface (TS) map (refer to details in Sec. III. A of \cite{zhou2021event}) and our novel representation called 
the adaptive accumulation (AA) map of events (see Sec.~\ref{subsec:Adaptive Accumulation of Events}).
Both of them can be refreshed in either an asynchronous manner (\eg, at the occurrence of a certain number of events) or at a fixed rate (\eg, 100 Hz in our implementation) and are stored separately in a database for access by other modules.

Secondly, IMU pre-integration is used as the motion prior to initialize the camera pose estimator, which is formulated as a 3D-2D spatio-temporal registration problem 
(alignment of the 3D local map projected onto the time-surface domain).
The resulting pose estimates are stored in a database of coordinate transforms (\eg, TF in ROS \cite{Quigley09icraoss}).

Thirdly, the mapping module takes the contour points sampled on the AA maps, time surfaces, and camera pose estimates to keep updating a semi-dense local depth map.
Specifically, two kinds of stereo methods are deployed to handle points with different observability properties on the spatio-temporal profile of events.
The resulting depth estimates are merged to refresh the local map, which is constantly pushed into a database of global point cloud for potential visualization.
Finally, the back-end realizes a sliding-window optimization that keeps updating the estimates of IMU biases and camera velocity at a fixed rate.


The bootstrap of our system creates an initial local map and initializes the unknowns in the back-end (\ie, the IMU biases and camera velocity).
The former operation is identical to that in ESVO.
A loosely-coupled method is performed in the latter operation, which determines the back-end parameters by aligning two trajectories obtained independently from events and IMU measurements.
Different from what has been proposed in the initialization of \cite{liu2023esvio}, we find that the refinement of scale is unnecessary as long as the extrinsic parameters are sufficiently accurate.

\section{Front-End}
\label{sec:method}

We detail our front-end in this section.
First, we introduce a novel image-like representation of event data that facilitates efficient sampling of contour points (Sec.~\ref{subsec:Adaptive Accumulation of Events}).
Second, we discuss our modifications to ESVO's mapping that lead to improvements in terms of structure completeness and local smoothness (Sec.~\ref{subsec:mapping}).
Then, we demonstrate how to further enhance the camera pose estimation by incorporating the IMU pre-integration as motion priors
and employing offset-free time surfaces in the spatio-temporal registration (Sec.~\ref{subsec:camera pose tracking}).

The front-end interacts with a compact and efficient back-end (Sec.~\ref{sec:backend}) that helps suppress drift in the estimated IMU biases, which is key to globally consistent trajectory estimates.



\subsection{Pre-processing: Adaptive Accumulation of Events}
\label{subsec:Adaptive Accumulation of Events}


The computational efficiency of the mapping method in ESVO is limited by several aspects, and one of them is the way that the edge-pixel locations are determined.
In ESVO, an edge map in a virtual reference frame is created by applying motion compensation to events occurring within a short time interval (\eg, 10 ms).
This operation will become computationally expensive as a pre-processing step when the streaming rate of events exceeds a certain range\footnote{The event streaming rate is in proportion to several aspects, including the scene dynamics, scene texture, and spatial resolution of event cameras.}.
Besides, we observe that the extracted events are typically concentrated in certain regions with large optical flow.
To alleviate such an uneven distribution of contour points, the method has to sample a large number of points, which is redundant and becomes a computational burden to mapping.
Therefore, a more efficient way is needed to determine contour points.
The synthesized event map obtained by a naive accumulation of events in \cite{Rebecq17ral} can be used as an approximate edge map.
However, this approximation can become considerably inaccurate (\ie, blurred edges or invisible edges) when the depth varies significantly in the scene.
This is because a global threshold for event accumulation cannot handle diverse local dynamics of events.
To address this, we propose AA, a novel representation of event data,
that can control the amount of events to be accumulated according to the local event dynamics.
In \cite{nunes2023adaptive}, the event dynamic is represented by event activity $a(t, s)$, which is defined as the number of events within a given time interval $(s, t]$. 
A temporal decay strategy is applied to the event activity, which is updated every time an event arrives. 
The event activity at the arrival of the $i_{\text{th}}$ event can be represented as
\begin{figure}[t]
	\centering
	\subfigure[Convergence behavior of event activity.]{\includegraphics[width=0.23\textwidth]{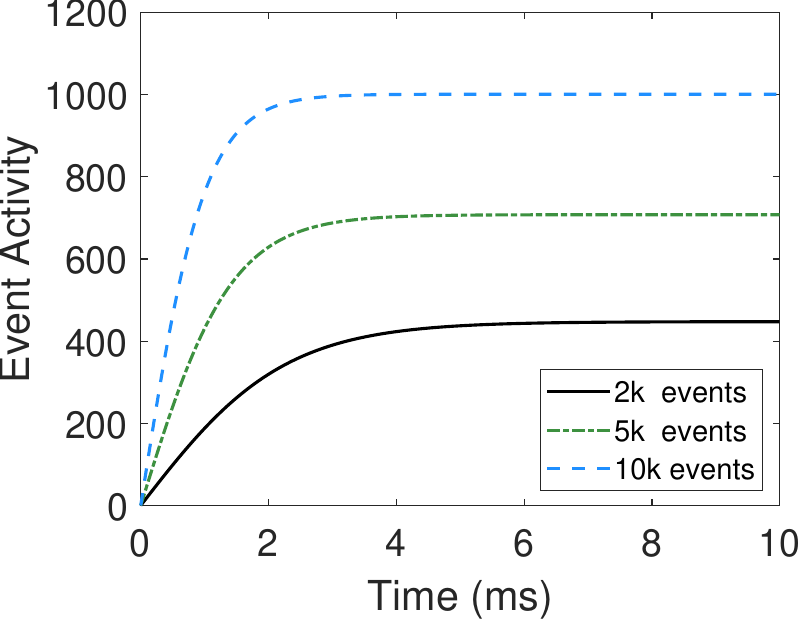}
    \label{fig:convergence_AA_a}}\hspace{1pt}
    \subfigure[Irrelevance to temporal distribution of events.]{\includegraphics[width=0.23\textwidth]{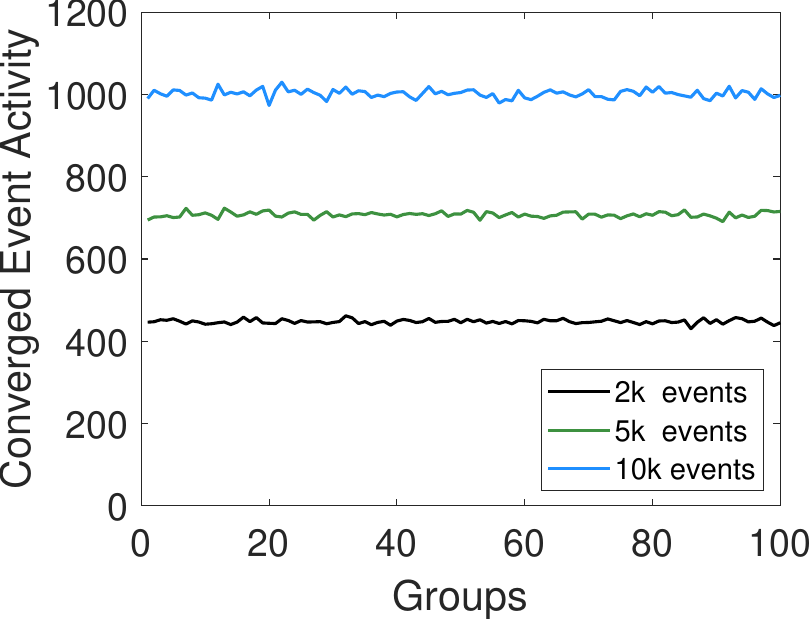}
    \label{fig:convergence_AA_b}}
    \caption{
    \emph{Convergence of event activities.}
    Panel (a) illustrates the converging process of event activities with three different event dynamics (2k, 5k, and 10k events per 10 ms, respectively).
    Panel (b) reports that the value of event activities at convergence is roughly independent of whether events are triggered uniformly in the temporal domain, as justified by 100 groups of independent simulations.
    }
    \label{fig:convergence_AA}
    \vspace{-0.5em}
\end{figure}
\begin{equation}
\label{eq:event activity update}
  a_{i} = \beta_{i} a_{i - 1} + 1, 
\end{equation}
where $\beta_{i}$ is the decay rate that varies with event activity $a_{i - 1}$ and is determined by the time interval $(t_{i} - t_{i-1})$ between two consecutive events:
\begin{equation}
\label{eq:beta update}
 \beta_{i} = \frac{1}{1 + a_{i - 1}(t_{i} - t_{i-1})}.  
\end{equation}
The method in \cite{nunes2023adaptive} utilizes the event activity within a time interval to represent the global event dynamics, controlling the decay rate depending on the event activity to generate adaptive-decay time surfaces.
Inspired by \cite{nunes2023adaptive}, our method monitors the event activity to determine the termination time of event accumulation.
Intuitively, a greater event activity corresponds to higher event dynamics, indicating a shorter duration for event accumulation.
We also observe that the event dynamics remain relatively stable within a short time interval (\eg, 10 ms), which leads to event activities converging towards a constant value incrementally, as shown in Fig.~\ref{fig:convergence_AA_a}. 
Assuming the events are generated at equal time intervals, 
the converged event activity can be roughly calculated as
\begin{equation}
\label{eq:convergence}
 \lim_{t_i \to t}a_i = (1 + \sqrt{1 + 4 / b}) / 2 \approx 1 / \sqrt{b},~t_i \in (s, t],
\end{equation}
where $b$ denotes the length of the time interval in seconds.
As seen in Fig.~\ref{fig:convergence_AA_b}, this estimation of converged event activity also holds approximately when the time interval $b$ is not constant. 
Therefore, we terminate event accumulation when the event activity converges to obtain a sufficiently sharp image.
To handle the spatial variation of local event dynamics, we divide the image plane into small regions and accumulate events independently according to the local event activity. 

The generation of an AA map is summarized in Algorithm~\ref{alg:AA algorithm}.
Specifically, we first divide the image plane evenly into $N$ small regions 
and the accumulation of events is carried out in each region independently.
Events are sequentially accumulated in a region of an event count image until the event activity converges to the steady state value estimated by \eqref{eq:convergence}.
The pixel value in an AA map represents the number of events accumulated at this pixel.
Pixels with higher values are more likely to be associated with contour points.

\setlength{\textfloatsep}{0pt}
\begin{algorithm}[t]
\caption{Adaptive Accumulation of Events
}
\label{alg:AA algorithm}
\renewcommand{\algorithmicrequire}{\textbf{Input:}}
\renewcommand{\algorithmicensure}{\textbf{Output:}}
\begin{algorithmic}[1]
\REQUIRE
All involved events $\{e_k \doteq (x_k, y_k, t_k, p_k)\}_{k=1}^{N_e}$, and the amount of regions $N$.
\ENSURE
Adaptive accumulation map $\mathbf{A}(x, y)$.
\STATE Initialize $\mathbf{A}(x, y)$ with zero values.
\STATE Divide $\mathbf{A}(x, y)$ into $N$ regions and assign each one a boolean flag $\mathcal{F}_{j} $ = \text{False} and two variables ${a}_{j}$ and ${a}_{j}^{\text{conv}}$ for storing the value of event activities.


\FOR{$j=1,\ldots,N$}
\STATE Estimate the average time interval $b^{j}$  within the $j_{\text{th}}$ region and calculate ${a}_{j}^{\text{conv}}$ using Eq.~\eqref{eq:convergence}.
\FOR{$i=1,\ldots,N_e^{j}$}
\IF{ $\mathcal{F}_{j}$ == \text{True}  }
\STATE Break.
\ENDIF
\STATE Update $a_{j}$ using Eq.~\eqref{eq:event activity update} and Eq.~\eqref{eq:beta update}. 
\STATE $\mathbf{A}(x_i, y_i)$++.
\IF{$|a_{j} - {a}_{j}^{\text{conv}}| / {a}_{j}^{\text{conv}}< 0.02$}
\STATE $\mathcal{F}_{j}$ = True.
\ENDIF
\ENDFOR
\ENDFOR
\end{algorithmic}
\end{algorithm}

\begin{figure*}[t]
	\centering
    \subfigure[Edge Map (from image)]{\gframe{\includegraphics[width=0.19\textwidth]{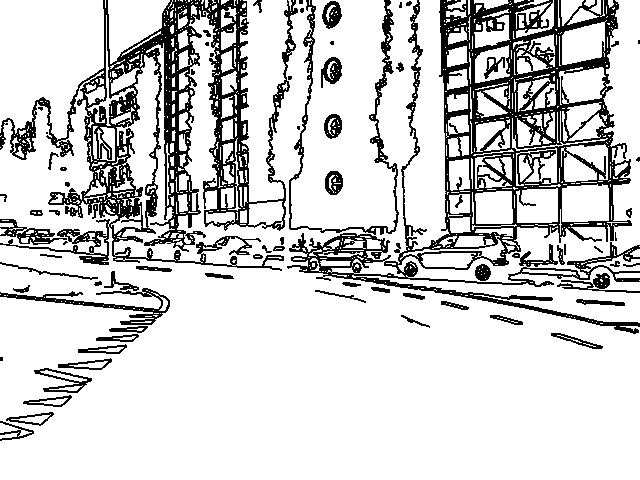}}}
    \subfigure[\textbf{T}ime \textbf{S}urface~\cite{Lagorce17pami}]{\gframe{\includegraphics[width=0.19\textwidth]{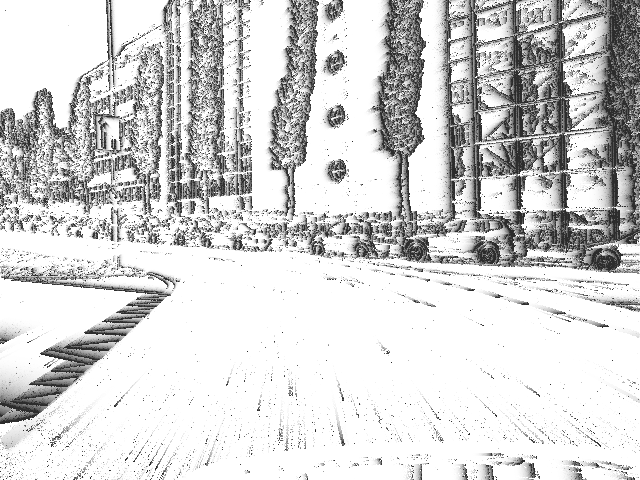}}}
     \subfigure[SILC~\cite{Manderscheid19cvpr}]{\gframe{\includegraphics[width=0.19\textwidth]{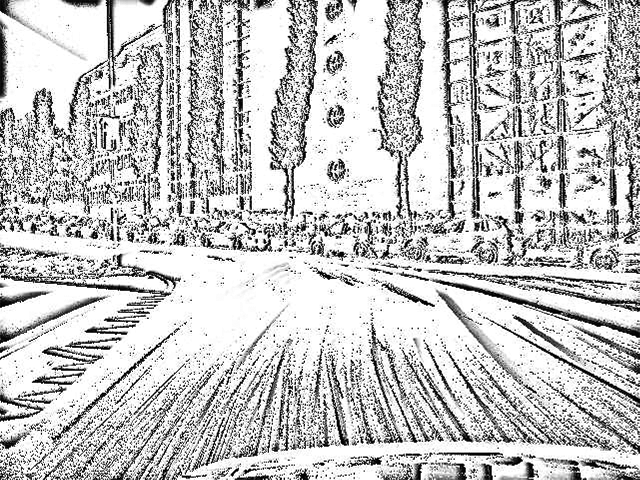}}}
     \subfigure[TOS~\cite{glover2021luvharris}]{\gframe{\includegraphics[width=0.19\textwidth]{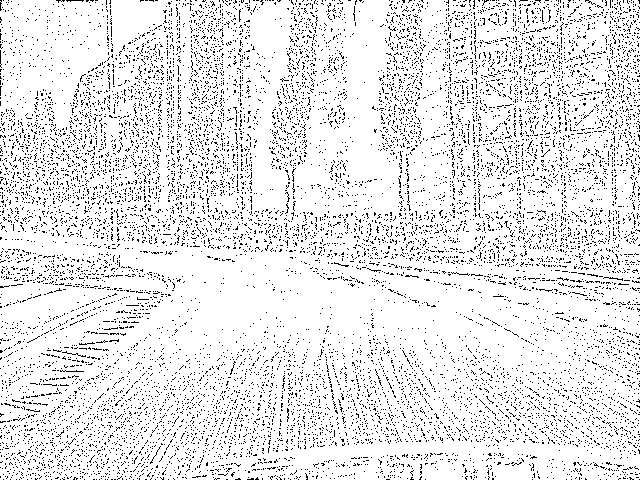}}}
	\subfigure[\textbf{A}daptive \textbf{A}ccumulation]{\gframe{\includegraphics[width=0.19\textwidth]{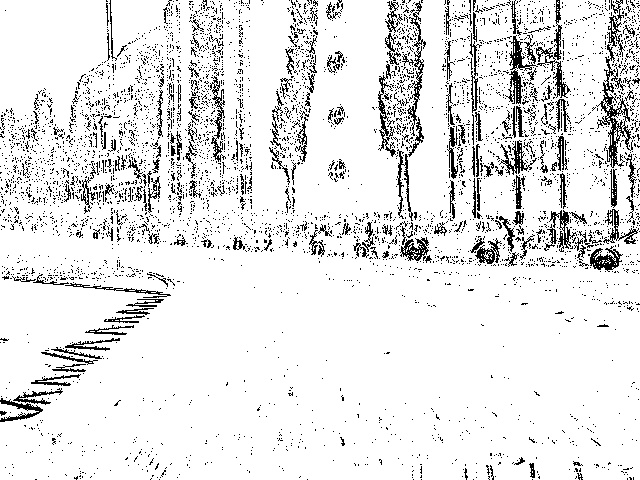}}}
	\caption{\label{fig:image-like representation comparision}
 ``Event maps'' generated using different event data representations, including TS~\cite{Lagorce17pami}, SILC~\cite{Manderscheid19cvpr}, TOS~\cite{glover2021luvharris}, and the proposed AA map.
 The data used are from the \emph{dsec\_city\_04a} sequence, which is recorded using a Prophesee Gen3.1 (VGA) event camera.
 The leftmost edge map is extracted from an intensity image and used for visualization only.}
\end{figure*}

\begin{figure}[t]
	\centering
    \subfigure[ESVO (TEPR: 54.6\%)]{{\includegraphics[width=0.48\linewidth]{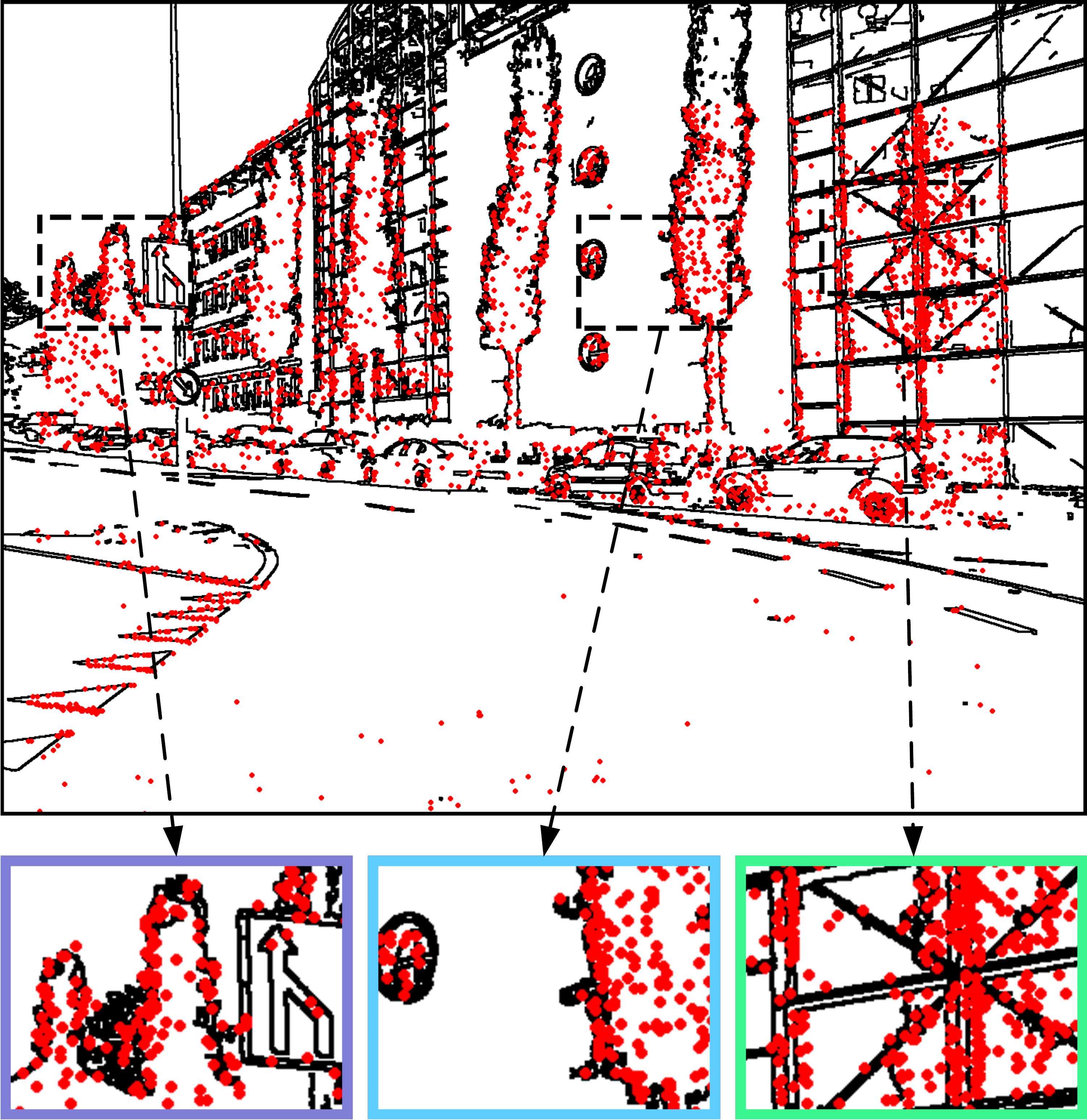}}}\hspace{1pt}
	\subfigure[Ours (TEPR: 73.7\%)]{{\includegraphics[width=0.48\linewidth]{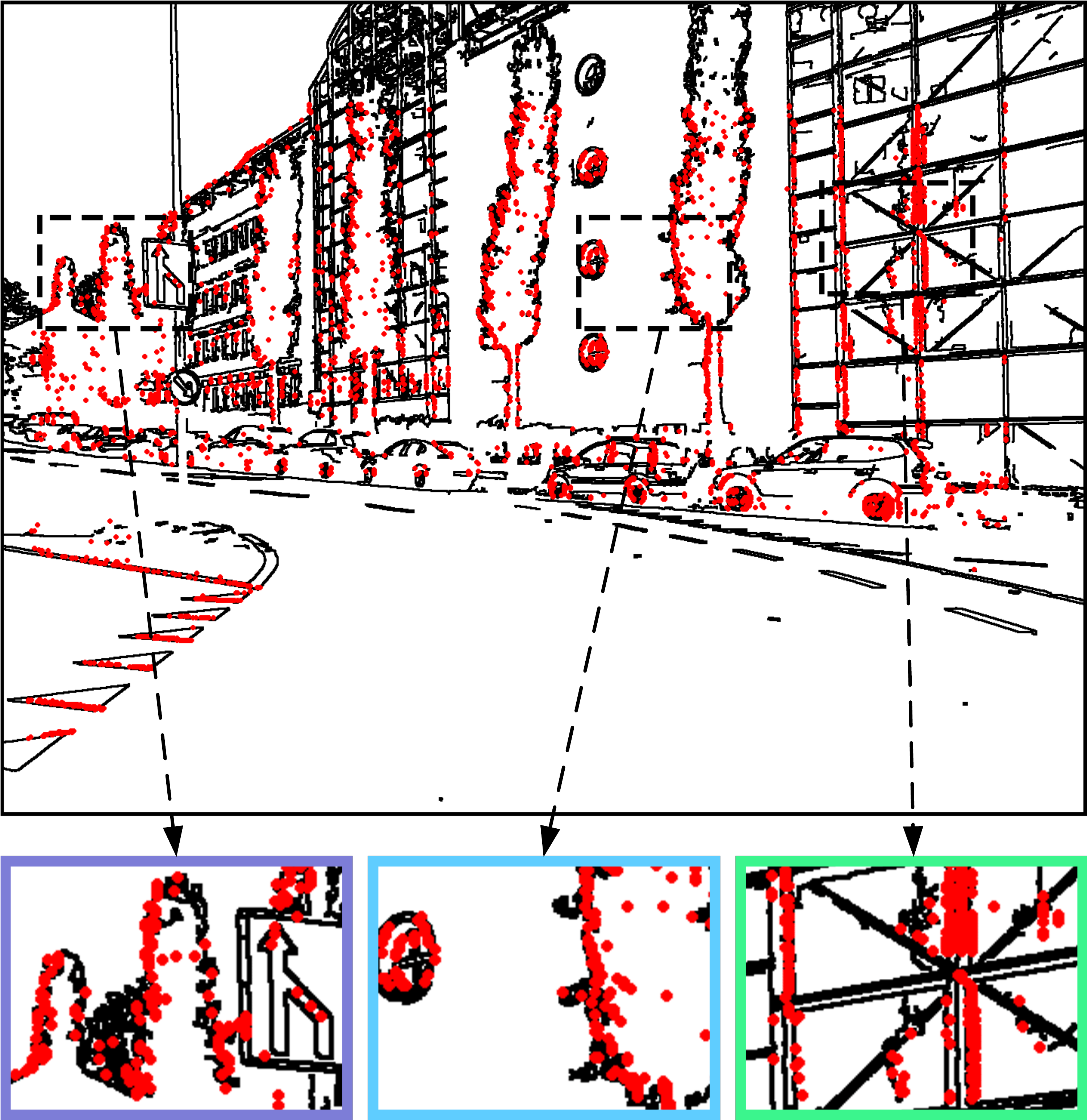}}}
 \caption{\label{fig:sampling result comparison}
 \emph{Point sampling.}
 Panels (a) and (b) compare the results of sampling 4000 points by ESVO and our method, respectively.
 The background is an edge map obtained by processing the grayscale image with an edge detection algorithm.
 The values in parenthesis denote the proportion of contour points (black) covered by the sampled points (red).
 }
\vspace{1.5em}
\end{figure}

To assess the effectiveness of the proposed AA method, we compare it against three image-like representations, including time surfaces~\cite{Lagorce17pami} and two speed-invariant ones, namely \emph{Speed Invariant
Learned Corners} (SILC)~\cite{Manderscheid19cvpr} and \emph{Threshold-Ordinal Surface} (TOS)~\cite{glover2021luvharris}.
As illustrated in Fig.~\ref{fig:image-like representation comparision}, the AA map preserves relatively complete edges with fewest redundant points while keeping the highest true-edge-point ratio (TEPR) \footnote{By ``TEPR'', we mean the proportion of contour points covered by the sampled points, as illustrated in Fig.~\ref{fig:sampling result comparison}.}.
This superior performance can also be justified quantitatively by comparing each representation against the edge map obtained from the corresponding grayscale image.
As shown in Tab.~\ref{tab:similarity evaluation of TS, TOS, SILC and AA}, the proposed AA representation leads to the highest score using the multi-component structure similarity index measure (3-SSIM)~\cite{wang2003multiscale} as evaluation metric. 

\begin{table}[t]
\vspace{0.4em}
\begin{center}
\caption{
The multi-component SSIM (3-SSIM) of various event maps compared to the edge map obtained from the corresponding intensity image.
The data used for the 3-SSIM computation are identical to those used in Fig.~\ref{fig:sampling result comparison}.}
\begin{tabular}{lcccc}
\toprule
{Event map} & {TS~\cite{Delbruck08issle}} & {SILC~\cite{Manderscheid19cvpr}} & {TOS~\cite{glover2021luvharris}}  & {AA}\\
\midrule
{3-SSIM} & {0.1801} & {0.1369} & {0.15203} & {\textbf{0.3387}}\\
\bottomrule
\end{tabular}
\label{tab:similarity evaluation of TS, TOS, SILC and AA}
\end{center}
\end{table}

Contour points 
are further sampled from each region independently. 
In general, the higher the AA pixel value in each region, the greater the probability that the pixel will be selected.
We compare the sampling result to the contour points obtained by ESVO.
As shown in Fig.~\ref{fig:sampling result comparison}, our sampled pixels better capture the edge structures, indicating that our sampling method can cover the edge patterns in the environment using fewer redundant input points.

\subsection{Mapping: Depth Estimation}
\label{subsec:mapping}

The mapping module aims at recovering depth information and maintaining a local 3D map.
In the original ESVO pipeline, these are achieved by constantly propagating recent depth estimates of sampled events to a virtual reference frame, on which overlapped estimates are further fused based on their probabilistic characteristics.
As for estimating the depth of each individual event, a two-step method (\ie, a block matching operation followed by a nonlinear refinement) is proposed.
More specifically, a triangulation operation is performed given the resulting disparity from the block matching, and the nonlinear refinement operates in a forward-projection manner (\ie, from the inverse depth domain to 2D event measurements). 
In general, the applied static stereo configuration in ESVO leads to unreliable depth estimation of structures parallel to the stereo baseline.
Moreover, the two-step method for depth estimation does not scale well with the number of individual queries.
To circumvent these two problems, we propose two modifications: 1) an additional temporal-stereo matching method and 2) a faster design of block matching.
We detail each modification in the following.

\subsubsection{Temporal Stereo Matching}
\label{subsubsec:temporal stereo matching}
The static stereo method in \cite{zhou2021event} can hardly recover accurate depth of structures that are parallel to the baseline of the stereo camera.
This is because the spatio-temporal profile of these structures is not distinctive along the baseline, and thus, multiple false-positive matches will be obtained during the block matching operation.
Inspired by \cite{Engel15iros}, we introduce the temporal stereo method (\ie, establishing stereo associations between adjacent views of the same camera) to resolve this problem.
It is known that as long as the stereo camera does not move in the direction of the baseline, epipolar lines defined between a temporal-stereo pair are no longer parallel to the static-stereo baseline.
This assumption always holds for forward moving stereo cameras, \eg,~those employed in driving scenarios.



Let us consider the commonly used horizontal stereo configuration.
We divide the sampled contour points from Sec.~\ref{subsec:Adaptive Accumulation of Events} into two groups according to their gradient direction on the TS.
In the first group, we collect pixels at which the ratio ($\eta$) between the vertical gradient and the horizontal gradient is smaller than a threshold.
These pixels are fed to the static stereo method.
The remaining sampled pixels go to the second group and are fed to the proposed temporal stereo method.

The key to the event-based temporal-stereo problem is to effectively exploit appearance similarity in event data.
This requires the applied event representation, on which the stereo data association is established, to possess the speed-invariant property to some extent.
Since the TS used in ESVO~\cite{zhou2021event} does not satisfy this property, 
it is not a good choice for temporal stereo matching.
To address this, we investigate the temporal stereo matching performance on three representations, namely TOS~\cite{glover2021luvharris}, SILC~\cite{Manderscheid19cvpr}, and our AA.
We find that AA is the optimal choice due to its highest contour points ratio.
Consequently, the epipolar matching is carried out on successive AA maps of the left camera, followed by triangulation and optionally by nonlinear refinement.

Note that the complementary property of the temporal-stereo configuration to the static one would hold as long as the temporal-stereo epipolar line is not parallel to the static-stereo baseline. 
This assumption is typically valid for most mobile robot platforms (\eg, ground vehicles and drones) when the stereo camera is set looking forward.


\subsubsection{Fast Block Matching}
\label{subsubsec:faster block matching}
The static stereo method in ESVO~\cite{zhou2021event} is initialized by a block-matching operation, which uses the zero-mean normalized cross-correlation (ZNCC) as a similarity measure: 
\begin{equation}
    Z_{x,y,d} = \frac{\mathrm{cov}_{x,y,d}(l, r)}{\sqrt{\mathrm{var}_{x,y}(l) \; \mathrm{var}_{x,y+d}(r)}},
    \label{eq:zncc_ori}
\end{equation}
where $\mathrm{cov}_{x,y,d}(l, r)$ denotes the intensity covariance between the patch $p_l$ centered at $(x,y)$ on image $l$ and the patch $p_{r,d}$ centered at $(x,y+d)$ on image $r$; $\mathrm{var}_{x,y}(l)$ and $\mathrm{var}_{x,y+d}(r)$ refer to the variance of $p_l$ and $p_{r,d}$, respectively.

\begin{table}[t]
\begin{center}
\caption{\emph{Computational complexity of different implementations for ZNCC}. 
The test is conducted on four time surfaces selected from sequences of different resolutions, and 300 blocks are selected on each time surface.
Block size is set to 15$\times$15 px, with a search distance of 100 px.}
\setlength{\tabcolsep}{4mm}{
\begin{tabular}{lcc}
\toprule
{Algorithm} & {Traditional ZNCC} & {Fast ZNCC}\\
\midrule
Mean time           & 30.70~$\mu$\text{s} & 5.761~$\mu$\text{s} \\
Median time         & 30.16~$\mu$\text{s} & 5.714~$\mu$\text{s} \\ 
25th percentile     & 29.97~$\mu$\text{s} & 5.571~$\mu$\text{s} \\
75th percentile     & 30.45~$\mu$\text{s} & 5.819~$\mu$\text{s} \\
Standard deviation  & 2.369~$\mu$\text{s} & 0.452~$\mu$\text{s} \\
\bottomrule
\end{tabular}
}
\label{tab:zncc_cost}
\end{center}
\end{table}

Although the original ZNCC leads to satisfactory accuracy, it is notably time consuming. 
To speed up, we propose a fast version of ZNCC based on the approach presented in~\cite{LIN201764}.
It adopts a recursive method that calculates the required sums and squared sums for the computation of ZNCC:
\begin{align}
&    \mathrm{cov}_{x,y,d}(l, r) = \frac{\langle {p_l}, {p_{r,d}} \rangle - \mu_l T_{r,d}}{mn},
    \label{eq:fast_zncc_cov}\\
&
    \mathrm{var}_{x,y,d}(r) = \frac{mnT_{r^2,d} - T_{r,d} T_{r,d}}{(mn)^3},
    \label{eq:fast_zncc_var}\\
    &T_{r,d} =
    T_{r,d-1} - \sum_{i=1}^{m}p_{r,d-1}(i,1) + \sum_{i=1}^{m}p_{r,d}(i,n),
    \label{eq:fast_zncc_tr}\\
    &T_{r^2,d} =
    T_{r^2,d-1} - \sum_{i=1}^{m}(p_{r,d-1}(i,1))^2 + \sum_{i=1}^{m}(p_{r,d}(i,n))^2,
    \label{eq:fast_zncc_tr2}
\end{align}
where $m$ and $n$ denote the number of rows and columns of $p_l$ and $p_{r,d}$, $\mu_l$ the mean intensity of $p_l$, $T_{r,d}$ the sum of the intensities in $p_{r,d}$, and $T_{r^2,d}$ the sum of the squared intensities in $p_{r,d}$. 
In this way, the approach effectively utilizes intermediate results from previous computations to reduce redundant computation in ZNCC.
As shown in Table.~\ref{tab:zncc_cost}, this recursive method significantly reduces the computational complexity of block matching,
leading to a speedup of approximately an order of magnitude.

\subsubsection{Depth Merge}
Given as input the relative pose between successive AA maps of the left camera, the proposed temporal-stereo method returns the depth information of structures lacking in the static-stereo method.
Ideally, these stereo results can be fused in the same way as done in \cite{zhou2021event}, because the temporal-stereo residuals $r_{\text{temporal}}$ evaluated on AAs and the static-stereo residuals $r_{\text{static}}$ evaluated on TSs approximately obey the \emph{Student's t} distribution
\begin{equation}
\label{eq:t distribution}
    r \sim St(\mu_{r}, s_{r}, \nu_{r}),
\end{equation}
where $\mu_{r}$, $s_{r}$, $\nu_{r}$ are the model parameters, namely the mean, scale, and degree of freedom.
However, this is not sensible because the uncertainty of the temporal-stereo results is always much smaller, leading to a bias towards the temporal-stereo estimates.
Hence, we straightforwardly merge the results of the two stereo methods to obtain a more complete depth map, as shown in Sec.~\ref{subsec:comparison of mapping}.
Specifically, the employed merging operation consists of two steps.
First, the depth estimates from the static-stereo and temporal-stereo methods are fused independently into two separate depth maps according to their respective probabilistic characteristics, in the same way as described in \cite{zhou2021event}. 
Afterwards, we fill the non-occupied pixels in the static-stereo depth map with the results in the temporal-stereo depth map.
Also note that we get rid of the nonlinear refinement of depth in the static-stereo and temporal-stereo methods in our implementation.
We find that the absence of the nonlinear refinement guarantees the real-time performance of mapping, while not bringing any notable degradation in the accuracy of camera pose tracking.
The effect of omitting the nonlinear refinement is discussed in Sec.~\ref{subsubsec:Ablation study refinement}.

\begin{figure}[t]
	\centering
    \vspace{-1em}
    \subfigure[Original TS]{\includegraphics[width=0.15\textwidth]{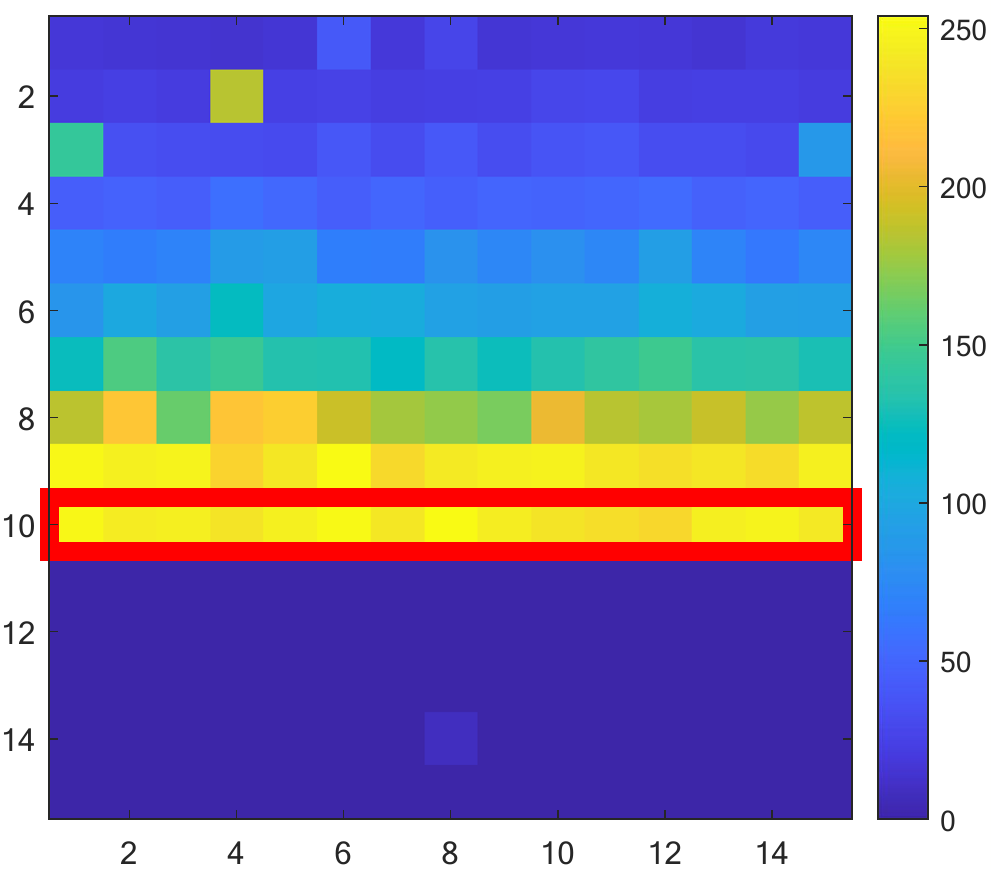}\label{fig:TS_discussion_TS_ori}}
    \subfigure[Blurred TS]{\includegraphics[width=0.15\textwidth]{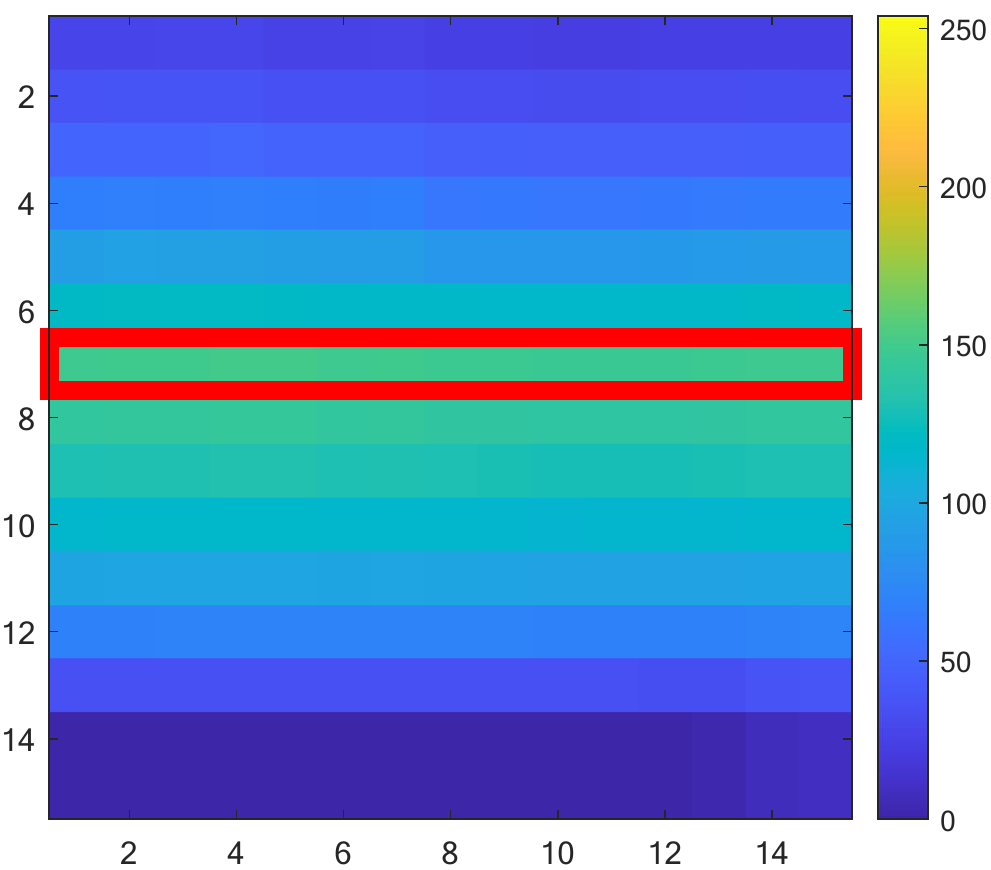}\label{fig:TS_discussion_TS_blur}}
    \subfigure[Offset-free TS]{\includegraphics[width=0.15\textwidth]{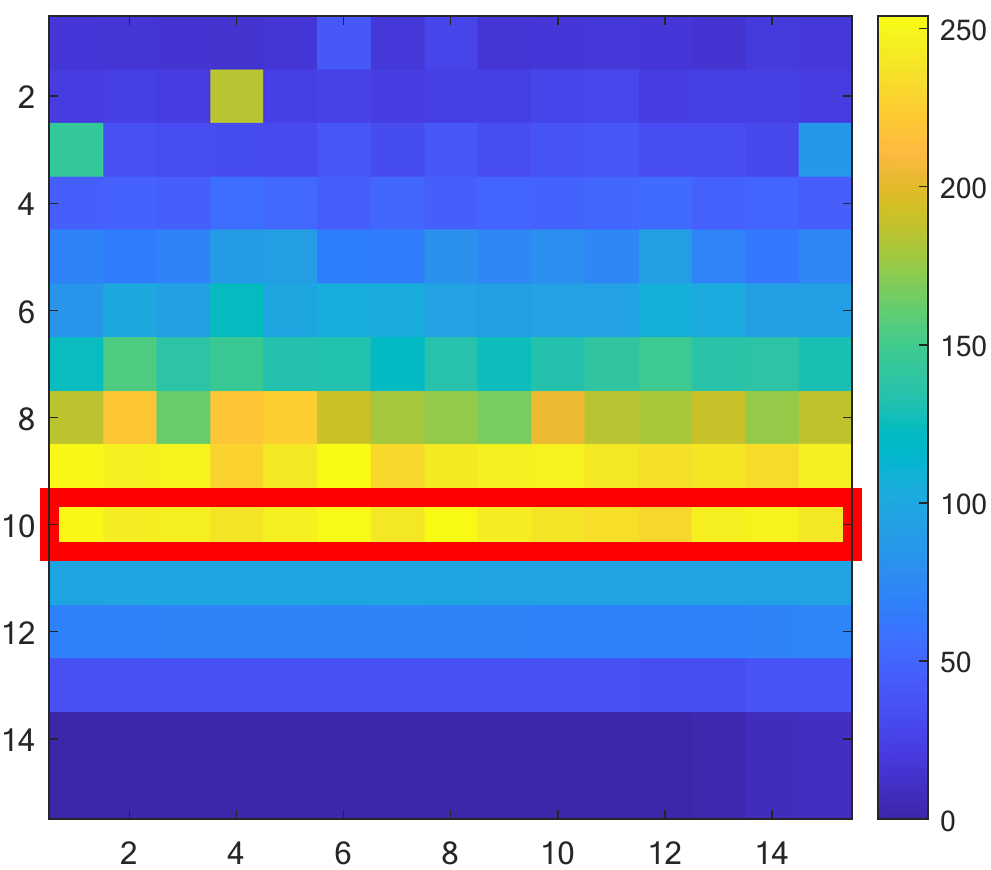}\label{fig:TS_discussion_TS_composite}}
 \caption{
 \emph{Comparison of the original TS, Gaussian-blurred TS, and offset-free smoothed TS (OS-TS) for the same patch.}
 The color from blue to yellow represents an increase in pixel value.
 The pixel row with the highest values is highlighted using a red rectangle.
 Real data from \emph{dsec\_city\_04a} are used in rendering these three representations.
 }
    \label{fig:TS_discussion}
\end{figure}

\subsection{Localization: Camera Pose Estimation}
\label{subsec:camera pose tracking}
Our front-end follows the tracking module in ESVO~\cite{zhou2021event}, which takes as input a TS and a local 3D map, and computes the pose of the stereo rig with respect to the map.
Let $\mathcal{S}^{\mathcal{F}_{\mathbf{ref}}} = \{ \mathbf{x}_{i}\}$ represent a set of pixels with inverse depth values $\{\rho_i\}$ in the reference frame, and $\mathcal{T}(\mathbf{x}, t)$ and $\overline{\mathcal{T}}(\mathbf{x}, t) = 1 - \mathcal{T}(\mathbf{x}, t)$ be the TS and negative TS at time $t$.
The objective is, as in \cite{zhou2021event}, to determine the optimal motion parameters $\boldsymbol{\theta}$ by solving
\begin{equation}
\label{eq: optimization}
\boldsymbol{\theta}^{\ast} = \mathop{\arg\min}\limits_{\boldsymbol{\theta}} \sum\limits_{\mathbf{x}_i \in \mathcal{S}^{\mathcal{F}_{\mathbf{ref}}}} \overline{\mathcal{T}}_\text{left}(\boldsymbol{W}(\mathbf{x}_i, \rho_i; \boldsymbol{\theta})),
\end{equation}
where the warp $\boldsymbol{W}(\mathbf{x}_i, \rho_i; \boldsymbol{\theta}) = \pi_\text{left}(T (\pi^{-1}_\text{ref}(\mathbf{x}_i,\rho_i), G(\boldsymbol{\theta})))$
transfers points from the reference frame $\mathcal{F}_{\mathbf{ref}}$ to the latest TS using the inverse depth $\rho_i$ and projection operator $\pi$ (see \cite{zhou2021event}).
The transformation matrix $G(\boldsymbol{\theta}): \mathbb{R}^6 \mapsto SE(3)$ corresponds to the motion parameters $\boldsymbol{\theta} \doteq (\mathbf{c}^{\top}, \mathbf{t}^{\top})^{\top}$,
where $\bc = (c_{1},c_{2},c_{3})^{\top}$ are the Cayley parameters~\cite{cayleyparameter} for rotation and $\mathbf{t} = (t_{x}, t_{y},t_{z})^{\top}$ are translation parameters.

The spatio-temporal registration module leverages the distance-transform property of time surfaces and formulates the camera pose estimation as a 3D-2D registration problem.
The employment of time surfaces smoothed with a blur kernel, on the one hand, enhances the differentiability of the objective function.
On the other hand, the true location of edge points is shifted on the image plane, leading to a bias in the registration result.
The nonlinear property of the resulting 3D-2D registration problem must be initialized properly, especially in the presence of a sudden variation in the orientation.
Also, it may come across a degeneration in recovering the pitch and yaw components, and thus, an additional sensor (e.g., an IMU) is needed as a supplement.
We discuss our solutions to these problems in the following.

\begin{algorithm}[t]
\caption{Offset-free Smoothed Time Surface (OS-TS)}
\label{alg:discussion TS}
\renewcommand{\algorithmicrequire}{\textbf{Input:}}
\renewcommand{\algorithmicensure}{\textbf{Output:}}
\begin{algorithmic}[1]
\REQUIRE
Original time surface $\text{TS}_\text{ori}$.\\
\ENSURE
Offset-free smoothed time surface $ \text{TS}_\text{offset\_free} $.
\STATE $\text{TS}_\text{blur}$ $\gets$ Apply Gaussian blur to $\text{TS}_\text{ori}$.
\FOR{each pixel $(i,j)$ in $\text{TS}_\text{ori}$}
\IF{$\text{TS}_\text{ori}(i,j) == 0$}
\STATE $\text{TS}_\text{offset\_free}(i,j) = \text{TS}_\text{blur}(i,j)$.
\ELSE
\STATE $\text{TS}_\text{offset\_free}(i,j) = \text{TS}_\text{ori}(i,j)$.
\ENDIF
\ENDFOR
\end{algorithmic}
\end{algorithm}
\setlength{\textfloatsep}{2ex}

\subsubsection{Offset-free Smoothed Time Surface (OS-TS)}
\label{subsubsec:Offset-free Smoothed Time Surface}
The original TS provides a truncated distance field with slopes ascending towards the true location of contours / edges.
The lack of gradients in the truncated area, as shown in Fig.~\ref{fig:TS_discussion_TS_ori}, leads to a one-sided convergence basin.
To this end, a blurring filter is applied to the TS in ESVO~\cite{zhou2021event} to obtain a two-sided convergence basin, as illustrated in Fig.~\ref{fig:TS_discussion_TS_blur}.
The smoothing operation, however, shifts the basin, leading to a notable offset in the location of the edges.
This phenomenon can be seen by comparing the region with the highest pixel values in Fig.~\ref{fig:TS_discussion_TS_ori} and Fig.~\ref{fig:TS_discussion_TS_blur}.
Consequently, the spatio-temporal registration with blurred TS is no longer optimal because the 3D map is aligned with the shifted observation of edges.

To address this issue, we propose an alternative design, called offset-free smoothed time surface (OS-TS), as outlined in Alg.~\ref{alg:discussion TS}.
Specifically, it fills the side lacking valid gradient information in the original TS (Fig.~\ref{fig:TS_discussion_TS_ori}) using the pixel values in the corresponding area of the smoothed TS (Fig.~\ref{fig:TS_discussion_TS_blur}).
As illustrated in Fig.~\ref{fig:TS_discussion_TS_composite}, the result exhibits no offset while preserving valid gradient information on both sides of the edge.
Finally, the negative OS-TS is computed and used in the spatio-temporal registration.

\subsubsection{Using IMU Pre-integration as Motion Priors}
\label{subsubsec:Using IMU Pre-integration as Motion Priors}
To deal with sudden variations in orientation and provide a proper initialized value for the spatio-temporal registration, we use as motion priors the pre-integration of an IMU~\cite{Lupton12tro}.
For an IMU with a 3-axis accelerometer and gyroscope, the measurements of acceleration $\tilde{\boldsymbol{a}}^{b}$ and angular velocity $\tilde{\boldsymbol{\omega}}^{b}$ can be denoted as
\begin{equation}
\label{eq:gyro measurement}
\begin{split}
  \tilde{\boldsymbol{a}}^{b} &= \boldsymbol{a}^{b} + \mathbf{b}_{a}^{b} +  \mathbf{n}_{a}^{b}, \\
  \tilde{\boldsymbol{\omega}}^{b} &= \boldsymbol{\omega}^{b} + \mathbf{b}_{g}^{b} +  \mathbf{n}_{g}^{b},  
\end{split}
\end{equation}
where $\mathbf{b}_{a}^{b}$, $\mathbf{b}_{g}^{b}$, $\mathbf{n}_{a}^{b}$ and $\mathbf{n}_{g}^{b}$ are the bias and noise of the accelerometer and gyroscope in the IMU frame, respectively.
The IMU measurement, \eg,~from time $t_i$ to $t_{i+1}$, expressed in the IMU frame $b_{i}$, can be calculated by
\begin{equation}
\label{eq:integration of angular velocity}
\begin{split}
    \boldsymbol{\hat \alpha}_{b_{i+1}}^{b_{i}} &= \iint_{t \in [t_{i}, t_{i+1}]} {\mathbf R}^{b_{i}}_{b_{t}}(\boldsymbol{a}^{b_{t}})  \emph{dt}^{2},\\ 
    \boldsymbol{\hat \beta}_{b_{i+1}}^{b_{i}} &=  \int_{t \in [t_{i}, t_{i+1}]} {\mathbf R}^{b_{i}}_{b_{t}}(\boldsymbol{a}^{b_{t}} )  \emph{dt}, \\
    \boldsymbol{\hat \gamma}_{b_{i+1}}^{b_{i}} &= \int_{t \in [t_{i}, t_{i+1}]} \frac{1}{2} \boldsymbol{\Omega}(\boldsymbol{\omega}^{b_{t}}) \boldsymbol{\hat \gamma}_{b_{t}}^{b_{i}} \emph{dt},
\end{split}
\end{equation}
where $\boldsymbol{\Omega}(\boldsymbol{\omega}) = \begin{bmatrix}  - {\lfloor \boldsymbol{\omega} \rfloor}_{\times} & \boldsymbol{\omega} \\  -\boldsymbol{\omega}^{\top} & 0 \end{bmatrix}$, and ${\lfloor \cdot \rfloor}_{\times}$ is the cross-product matrix operator of a vector.
The linear velocity and biases (\ie, $\mathbf{v}_{b_{i}}$, $\mathbf{b}_{a}^{b}$ and $\mathbf{b}_{g}^{b}$) are initialized empirically and constantly updated in the back-end (see Sec.~\ref{sec:backend}).
We estimate the position prior from $t_i$ to $t_{i+1}$ by $(t_{i+1} -t_{i})\mathbf{v}_{b_{i}}$, while neglecting the impact of acceleration due to the high operating rate of our tracking module, and employ $\boldsymbol{\hat \gamma}_{b_{i+1}}^{b_{i}}$ as rotation prior.
The prior position and orientation are used to initialize the search for the optimal pose in \eqref{eq: optimization}.
\section{Back-end Optimization}
\label{sec:backend}


To suppress the drift in the motion estimates, a bundle adjustment (BA) optimization is typically needed as a back-end.
One of the most widely used strategies is the sliding-window approach, which jointly refines the poses of all involved key frames and, optionally, the depth information of co-visible 3D points by minimizing the inconsistency among heterogeneous measurements.
However, our geometric method lacks explicit data association in the spatio-temporal domain, and thus, such a relaxation-and-refinement scheme~\cite{Forster14icra} is not applicable.
Moreover, we observe that the presence of camera poses among the optimization variables does not improve the camera tracking performance.
This is because the local map (\ie, the involved point cloud within the window) is a fusion result relying on existing camera poses.
It is by no means to establish additional constraints by introducing the 3D point cloud into the sliding-window optimization without relaxation. 
Therefore, unlike \cite{Qin18tro,liu2023esvio}, we attempt to create a compact and efficient back-end, which considers only the linear velocity and IMU's bias as the optimizing variables.
Experiments show that the back-end built on this strategy significantly improves the tracking accuracy of our system  (Sec~\ref{subsec: Ablation Studies}).

Specifically, we apply a sliding window (shown in Fig.~\ref{fig:sliding window}), in which the last $N$ pose estimates\footnote{Note that the involved poses are only used for providing the relative pose constraints rather than being updated.} and the corresponding IMU's measurements are involved.
The objective is to refine the estimated linear velocity and IMU biases at the corresponding time of each involved pose by minimizing the Mahalanobis norm of IMU measurement residuals:
\begin{figure}[t]
    \centering
    \includegraphics[width=\linewidth]{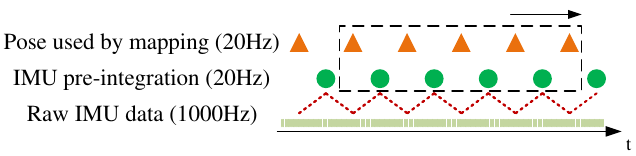}
    \caption{\emph{Illustration of data within the sliding window.}
    The data includes 5 poses used by mapping (orange triangles) and 4 results of the inter-frame IMU pre-integration (green circles).}
    \label{fig:sliding window}
\end{figure}
\begin{equation}
\label{eq:back-end optimization problem}
    \boldsymbol{\chi}^{*} = \mathop{\arg\min}\limits_{\boldsymbol{\chi}} \sum\limits_{{i} \in \mathcal{I}} \| r_{\mathcal{I}} 
 (\hat{\boldsymbol z}_{b_{i + 1}}^{b_{i}}, \boldsymbol{\chi}, \boldsymbol{\zeta}) \|_{\mathbf{P}^{b_i}_{b_{i+1}}}^{2},
\end{equation}
where $\boldsymbol{\chi}^{*}$ denotes the optimizing variables (\ie, the linear velocities and the biases of IMU), $\boldsymbol{\zeta}$ the involved pose estimates, and $\hat{\boldsymbol z}$ the IMU pre-integrated measurements.
In addition, $\mathcal{I}$ denotes the set of IMU pre-integration results in the sliding window. 
Furthermore, $\mathbf{P}^{b_{i}}_{b_{i+1}}$  is the covariance matrix of the IMU pre-integration during the propagation process, and it is used as the weight for the IMU measurements.
Detailed derivation can be found in Sec. IV.B of \cite{Qin18tro}.
The relative pose residual $r_{\mathcal{I}} (\hat{\boldsymbol z}_{b_{i + 1}}^{b_{i}}, \boldsymbol{\chi}, \boldsymbol{\zeta})$ is specifically defined as:
$$
r_{\mathcal{I}} 
 (\hat{\boldsymbol z}_{b_{i + 1}}^{b_{i}}, \boldsymbol{\chi}, \boldsymbol{\zeta}) 
=
\begin{bmatrix}
\delta \boldsymbol{\alpha}^{b_i\top}_{b_{i+1}},
\delta \boldsymbol{\beta}^{b_i\top}_{b_{i+1}},
\delta \boldsymbol{\gamma}^{b_i\top}_{b_{i+1}}, 
\delta \mathbf{b}_a^\top,
\delta \mathbf{b}_g^\top
\end{bmatrix}^\top
$$
\begin{equation}
=
\begin{bmatrix}
\mathbf{R}_{w}^{b_i}\left(\mathbf{p}_{b_{i+1}}^w - \mathbf{p}_{b_i}^w + \frac{1}{2} \mathbf{g}^w \Delta t_i^2 - \mathbf{v}_{b_i}^w \Delta t_i\right) - \hat{\boldsymbol{\alpha}}^{b_i}_{b_{i+1}} \\
\mathbf{R}_{w}^{b_i}\left(\mathbf{v}_{b_{i+1}}^w + \mathbf{g}^w \Delta t_i - \mathbf{v}_{b_i}^w\right) - \hat{\boldsymbol{\beta}}^{b_i}_{b_{i+1}} \\
2\left[{(\mathbf{q}^{w}_{b_i})}^{-1} \otimes \mathbf{q}_{b_{i+1}}^{w} \otimes \left(\hat{\boldsymbol{\gamma}}^{b_i}_{b_{i+1}}\right)^{-1}\right]_{xyz} \\
\mathbf{b}_{a}^{b_{i+1}} - \mathbf{b}_{a}^{b_i} \\
\mathbf{b}_{g}^{b_{i+1}} - \mathbf{b}_{g}^{b_{i}}
\end{bmatrix},
\end{equation}
where $\mathbf{g}^{w}$ indicates the gravity in the world coordinate system, $\mathbf{p}$ and $\mathbf{q}$ denote the position and rotation, respectively. ${[ \cdot ]}_{xyz}$ extracts the vector portion of the quaternion $\mathbf{q}$ for representing the error state.
In detail, the size of the sliding window is set to 5, and we apply the Levenberg-Marquardt (LM) algorithm to solve the optimization problem.
The sliding window shifts forward when a new pose estimate is obtained, while the oldest pose estimate and corresponding IMU measurements are removed. 

\label{subsec:system}

\section{Experiments}
\label{sec:evaluation}

In this section, we comprehensively evaluate our system.
First, we introduce all datasets used and explain why some of them are not applicable for the evaluation (Sec.~\ref{subsec:experimental setup and datasets used}).
Second, we conduct both qualitative and quantitative comparisons of the proposed mapping module against that of ESVO, justifying the benefits brought by our modifications (Sec.~\ref{subsec:comparison of mapping}).
Third, we evaluate the overall performance of our system by comparing the localization results against those of various stereo event-based pipelines and demonstrate our superior performance in terms of trajectory accuracy, especially in driving scenarios (Sec.~\ref{subsec:system evaluation}).
Besides, an extended evaluation using our dataset is provided to justify the versatility of our system (Sec.~\ref{subsec: Outdoor Evaluation}).
Moreover, we investigate the benefits brought by the newly incorporated functions and modules on the tracking performance (Sec.~\ref{subsec: Ablation Studies}) and provide a computational-complexity analysis of each function in the entire system (Sec.~\ref{subsec:implementation details and computational efficiency}).
Finally, we discuss some limitations of our method in small-scale and narrow indoor scenes (Sec.~\ref{subsec:limitations}).

\subsection{Public Datasets Used}
\label{subsec:experimental setup and datasets used}

Our system is tested on five publicly available datasets with stereo event cameras and IMUs, which feature different event-camera types, spatial resolutions, and scene geometries, including \emph{DSEC}~\cite{Gehrig21ral}, \emph{VECtor}~\cite{gao2022vector}, \emph{MVSEC}~\cite{Zhu18ral}, \emph{rpg}~\cite{Zhou18eccv} and \emph{TUM-VIE}~\cite{klenk2021tumvie}.
\emph{DSEC} is a large-scale dataset capturing outdoor driving scenes, and the others feature small indoor scenes with different sensor resolutions. 
Sequences in \emph{MVSEC} are recorded using a stereo event camera mounted on drones, whereas a hand-held rig is used in \emph{VECtor}, \emph{rpg}, and \emph{TUM-VIE}.
The characteristics of each dataset are summarized in Tab.~\ref{tab:camera parameters}.

\begin{table}[t]
\centering
\caption{\label{tab:camera parameters}Event camera parameters of the datasets used in the experiments.}
\begin{adjustbox}{max width=\linewidth}
\renewcommand{\arraystretch}{1.2}
\setlength{\tabcolsep}{1.2mm}{
\begin{tabular}{llccc}
\toprule
\textbf{Dataset} & \textbf{Cameras} & \textbf{Resolution} [px] & \textbf{BL} [cm] & \textbf{FOV} [$^\circ$] \\
\midrule

\emph{rpg}~\cite{Zhou18eccv} & DAVIS240C & 240$\times$180 & 14.7 & 62.9\\
\emph{MVSEC}\protect~\cite{Zhu18ral} & DAVIS346 & 346$\times$260 & 10.0 & 74.8\\
\emph{DSEC}\protect~\cite{Gehrig21ral} & Prophesee Gen3.1 & 640$\times$480 & 59.9 & 60.1\\
\emph{VECtor}\protect~\cite{gao2022vector} & Prophesee Gen3 & 640$\times$480 & 17.0 & 67.0\\
\emph{TUM-VIE}\protect~\cite{klenk2021tumvie} & Prophesee Gen4 & 1280$\times$720~ & 11.8 & 65.0\\
\emph{Ours} (Sec.~\ref{subsec: Outdoor Evaluation}) & DVXplorer & 640$\times$480 & 51.0 & 72.4\\
\bottomrule
\end{tabular}
}
\end{adjustbox}
\end{table}


Note that some of the other released datasets using stereo event cameras are not applicable to our evaluation.
For example, an unsatisfactory rectification is witnessed in the \emph{HKU} dataset~\cite{chen2023esvio} due to inaccuracies in the extrinsic calibration.
The \emph{ECMD} dataset \cite{chen2023ecmd} exhibits a significant discrepancy in the streaming rate of events between the left and right event cameras, thus breaking the assumption of temporal coherence between the two cameras.
Last but not the least, the ``Kirschbaum problem'' \cite{ponitz122010efficient} is witnessed in the \emph{M3ED} dataset \cite{chaney2023m3ed}, which employs event cameras with a spatial resolution of 1280 $\times$ 720 pixels and collects data in urban environments filled with high-frequency texture patterns (\eg, trees and lawns).
According to \cite{gehrig2022high}, the higher the spatial resolution, the more sensitive each pixel becomes to brightness changes. 
Therefore, such a sensor-scene configuration results in an excessive number of events recorded in this dataset, leading to a lack of clear structure information in the spatio-temporal profile of events.
This may result in failures of the proposed direct method.
\def\figWidth{0.31\linewidth} 
\begin{figure}[t]
\centering
\vspace{-0.2em}
\setlength\tabcolsep{0pt}
{\small
\begin{tabular}{cccc}
    {} & Intensity image\  & ESVO\  & Our method\   \\ 
    \rotatebox{90}{\makecell[c]{\ \emph{dsec\_city04\_a}}}&
    \begin{minipage}[b]{\figWidth}
		\centering
		{\includegraphics[width=\linewidth]{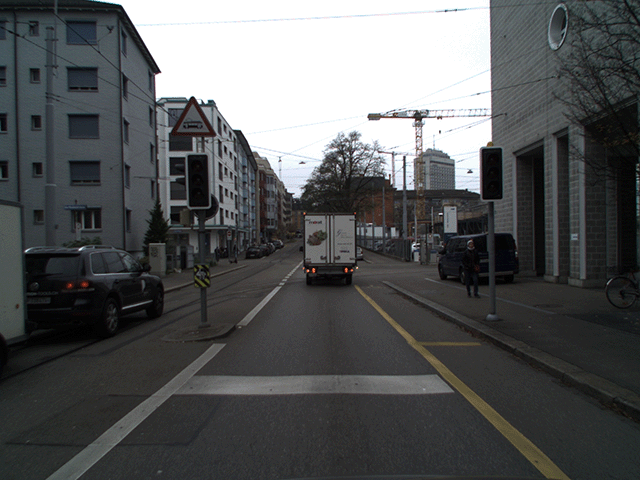}}
    \end{minipage} \hspace{4pt}&     
	\begin{minipage}[b]{\figWidth}
		\centering
		{\includegraphics[width=\linewidth]{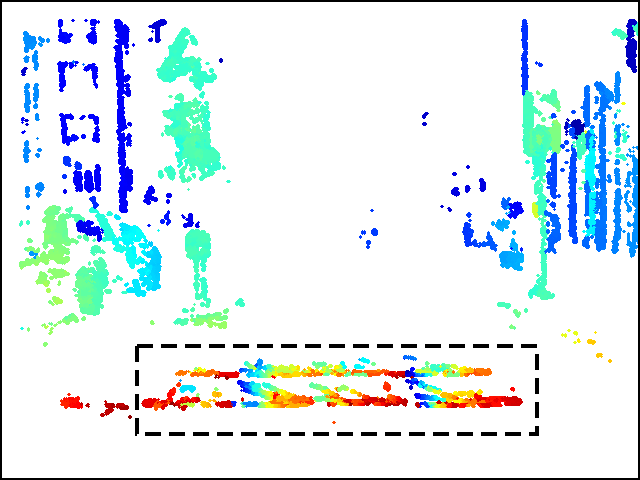}}
	\end{minipage} \hspace{4pt}&      
	\begin{minipage}[b]{\figWidth}
		\centering
		{\includegraphics[width=\linewidth]{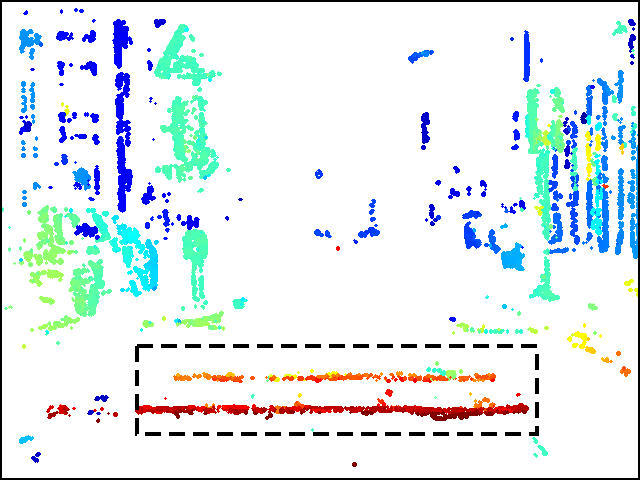}}
	\end{minipage}  \vspace{0pt}\\
 
    \rotatebox{90}{\makecell[c]{\ \emph{dsec\_city04\_c}}}&
    \begin{minipage}[b]{\figWidth}
		\centering
		{\includegraphics[width=\linewidth]{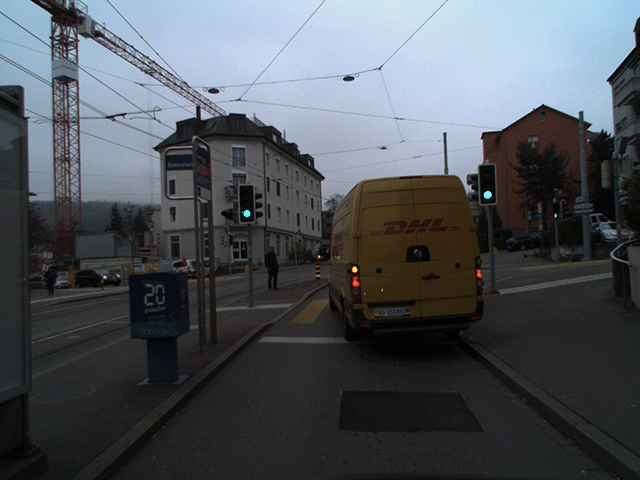}}
    \end{minipage} \hspace{4pt}&     
	\begin{minipage}[b]{\figWidth}
		\centering
		{\includegraphics[width=\linewidth]{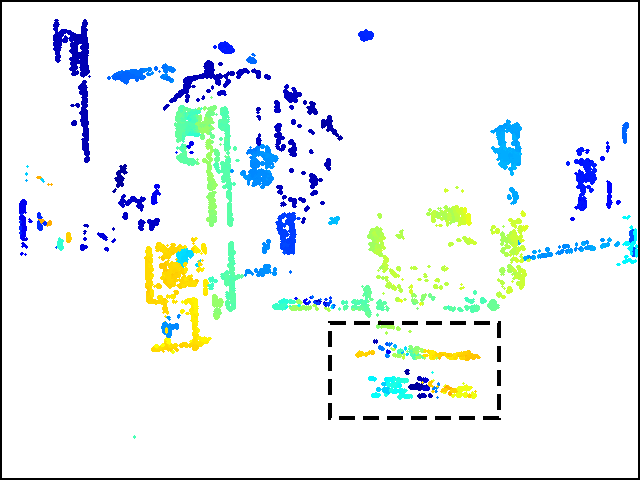}}
	\end{minipage} \hspace{4pt}&      
	\begin{minipage}[b]{\figWidth}
		\centering
		{\includegraphics[width=\linewidth]{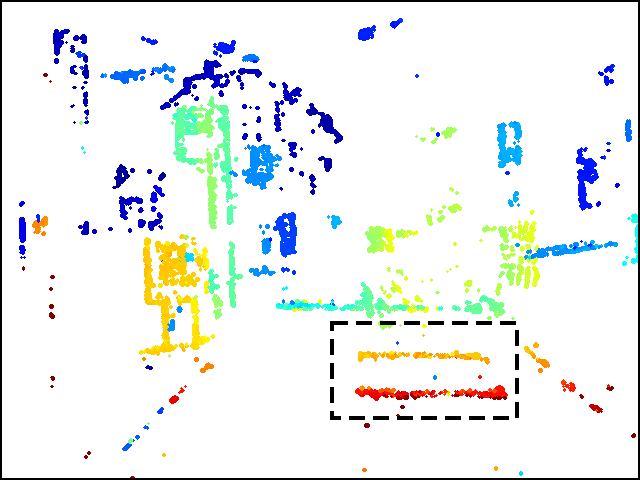}}
	\end{minipage}  \vspace{0pt}\\
 
 
 
    \rotatebox{90}{\makecell[c]{\ \emph{rpg\_monitor2}}}&
    \begin{minipage}[b]{\figWidth}
		\centering
		{\includegraphics[width=\linewidth]{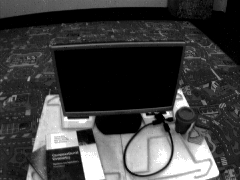}}
    \end{minipage} \hspace{4pt}&     
	\begin{minipage}[b]{\figWidth}
		\centering
		{\includegraphics[width=\linewidth]{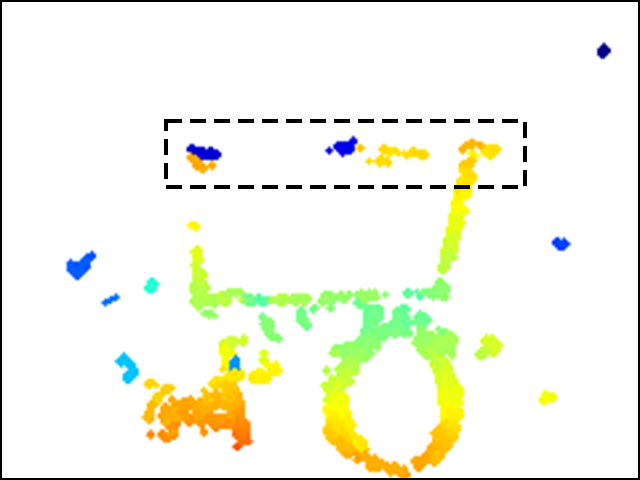}}
	\end{minipage} \hspace{4pt}&      
	\begin{minipage}[b]{\figWidth}
		\centering
		{\includegraphics[width=\linewidth]{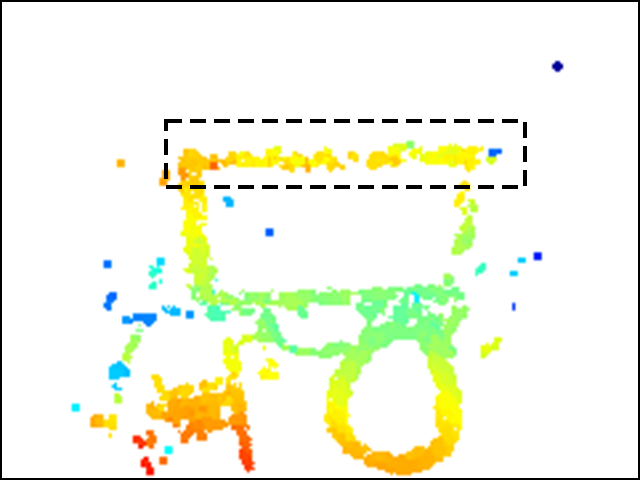}}
	\end{minipage}  \vspace{0pt}\\
 
    \rotatebox{90}{\makecell[c]{\ \ \ \emph{rpg\_reader}}}&
    \begin{minipage}[b]{\figWidth}
		\centering
		{\includegraphics[width=\linewidth]{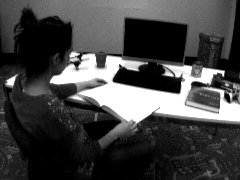}}
    \end{minipage} \hspace{4pt}&
	\begin{minipage}[b]{\figWidth}
		\centering
		{\includegraphics[width=\linewidth]{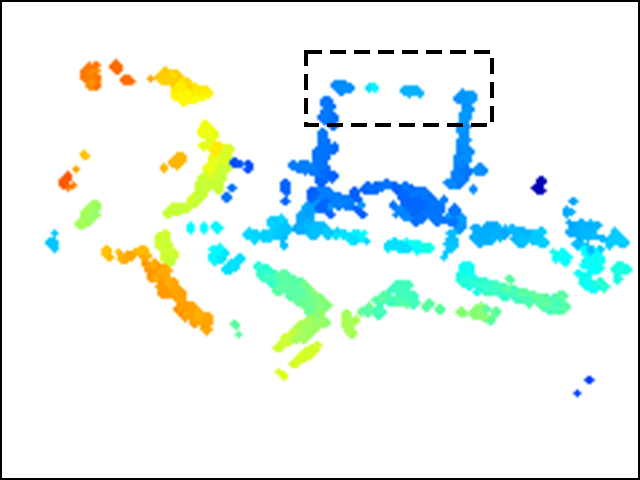}}
	\end{minipage} \hspace{4pt}&      
	\begin{minipage}[b]{\figWidth}
		\centering
		{\includegraphics[width=\linewidth]{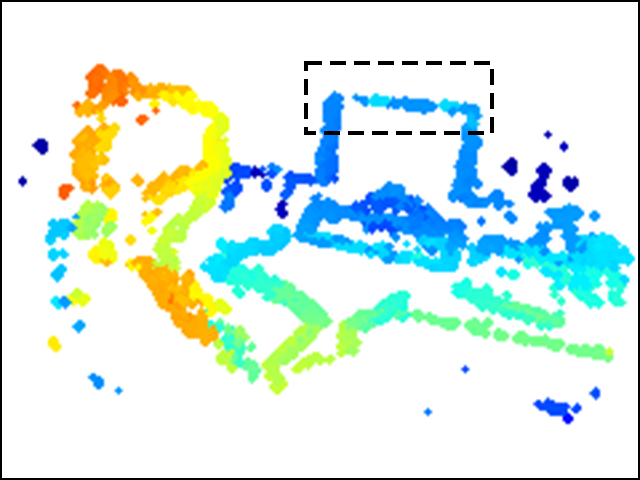}}
	\end{minipage}  \vspace{0pt}\\
    \rotatebox{90}{\makecell[c]{\emph{tumvie\_6d}}}&
    \begin{minipage}[b]{\figWidth}
		\centering
		{\includegraphics[width=\linewidth]{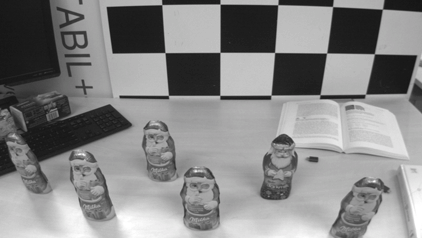}}
    \end{minipage} \hspace{4pt}&     
	\begin{minipage}[b]{\figWidth}
		\centering
		{\includegraphics[width=\linewidth]{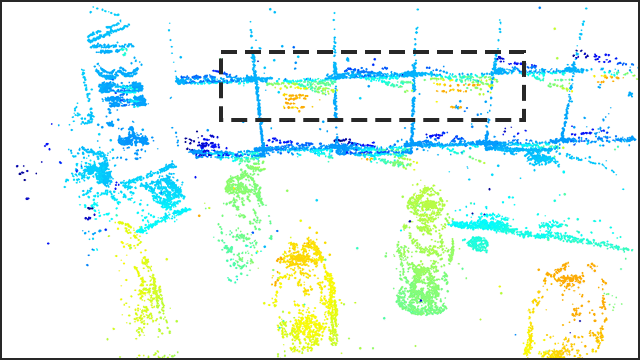}}
	\end{minipage} \hspace{4pt}&      
	\begin{minipage}[b]{\figWidth}
		\centering
		{\includegraphics[width=\linewidth]{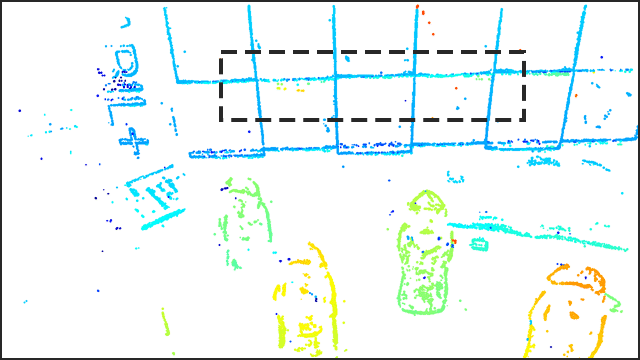}}
	\end{minipage} \vspace{0pt}\\
    \rotatebox{90}{\makecell[c]{\emph{tumvie\_3d}}}&%
    \begin{minipage}[b]{\figWidth}
		\centering
		{\includegraphics[width=\linewidth]{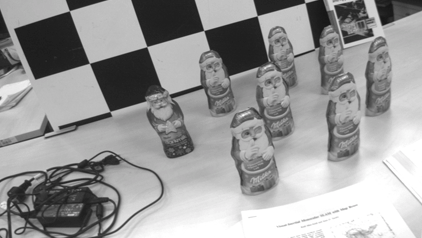}}
    \end{minipage} \hspace{0pt}&
	\begin{minipage}[b]{\figWidth}
		\centering
		{\includegraphics[width=\linewidth]{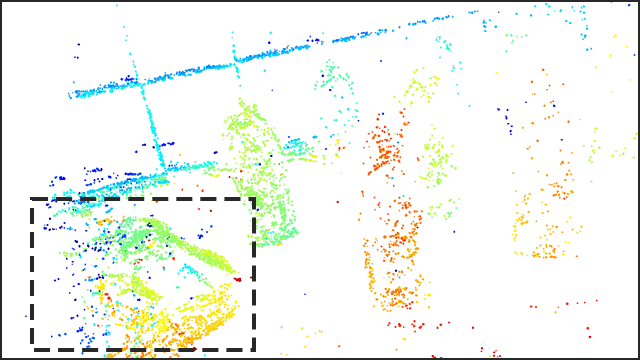}}
	\end{minipage} \hspace{0pt}&
	\begin{minipage}[b]{\figWidth}
		\centering
		{\includegraphics[width=\linewidth]{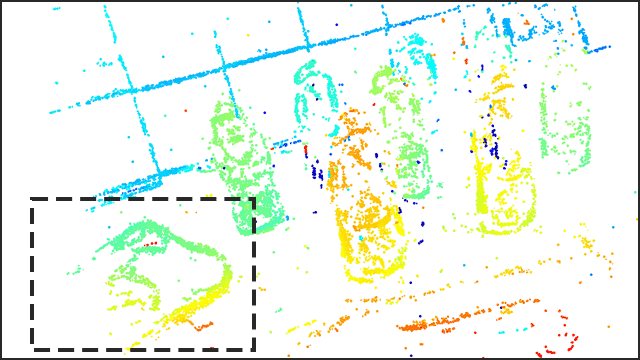}}
	\end{minipage} \vspace{0pt}
\end{tabular}
}
\caption{\label{fig:mapping evaluation on DSEC/RPG}\emph{Qualitative comparison of mapping results}.
Intensity images in the first column are used only for visualization.
The second and third columns show the estimated inverse depth maps by ESVO~\cite{zhou2021event} and our method, respectively. 
Note that our method returns more accurate and complete reconstruction results for horizontal edges.
Inverse depth maps are color coded, from red (close) to blue (far) over a white background, in the range of $\textrm{$0.5$ m--$10$ m}$ for the \emph{rpg} dataset, $\textrm{$1$ m--$50$ m}$ for the \emph{DSEC} dataset, and $\textrm{$0.33$ m--$10$ m}$ for the \emph{TUM-VIE} dataset.}
\end{figure}




\subsection{Comparison of Mapping: ESVO vs Ours}
\label{subsec:comparison of mapping}


We compare our mapping results against those of the original ESVO pipeline~\cite{zhou2021event} to demonstrate the performance improvement of our system in mapping.
As illustrated in Fig.~\ref{fig:mapping evaluation on DSEC/RPG}, the results of ESVO exhibit that horizontal structures in the environment are either not reconstructed or inaccurate in depth estimates.
This is because the static-stereo method cannot precisely recover the depth of structures that are parallel to the stereo-camera baseline. 
In comparison, our system additionally incorporates a temporal stereo method, which allows for accurate depth estimation of structures parallel to the stereo-camera baseline (see Sec.~\ref{subsubsec:temporal stereo matching}). 
Besides, the sampled points from AA better capture edge structures with less noise (see last row of Fig.~\ref{fig:mapping evaluation on DSEC/RPG}).
All of these lead to better performance in terms of reconstruction completeness and local smoothness.

We also provide a quantitative evaluation, where the figures of merit consist of the \emph{mean} depth error (defined as $\frac{1}{N}\sum_{i}^{N}\vert d_{i,\text{est}} - d_{i,\text{gt}}\vert$), the \emph{median} depth error, and the relative error (defined as $\frac{1}{N}\sum_{i}^{N}\vert d_{i,\text{est}} - d_{i,\text{gt}}\vert / d_{i,\text{gt}}$), respectively.
Sequences that frequently capture horizontal structures are selected for the evaluation.
Depth measurements from a LiDAR are adopted as ground truth, and only those within 20~m are used in the evaluation because we find that some of them beyond that range are either less accurate or not aligned well with visual data.
As shown in Tab.~\ref{tab:mapping evaluation}, the best results under each metric are highlighted in bold, demonstrating that our method outperforms ESVO in almost all evaluation metrics. 


\begin{table}[t]
\centering
\caption{\label{tab:mapping evaluation}\emph{Quantitative comparison of mapping results.}
Depth range refers to the mean of true depth of the evaluated points.}
\begin{adjustbox}{max width=\linewidth}
\begin{tabular}{lrlcc}
\toprule
\textbf{Sequence} & \textbf{Depth range} & \textbf{Statistic} & \textbf{ESVO} & \textbf{Ours}\\
\midrule

\emph{dsec\_city04\_a} 
& 11.54 m
& Mean error & 1.84 m & \textbf{1.64 m}\\
& & Median error & 1.45 m & \textbf{1.29 m}\\
& & Relative error & 15.15\% & \textbf{12.57\%} \\[0.9ex]

\emph{dsec\_city04\_c} 
& 9.75 m 
& Mean error & 1.22 m & \textbf{0.99 m}\\
& & Median error & \textbf{0.37 m} & 0.42 m\\
& & Relative error & 11.45\% & \textbf{9.53\%} \\[0.9ex] 

\emph{dsec\_city04\_d} 
& 11.37 m 
& Mean error & 1.72 m & \textbf{1.59 m}\\
& & Median error & 1.83 m & \textbf{1.35 m}\\
& & Relative error & 13.52\% & \textbf{11.20\%} \\ 
\bottomrule
\end{tabular}
\end{adjustbox}
\end{table}
\subsection{Full System Evaluation}
\label{subsec:system evaluation}

To assess the performance of camera pose tracking of our system, we conduct tests using all five datasets of different resolutions and compare the motion estimation results with another five stereo event-based VO/VIO pipelines~\cite{zhou2021event, liu2023esvio, niu2024imu, chen2023esvio, Ghosh24eccvw}.
Among them, ESVO~\cite{zhou2021event} is the first stereo event-based VO system that employs a direct method. 
Successive works, \eg, ESVIO~\cite{liu2023esvio} and our ICRA publication~\cite{niu2024imu}, are both direct methods that additionally introduce an IMU.
Note that a feature-based method, presented in \cite{chen2023esvio}, leverages both standard cameras and event cameras.
To ensure a fair comparison, only the pure event-based part (denoted by ESIO~\cite{chen2023esvio}) is enabled and used as a comparative.
Event-based Stereo Parallel Tracking and Mapping (ES-PTAM)~\cite{Ghosh24eccvw} is the latest event-only direct stereo VO system, with its mapping module built upon Multi-Camera Event-based Multi-View Stereo (MC-EMVS)~\cite{Ghosh22aisy}.
Note that some systems only provide raw data over several sequences. 
Therefore, we can only use the provided trajectories in our comparison and denote the missing ones with ``-'' in 
the tables.
We use \text{``failed''} to denote cases where the open-sourced systems failed to run successfully.

\begin{table*}[!ht]
\begin{minipage}{\textwidth}
\centering
\caption{\label{tab:ate_eval} Absolute rotation error and Absolute trajectory error (RMS) on
datasets of multiple resolutions [$\mathbf{ARE}$:$^\circ$, $\mathbf{ATE}$:cm].
ESVO and ES-PTAM are event-only methods, whereas the rest process input stereo events and inertial (IMU) data.
}
\begin{adjustbox}{max width=\linewidth}
\renewcommand{\arraystretch}{1.11}
\setlength{\tabcolsep}{10pt}
\begin{tabular}{ll*{12}{S[table-format=2.2,table-number-alignment=center]}}
\toprule
\multirow{2}{*}{\textbf{Dataset}} 
& \multirow{2}{*}{\textbf{Sequence}} 
& \multicolumn{2}{c}{\textbf{ESVO}~\cite{zhou2021event}} 
& \multicolumn{2}{c}{\textbf{ES-PTAM}~\cite{Ghosh24eccvw}} 
& \multicolumn{2}{c}{\textbf{ESIO}~\cite{chen2023esvio}} 
& \multicolumn{2}{c}{\textbf{ESVIO}~\cite{liu2023esvio}} 
& \multicolumn{2}{c}{\textbf{ICRA'24}~\cite{niu2024imu}} 
& \multicolumn{2}{c}{\textbf{Ours}}\\
\cmidrule(l{1mm}r{1mm}){3-4} 
\cmidrule(l{1mm}r{1mm}){5-6} 
\cmidrule(l{1mm}r{1mm}){7-8} 
\cmidrule(l{1mm}r{1mm}){9-10}
\cmidrule(l{1mm}r{1mm}){11-12}
\cmidrule(l{1mm}r{1mm}){13-14} 
{~}&{~}&$\mathbf{ARE}$&$\mathbf{ATE}$&$\mathbf{ARE}$&$\mathbf{ATE}$&$\mathbf{ARE}$&$\mathbf{ATE}$&$\mathbf{ARE}$&$\mathbf{ATE}$&$\mathbf{ARE}$&$\mathbf{ATE}$&$\mathbf{ARE}$&$\mathbf{ATE}$\\
\midrule

\multirow{4}{*}{\emph{rpg}} & \emph{box} & 5.95 & 5.80 & 6.62 & \bnum{4.06} & 12.35 & 11.38 & 4.25 & 4.41 & 3.30 & 6.67 & \bnum{2.79} & 4.31 \\
{~} & \emph{monitor } & 6.21 & 3.30 & \bnum{1.52} & 2.34 & 17.67 & 7.87 & 3.79 & 3.48 & 3.53 & 2.80 & 2.19 & \bnum{2.31} \\
{~} & \emph{bin}  & 1.98 & 2.80 & 3.29 & 2.57 & 11.76 & 7.08 & 3.87 & 2.28 & 3.54 & 5.90 & \bnum{1.23} & \bnum{2.27} \\
{~} & \emph{desk}  & 10.49 & 3.20 & \bnum{3.44} & 2.84 & 3.60 & 3.16 & 7.09 & 2.03 & 6.72 & 5.33 & 4.11 & \bnum{1.57} \\
{~} & \emph{reader}  & 3.80 & 6.60 & \novalue & \novalue & \text{failed} & \text{failed} & \novalue & \novalue & 1.78 & 3.88 & \bnum{1.54} & \bnum{2.68} \\
\noalign{\vskip 2pt} \hline \noalign{\vskip 2pt}

\multirow{4}{*}{\emph{MVSEC}} & \emph{indoor\_1}  & 4.40 & 16.59 & 14.93 & 15.02 & 22.32 & 820.36 & 6.57 & 9.63 & 11.16 & 17.65 & \bnum{1.69} & \bnum{7.63} \\
{~} & \emph{indoor\_2}  & 5.69 & 14.94 & \novalue & \novalue & 43.49 & 417.85 & \novalue & \novalue & 12.14 & 17.52 & \bnum{4.53} & \bnum{10.05} \\
{~} & \emph{indoor\_3}  & 2.94 & 10.03 & \novalue & \novalue & \text{failed} & \text{failed} & 3.01 & 8.06 & 2.73 & 10.45 & \bnum{2.63} & \bnum{7.35} \\
{~} & \emph{indoor\_4}  & \text{failed} & \text{failed} & \novalue & \novalue & 32.15 & 173.51 & \novalue & \novalue & \text{failed} & \text{failed} & \bnum{10.96} & \bnum{5.59} \\ 
\noalign{\vskip 2pt} \hline \noalign{\vskip 2pt}

\multirow{9}{*}{\emph{DSEC}} & \emph{city04\_a}  & 8.17 & 370.32 & \bnum{3.17} & 131.62 & 5.07 & 940.80 & {5.59} & 201.53 & 4.36 & 103.85 & 3.33 & \bnum{56.17} \\
{~} & \emph{city04\_b}  & 1.97 & 115.56 & 2.04 & 29.02 & 3.69 & 434.87 & \bnum{1.38} & \bnum{48.33} & 2.64 & 66.80 & 1.55 & 73.83 \\
{~} & \emph{city04\_c}  & 14.04 & 932.84 & 6.02 & 1184.37 & \bnum{7.35} & 1153.69 & 14.62 & 1400.76 & 11.39 & 637.13 & 10.26 & \bnum{508.71} \\
{~} & \emph{city04\_d}  & 21.62 & 2676.11 & 37.13 & 1053.87 & \bnum{3.63} & 6822.53 & \novalue & \novalue & 14.14 & 732.13 & 8.87 & \bnum{546.58} \\
{~} & \emph{city04\_e}  & 6.89 & 792.93 & 3.97 & 75.9 & 5.57 & 1036.23 & 6.28 & 331.70 & 5.08 & 115.82 & \bnum{3.54} & \bnum{52.94} \\
{~} & \emph{city04\_f}  & 6.36 & 1400.26 & 10.65 & 522 & 6.19 & 4595.01 & 20.78 & 1765.48 & 5.52 & 579.66 & \bnum{5.00} & \bnum{257.75} \\
{~} & \emph{city09\_b}  & 5.166670 & 606.0886 & \bnum{1.748680} & 195.1363 & \text{failed} & \text{failed} & \novalue & \novalue & 2.918167 & 192.4361 & 2.167177 & \bnum{87.8303}  \\ 
{~} & \emph{city11\_a}  & 1.96 & 366.22 & \novalue & \novalue & 2.40 & 107.36 & 7.71 & 406.11 & 4.52 & 95.75 & \bnum{1.85} & \bnum{48.77} \\
{~} & \emph{city11\_b}  & 44.57 & 3241.69 & \novalue & \novalue & \bnum{3.13} & \bnum{300.14} & \novalue & \novalue & 12.78 & 869.77 & 10.61 & 441.79  \\ 
\noalign{\vskip 2pt} \hline \noalign{\vskip 2pt}

\multirow{5}{*}{\emph{VECtor}} & \emph{robot\_normal}  & 19.79 & 7.32 & \novalue & \novalue & 5.61 & 5.17 & \novalue & \novalue & 20.46 & 15.20 & \bnum{5.03} & \bnum{4.81} \\
{~} & \emph{robot\_fast}  & \text{failed} & \text{failed} & \novalue & \novalue & \text{failed} & \text{failed} & \novalue & \novalue & \text{failed} & \text{failed} & \bnum{16.290511} & \bnum{24.1768} \\
{~} & \emph{corner\_slow}  & 9.63 & 13.70 & \novalue & \novalue & 32.67 & 2.67 & \novalue & \novalue & 6.15 & 5.52 & \bnum{2.96} & \bnum{2.15} \\
{~} & \emph{hdr\_normal}  & 24.54 & 18.40 & \novalue & \novalue & 18.42 & 27.85 & \novalue & \novalue & 14.11 & 16.06 & \bnum{7.82} & \bnum{13.53} \\
{~} & \emph{sofa\_normal}  & \text{failed} & \text{failed} & \novalue & \novalue & \bnum{12.11} & 43.94 & \novalue & \novalue & \text{failed} & \text{failed} & 20.11 & \bnum{40.28} \\
{~} & \emph{desk\_normal}  & 19.91 & 20.81 & \novalue & \novalue & \text{failed} & \text{failed} & \novalue & \novalue & 10.36 & 19.08 & \bnum{6.82} & \bnum{16.47} \\

\noalign{\vskip 2pt} \hline \noalign{\vskip 2pt}

\multirow{4}{*}{\emph{TUM-VIE}} & \emph{1d\_trans}  & 13.47 & 12.54 & \bnum{6.02} & \bnum{1.05} & \text{failed} & \text{failed} & \novalue & \novalue & 6.667275 & 3.8593 & 6.303151 & 3.3344 \\
{~} & \emph{3d\_trans}  & 19.20 & 17.19 & 15.62 & 8.53 & \text{failed} & \text{failed} & \novalue & \novalue  & 17.933383 & 18.8966 & \bnum{6.614262} & \bnum{7.2635} \\
{~} & \emph{6d\_trans}  & 17.59 & 13.46 & 14.01 & 10.25 & \text{failed} & \text{failed} &  \novalue & \novalue & \text{failed} & \text{failed} & \bnum{4.17} & \bnum{3.21} \\
{~} & \emph{desk}  & 14.56 & 12.92 & 3.37 & \bnum{2.5} & \text{failed} & \text{failed} & \novalue & \novalue & 6.946727 & 8.9920 & \bnum{2.396211} & 6.1611 \\
{~} & \emph{desk2}  & 5.86 & 4.42 & 10.12 & 7.2 & \text{failed}  & \text{failed} & \novalue & \novalue & 4.316407 & 9.4729 & \bnum{3.876920} & \bnum{4.0232} \\
\bottomrule
\end{tabular}

\end{adjustbox}
\begin{flushleft}
\vspace{-1ex}
{\emph{* ``-'' represents the lack of experimental results of the algorithm (according to the respective publication).}}
\end{flushleft}
\end{minipage}
\vspace{3em}

\begin{minipage}{\textwidth}
\centering
\caption{\label{tab:rpe_eval}Relative pose error (RMS) on
datasets of multiple resolutions [$\mathbf{R}$:$^\circ$/s, $\mathbf{t}$:cm/s].
ESVO and ES-PTAM are event-only methods, whereas the other methods process input stereo events and inertial (IMU) data.
}
\begin{adjustbox}{max width=\linewidth}
\renewcommand{\arraystretch}{1.11}
\setlength{\tabcolsep}{10pt}
\begin{tabular}{ll*{12}{S[table-format=2.2,table-number-alignment=center]}}
\toprule
\multirow{2}{*}{\textbf{Dataset}} 
& \multirow{2}{*}{\textbf{Sequence}} 
& \multicolumn{2}{c}{\textbf{ESVO}~\cite{zhou2021event}} 
& \multicolumn{2}{c}{\textbf{ES-PTAM}~\cite{Ghosh24eccvw}} 
& \multicolumn{2}{c}{\textbf{ESIO}~\cite{chen2023esvio}} 
& \multicolumn{2}{c}{\textbf{ESVIO}~\cite{liu2023esvio}} 
& \multicolumn{2}{c}{\textbf{ICRA'24}~\cite{niu2024imu}} 
& \multicolumn{2}{c}{\textbf{Ours}}\\
\cmidrule(l{1mm}r{1mm}){3-4} 
\cmidrule(l{1mm}r{1mm}){5-6} 
\cmidrule(l{1mm}r{1mm}){7-8} 
\cmidrule(l{1mm}r{1mm}){9-10}
\cmidrule(l{1mm}r{1mm}){11-12}
\cmidrule(l{1mm}r{1mm}){13-14} 
{~}&{~}&$\mathbf{R}$&$\mathbf{t}$&$\mathbf{R}$&$\mathbf{t}$&$\mathbf{R}$&$\mathbf{t}$&$\mathbf{R}$&$\mathbf{t}$&$\mathbf{R}$&$\mathbf{t}$&$\mathbf{R}$&$\mathbf{t}$\\
\midrule

\multirow{4}{*}{\emph{rpg}} & \emph{box} & 3.40 & 7.20 & \bnum{2.096231} & 4.222 & 2.910675 & 7.8742 & 5.11 & 8.71 & 2.65 & 5.16 & 3.02 & \bnum{4.18} \\
{~} & \emph{monitor}  & 1.70 & 3.20 &  \bnum{1.141536} & 2.5887 & 1.713174 & 4.2703 & 6.12 & 7.93 & 1.29 & 1.85 & 1.15 & \bnum{1.69} \\
{~} & \emph{bin}  & 1.20 & 3.10 & 1.000771 & \bnum{1.8234} & 2.301532 & 6.7769 & 4.14 & 7.64 & 1.23  &3.80 & \bnum{0.94} & 2.53 \\
{~} & \emph{desk}  & 3.10 & 4.50 & 1.770846 & 3.9895 & 3.94444 & 3.786 & 8.56 & 3.65  & 2.60 & 6.33 & \bnum{1.76} & \bnum{3.53} \\
{~} & \emph{reader}  & 2.50 & 5.60 & \novalue & \novalue & \text{failed} & \text{failed} & \novalue & \novalue & 1.37 & 4.65 & \bnum{1.32} & \bnum{2.12} \\
\noalign{\vskip 2pt} \hline \noalign{\vskip 2pt}

\multirow{4}{*}{\emph{MVSEC}} & \emph{indoor\_1}  & 1.09 & 7.38 & 1.68 & 6.89 & 2.91 & 228.84 & 1.28 & 5.92 & 2.09  & 10.72 & \bnum{0.98} & \bnum{5.05} \\
{~} & \emph{indoor\_2}  & 1.72 & 7.39 & \novalue & \novalue & 3.58 & 134.48 & \novalue & \novalue & 3.771  & 12.326 & \bnum{1.42} & \bnum{6.12} \\
{~} & \emph{indoor\_3}  & 1.0757 & 5.97 & \novalue & \novalue & \text{failed} & \text{failed} & 0.92 & 4.81 & 0.9878  & 6.052 & \bnum{0.77} & \bnum{4.75} \\
{~} & \emph{indoor\_4}  & \text{failed} & \text{failed} & \novalue & \novalue & 15.85 & 181.03 & \novalue & \novalue & \text{failed}  & \text{failed} & \bnum{3.84} & \bnum{10.36} \\ 
\noalign{\vskip 2pt} \hline \noalign{\vskip 2pt}

\multirow{9}{*}{\emph{DSEC}} & \emph{city04\_a}  & 1.02 & 69.14 & \bnum{0.529308} & 26.9175 & 0.66 & 187.03 & 0.95 & 62.85 & 0.82 & 20.91 & 0.60 & \bnum{16.98} \\
{~} & \emph{city04\_b} & 0.50 & 32.84 & 0.443704 & 26.4768 & 3.07 & 188.23 & 0.87 & 43.07 & 0.67 & 25.31 & \bnum{0.43} & \bnum{21.70} \\
{~} & \emph{city04\_c}  & 0.65 & 82.03 & 1.278875 & 86.2120 & 0.60 & 145.64 & 0.56 & 105.87 & 0.93 &  71.76 & \bnum{0.55} & \bnum{39.12} \\
{~} & \emph{city04\_d}  & 1.22 & 190.67 & 1.014114 & 68.0217 & 1.17 & 540.98 & \novalue & \novalue & 0.73 & 60.18 & \bnum{0.41} & \bnum{18.00} \\
{~} & \emph{city04\_e}  & 1.40 & 183.01 & \bnum{0.248299} & \bnum{14.0526} & 1.31 & 461.23 & 0.81 & 101.28 & 0.71 & 50.78 & 0.38 & 15.41 \\
{~} & \emph{city04\_f}  & 1.07 & 267.62 & 1.269052 & 49.3945 & 1.29 & 626.87 & 0.96 & 133.11 & 2.26 & 290.56 & \bnum{0.38} & \bnum{20.05} \\
{~} & \emph{city09\_b}  & 0.265498 & 38.5165 & \bnum{0.172920} & 22.4357 & \text{failed} & \text{failed} & \novalue & \novalue & 0.240206 & 20.2524 & 0.213839 & \bnum{17.8350} \\
{~} & \emph{city11\_a}  & 0.56 & 56.82 & \novalue & \novalue & 0.61 & 54.53 & 0.49 & 54.02 & 0.53 & 20.34 & \bnum{0.30} & \bnum{12.94} \\
{~} & \emph{city11\_b}  & 3.34 & 188.97 & \novalue & \novalue & 0.57 & 53.55 & \novalue & \novalue & 0.95 & 91.94 & \bnum{0.45} & \bnum{14.83}  \\ 
\noalign{\vskip 2pt} \hline \noalign{\vskip 2pt}

\multirow{5}{*}{\emph{VECtor}} & \emph{robot\_normal}  & 6.72 & 16.79 & \novalue & \novalue & \bnum{2.08} & 3.38 & \novalue & \novalue & 4.92 & 6.18 & 2.36 & \bnum{3.16} \\
{~} & \emph{robot\_fast}  & \text{failed} & \text{failed} & \novalue & \novalue & \text{failed} & \text{failed} & \novalue & \novalue & \text{failed} & \text{failed} & \bnum{5.047249} & \bnum{8.2239} \\
{~} & \emph{corner\_slow}  & 4.07 & 6.45 & \novalue & \novalue & 1.93 & 7.01 & \novalue & \novalue & 1.71 & 3.16 & \bnum{1.34} & \bnum{1.91} \\
{~} & \emph{hdr\_normal}  & 2.27 & 5.55 & \novalue & \novalue & 3.96 & 15.65 & \novalue & \novalue & 2.44  & 6.54 & \bnum{1.53} & \bnum{3.79} \\
{~} & \emph{sofa\_normal}  & \text{failed} & \text{failed} & \novalue & \novalue & \bnum{2.03} & 54.32 & \novalue & \novalue  & \text{failed} & \text{failed} & 7.15 & \bnum{13.585} \\
{~} & \emph{desk\_normal}  & 2.43 & 5.12 & \novalue & \novalue & \text{failed} & \text{failed} & \novalue & \novalue & 2.82 & 5.76 & \bnum{1.68} & \bnum{3.50} \\ 
\noalign{\vskip 2pt} \hline \noalign{\vskip 2pt}

\multirow{4}{*}{\emph{TUM-VIE}} & \emph{1d\_trans}  &  3.689574  & 8.4130 &  \bnum{0.230730} & \bnum{0.7062} & \text{failed} & \text{failed} & \novalue & \novalue &  0.973131 & 1.2108  & 0.813020 & 1.0602  \\
{~} & \emph{3d\_trans}  & 3.407809 & 7.4727 & 2.576669 & 4.9384 & \text{failed} & \text{failed} & \novalue & \novalue  &  1.793471  & 3.9808 & \bnum{0.830567} & \bnum{2.3176} \\
{~} & \emph{6d\_trans}  & 9.493119 & 14.2761 &  3.607863 & 12.0235 & \text{failed} & \text{failed} &  \novalue & \novalue &  \text{failed} & \text{failed} & \bnum{1.16} & \bnum{2.49}\\
{~} & \emph{desk}  & 2.93434 & 5.8707 &  0.989709 & \bnum{2.5250} & \text{failed} & \text{failed} & \novalue & \novalue & 1.002557 & 3.1374 & \bnum{0.984052} & 3.4097 \\
{~} & \emph{desk2}  & 2.122827 & 6.2277 & 2.978887 & 4.0370 & \text{failed}  & \text{failed} & \novalue & \novalue &  2.148878 & 4.1981  & \bnum{0.909677} & \bnum{2.4127}  \\

\bottomrule
\end{tabular}
\end{adjustbox}
\end{minipage}
\end{table*}
  
\begin{figure*}[!ht]
	\centering
    \vspace{0em}
    \begin{minipage}{\textwidth}
    \centering
    \gframe{\includegraphics[width=0.226\textwidth]{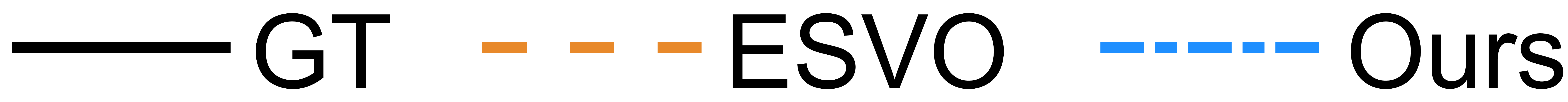}}\\
    
	\subfigure[\emph{rpg\_box}]{\includegraphics[width=0.225\textwidth]{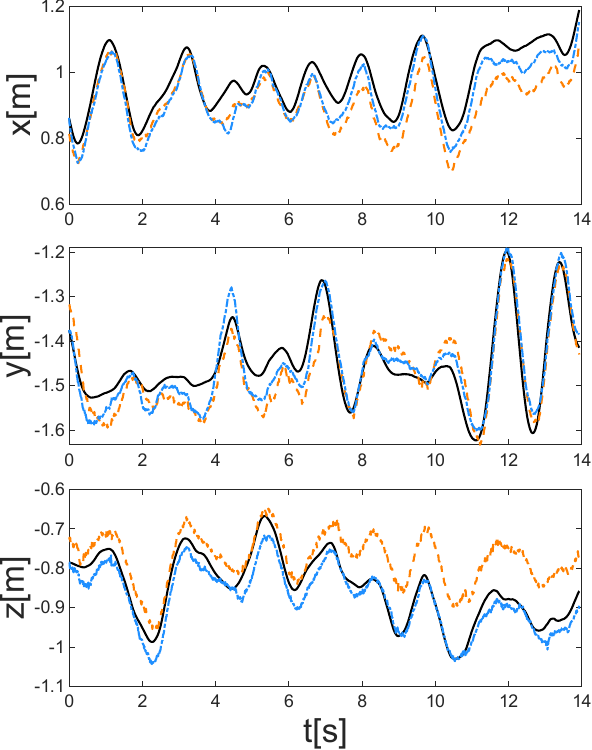}}\hspace{0pt}
	\subfigure[\emph{rpg\_monitor}]{\includegraphics[width=0.223\textwidth]{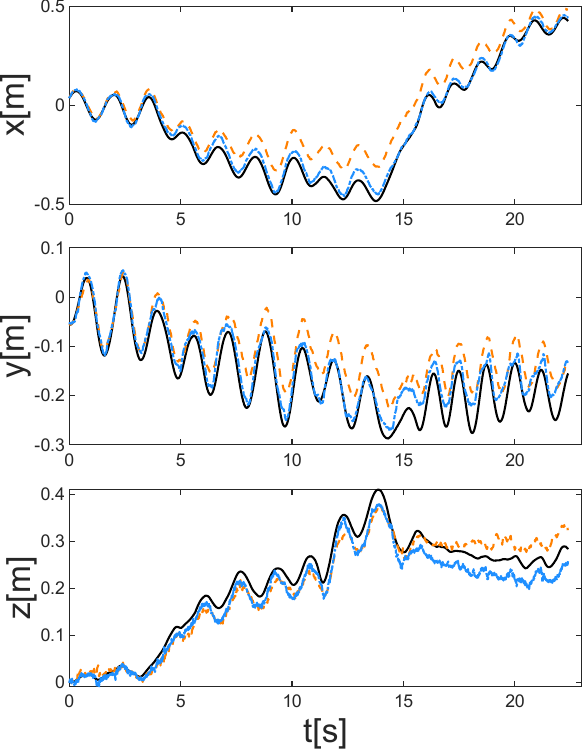}}\hspace{1pt}
    \subfigure[\emph{rpg\_reader}]{\includegraphics[width=0.2215\textwidth]{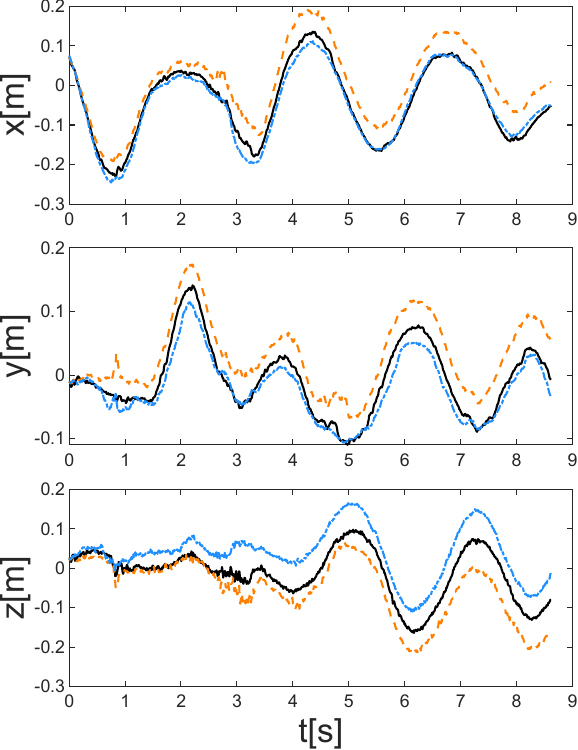}}
    \subfigure[\emph{rpg\_bin}]{\includegraphics[width=0.223\textwidth]{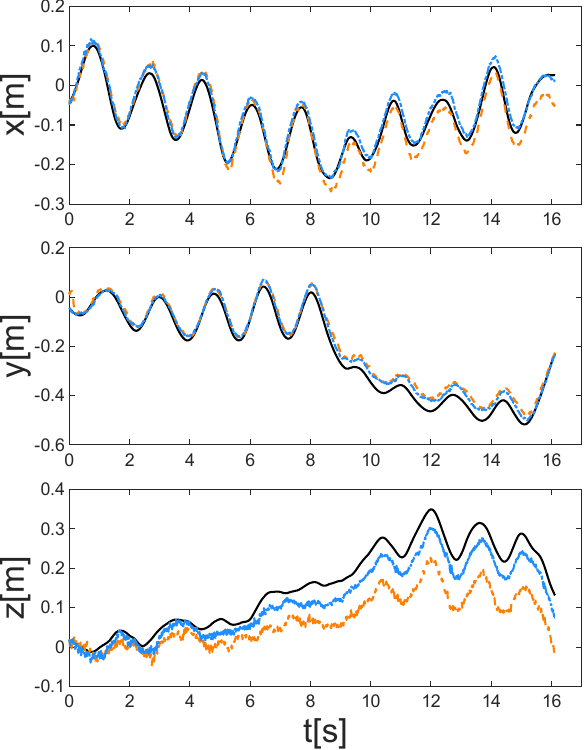}}\\[-0.5ex]
    
    \subfigure[\emph{rpg\_desk}]{\includegraphics[width=0.225\textwidth]{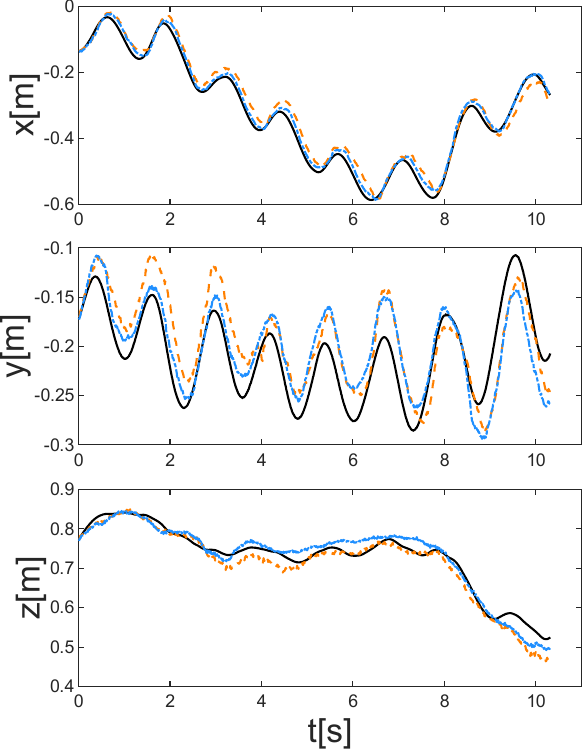}}\hspace{1pt}
	\subfigure[\emph{upenn\_indoor\_flying1}]{\includegraphics[width=0.22\textwidth]{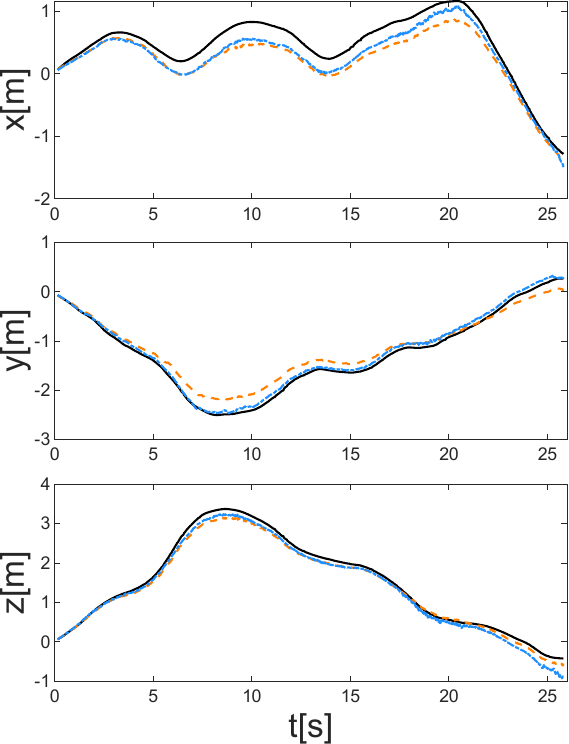}}\hspace{1pt}
    \subfigure[\emph{upenn\_indoor\_flying2}]{\includegraphics[width=0.225\textwidth]{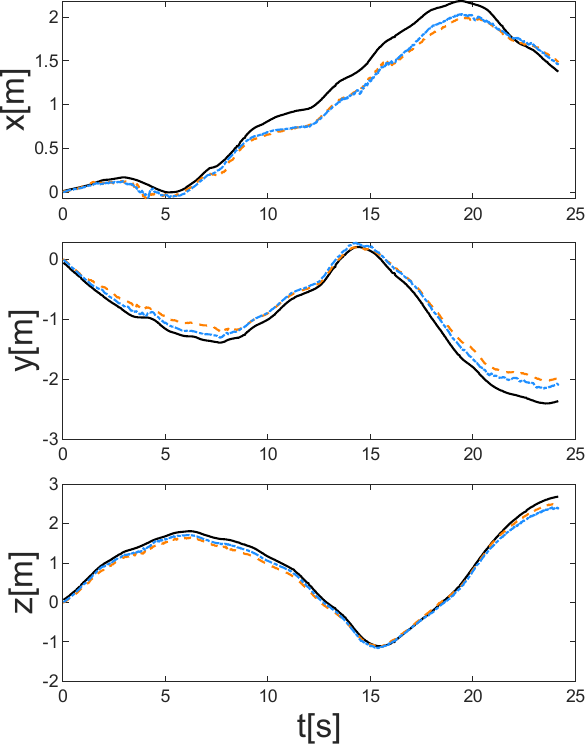}}\hspace{1pt}
    \subfigure[\emph{upenn\_indoor\_flying3}]{\includegraphics[width=0.225\textwidth]{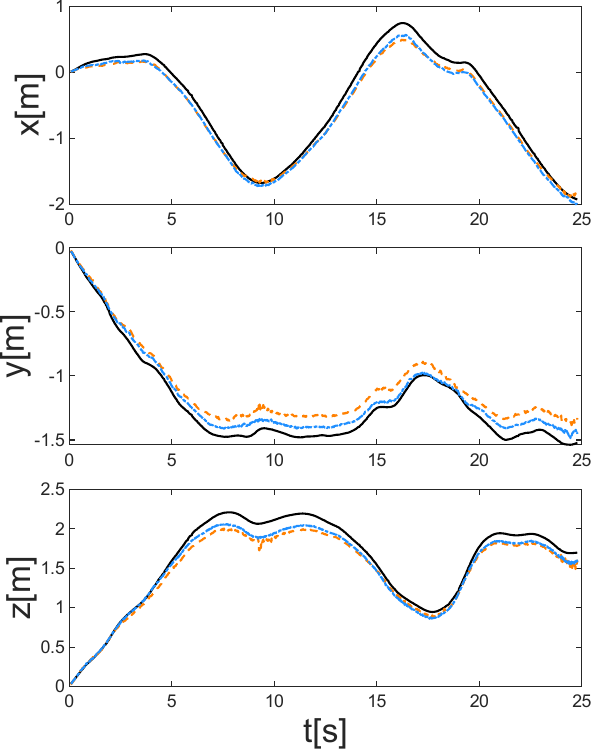}}\\[-0.5ex]
    
	\caption{\label{fig:traj_eval_rpg}Translation results on the \emph{rpg} and \emph{MVSEC} datasets, with ground truth provided by a motion capture system.
    \vspace{2em}}
    \end{minipage}
    \begin{minipage}{\textwidth}
    \centering
    \subfigure[\emph{dsec\_city04\_a}]{\includegraphics[width=0.222\textwidth]{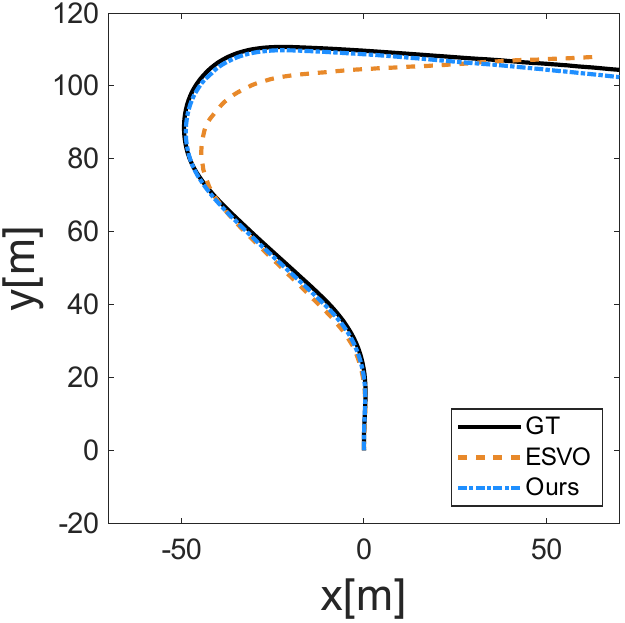}}\hspace{5pt}
	\subfigure[\emph{dsec\_city04\_b}]{\includegraphics[width=0.217\textwidth]{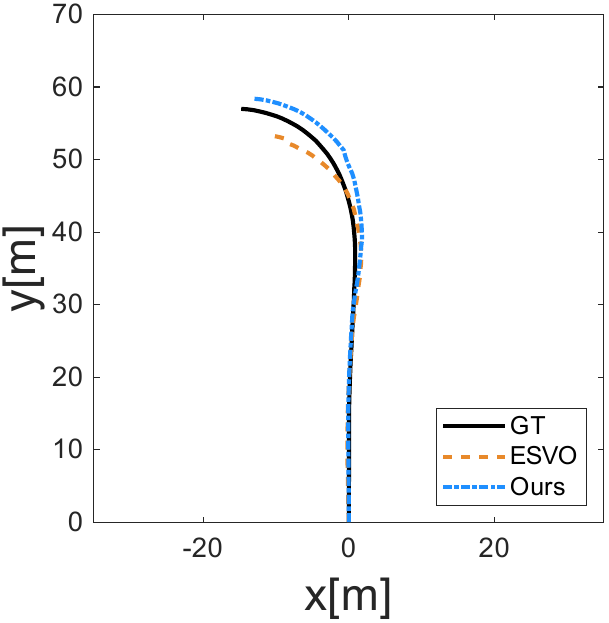}}\hspace{5pt}
    \subfigure[\emph{dsec\_city04\_c}]{\includegraphics[width=0.2275\textwidth]{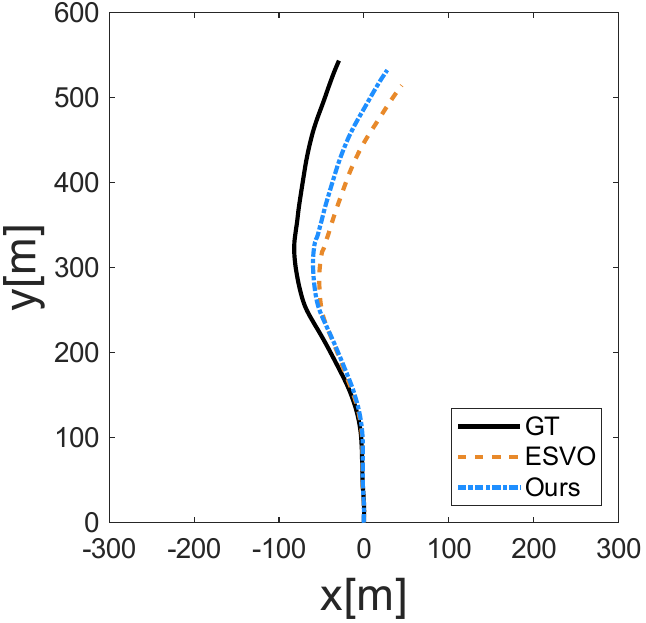}}\hspace{0pt}
	\subfigure[\emph{dsec\_city04\_d}]{\includegraphics[width=0.2275\textwidth]{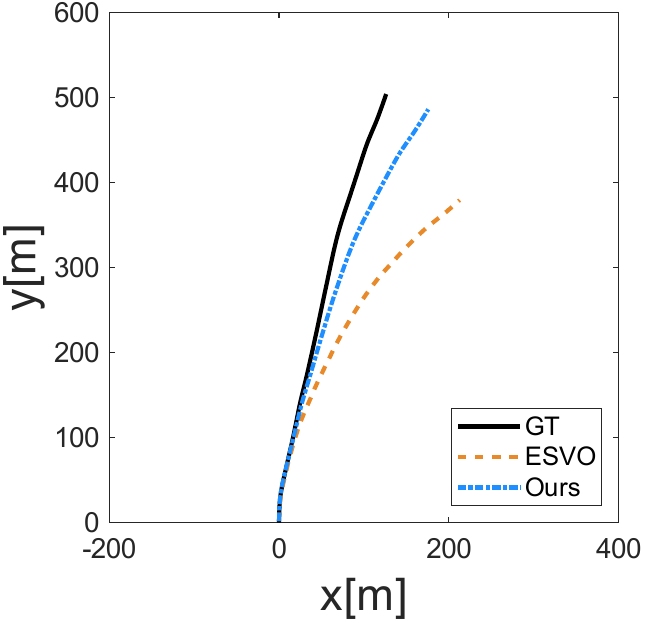}}\\[-0.5ex]
 
    \subfigure[\emph{dsec\_city04\_e}]{\includegraphics[width=0.228\textwidth]{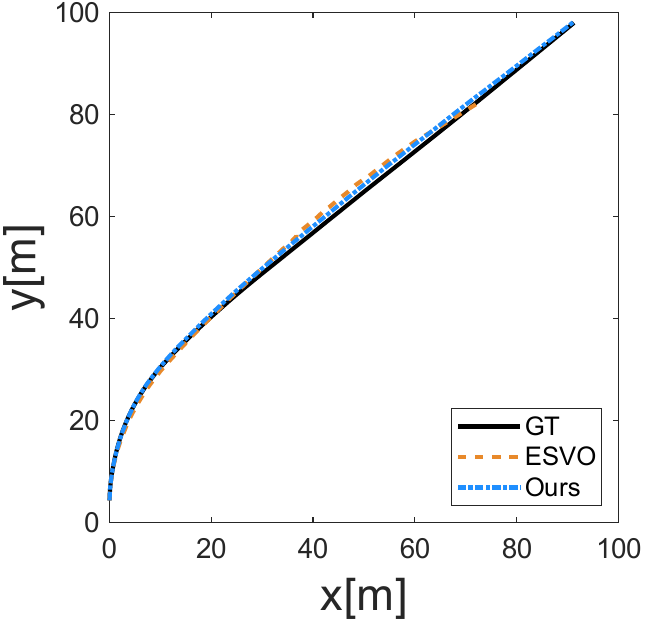}}\hspace{0pt}
	\subfigure[\emph{dsec\_city04\_f}]{\includegraphics[width=0.223\textwidth]{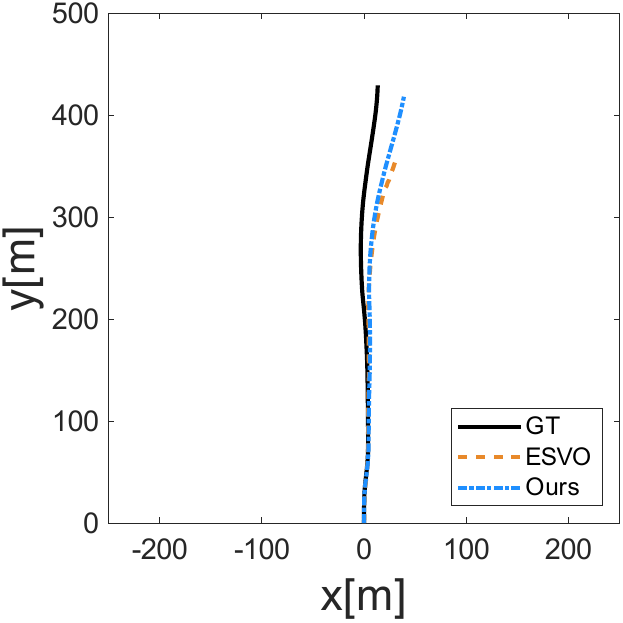}}\hspace{3pt}
    \subfigure[\emph{dsec\_city11\_a}]{\includegraphics[width=0.223\textwidth]{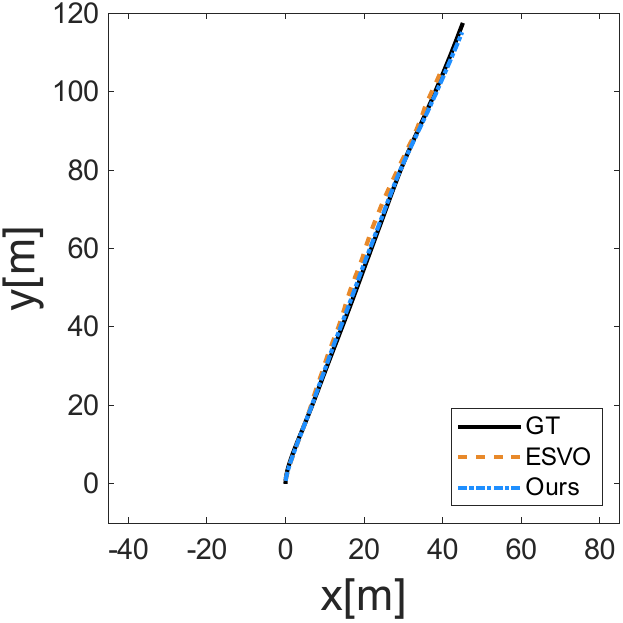}}\hspace{5pt}
	\subfigure[\emph{dsec\_city11\_b}]{\includegraphics[width=0.223\textwidth]{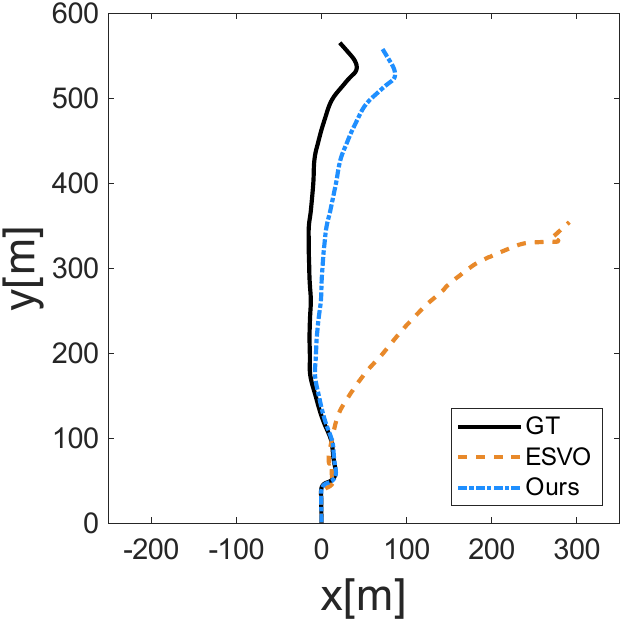}}\\[-0.5ex]
 
	\caption{\label{fig:traj_eval}Recovered trajectories on the \emph{DSEC} dataset, with ground truth provided by a LiDAR-IMU-based method~\cite{shan2020lio}.}
    \end{minipage}
\end{figure*}

To quantitatively evaluate the motion estimation results, we use two standard evaluation metrics: relative pose error (RPE) and absolute trajectory error (ATE) \cite{Sturm12iros}. 
The results of all methods can be found in Tabs.~\ref{tab:ate_eval} and \ref{tab:rpe_eval}, in which the best results are highlighted in bold.
In general, the proposed method (``Ours'') outperforms the others in terms of RPE and ATE.
Our method resolves most of the remaining issues of ESVO~\cite{zhou2021event}, thus leading to more accurate trajectory estimates than ESVO.

In contrast to ESVIO~\cite{liu2023esvio}, our method advances mainly in the pre-processing and mapping module.
The proposed contour-point sampling method and the additionally introduced temporal-stereo operation significantly improve the mapping performance, which benefits the entire system.
In addition, our better results justify that having the spatio-temporal-registration residual term incorporated into the back-end by ESVIO~\cite{liu2023esvio} does not bring obvious benefits to motion estimation while causing additional computational burden.
This may explain why ESVIO~\cite{liu2023esvio} does not demonstrate a real-time performance on event data with VGA resolution.

Compared with ESIO~\cite{chen2023esvio}, the only feature-based method in the table, our direct method makes better use of event data, circumventing the theoretical flaw of event-based feature matching.
This advantage of direct methods over feature-based methods is obvious in the results for sequences collected in large-scale environments (\eg, \emph{DSEC} in Tab.~\ref{tab:ate_eval}).
Notably, ESIO cannot operate properly on the \emph{TUM-VIE} dataset. 
We believe that one possible reason is that 
the smoothing effect on the gradient profile of the event representation, 
caused by the high spatial resolution of the camera, significantly hinders the detection and tracking of ARC*~\cite{Alzugaray18ral}, leading to the malfunction of the entire system.

In comparison to~\cite{niu2024imu}, our better performance is mainly attributed to the proposed back-end.
The better estimated IMU biases and linear velocities at recent poses facilitate camera pose tracking, thus leading to a notable improvement in tracking accuracy.
An in-depth discussion on the benefit brought by the back-end is given in Sec.~\ref{subsubsec:Ablation study Back-end}.
Finally, ES-PTAM~\cite{Ghosh24eccvw} performs exceptionally well in small-scale environments and those with repetitive textures, attributing it to the precise depth results via its multi-view stereo mapping method. 
However, thanks to the effective utilization of IMU data in the back-end, our system significantly outperforms ES-PTAM in large-scale scenarios while being on par in small-scale environments.

Additionally, ESVO2 demonstrates excellent performance in both high-speed motion and HDR scenes. 
The \emph{vector\_robot\_fast} sequence is captured indoors with a stereo event camera under aggressive motion, and only our method can operate on it and achieve good results.
Both \emph{dsec\_city\_09b} and \emph{vector\_hdr\_normal} are HDR sequences of nighttime driving and small-scale indoor scenarios, respectively. 
Benefiting from the accurate contour points sampling strategy, ESVO2 achieves the best performance on both of these sequences.

Figures~\ref{fig:traj_eval_rpg} and \ref{fig:traj_eval} provide a qualitative comparison of the trajectories estimated by ESVO and our method on sequences from \emph{rpg}, \emph{MVSEC}, and \emph{DSEC} data (To deliver a compact and clear visualization, only these trajectories are displayed.
A more complete qualitative comparison is provided in the Appendix).
Since \emph{DSEC} does not provide ground-truth trajectories, we use the result of a LiDAR-IMU--based method~\cite{shan2020lio} as ground truth.
In short, the trajectories obtained by our method align better with the ground truth, as evidenced by the quantitative analysis.
\begin{figure}[t]
    \centering
    \includegraphics[trim={0 8cm 0 0},clip,width=.88\linewidth]{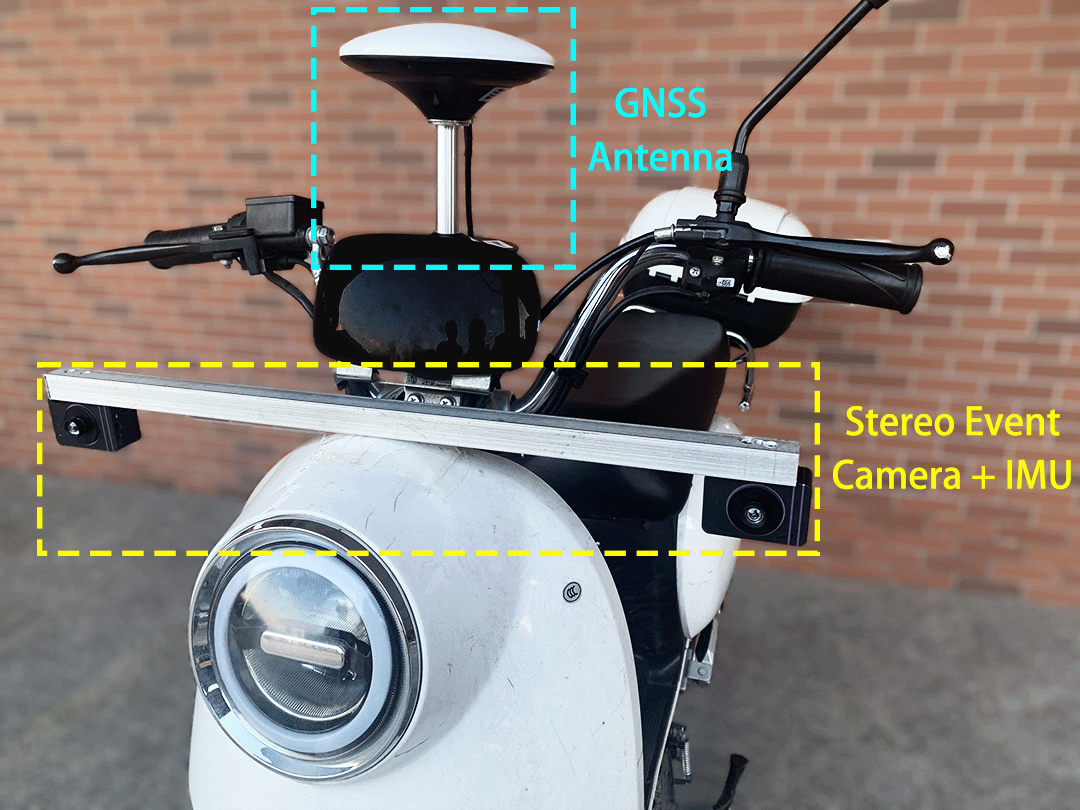}
    \caption{
    Illustration of the applied sensor suite and the motorbike used for data collection.
    }
    \label{fig:experimental setup}
    \vspace{1em}
\end{figure}

\begin{table*}[t]
\begin{center}
\caption{\label{tab:ablation_study_part1}
\emph{Results of the ablation studies (Secs.~\ref{subsubsec:Ablation study Back-end} to \ref{subsubsec:Ablation study refinement}).}
The up/down arrows indicate the variation trend compared to the reference group. 
The ``Time'' column demonstrates the average runtime of the mapping thread, followed with a tick or a cross indicating whether the mapping module can stably run in real time at a frequency of 20 Hz. 
[$\mathbf{ARE}$: $^\circ$, $\mathbf{ATE}$: m, $\mathbf{Time}$: ms]}
\setlength{\tabcolsep}{10pt}{
\begin{tabular}{l*{6}{S[table-format=2.2,table-number-alignment=center]} p{1.2cm} *{2}{S[table-format=2.2,table-number-alignment=center]} p{1.2cm}}
\toprule
\multirow{2}{*}{\textbf{Sequence}} & \multicolumn{2}{c}{\textbf{w/o back-end}} & \multicolumn{2}{c}{\textbf{w/o OS-TS}} & \multicolumn{3}{c}{\textbf{w/ refinement}} & \multicolumn{3}{c}{\textbf{Reference}} \\
\cmidrule(l{1mm}r{1mm}){2-3} 
\cmidrule(l{1mm}r{1mm}){4-5} 
\cmidrule(l{1mm}r{1mm}){6-8}
\cmidrule(l{1mm}r{1mm}){9-11}
{~}& $\mathbf{ARE}$&$\mathbf{ATE}$& $\mathbf{ARE}$&$\mathbf{ATE}$&$\mathbf{ ARE}$&$\mathbf{ATE}$ & $\mathbf{Time}$ &$\mathbf{ARE}$&$\mathbf{ATE}$ & $\mathbf{Time}$ \\
\midrule
\emph{dsec\_city04\_a} & 3.861907~$\uparrow$ & 1.34~$\uparrow$ & 4.773875~$\uparrow$ & 1.581142~$\uparrow$ & 3.01~$\downarrow$ & 0.59~$\uparrow$ & 44.70~$\uparrow$~(\ding{55})& {3.33} & {0.56} & {32.50}~(\ding{51})\\
\emph{dsec\_city04\_c} & 14.978016~$\uparrow$ & 12.90~$\uparrow$ & 13.168962~$\uparrow$ & 7.152484~$\uparrow$ & 9.50~$\downarrow$ & 4.49~$\downarrow$ & 47.55~$\uparrow$~(\ding{55}) & {10.26} & {5.09} & {34.12}~(\ding{51})\\
\emph{dsec\_city04\_d} & 9.830760~$\uparrow$ &  5.889172~$\uparrow$ & 14.867740~$\uparrow$ & 8.198073~$\uparrow$ & 7.67~$\downarrow$ & 5.37~$\uparrow$ & 47.03~$\uparrow$~(\ding{55}) & {8.87} & {5.04}  & {33.45}~(\ding{51})\\
\emph{dsec\_city04\_e} & 4.702666~$\uparrow$ & 2.63~$\uparrow$ & 4.339456~$\uparrow$ & 0.992723~$\uparrow$ & 3.10~$\downarrow$ & 0.61~$\downarrow$ & 50.85~$\uparrow$~(\ding{55}) & {3.54} & {0.76} & {30.09}~(\ding{51})\\
\emph{dsec\_city04\_f} & 5.314799~$\uparrow$ & 7.81~$\uparrow$ & 5.402240~$\uparrow$ & 3.445172~$\uparrow$ & 4.70~$\downarrow$ & 3.21~$\downarrow$ & 46.41~$\uparrow$~(\ding{55}) & {5.00} & {3.28} & {32.14}~(\ding{51})\\
\emph{dsec\_city11\_a} & 2.639011~$\uparrow$ & 1.34~$\uparrow$ & 2.436963~$\uparrow$ & 0.538048~$\uparrow$ & 2.15~$\uparrow$ & 0.54~$\uparrow$ & 49.93~$\uparrow$~(\ding{55}) & {1.85} & {0.48} & {31.96}~(\ding{51})\\
\emph{dsec\_city11\_b} & 16.307053~$\uparrow$ & 9.02~$\uparrow$ & 13.936225~$\uparrow$ & 8.011955~$\uparrow$ & 9.42~$\downarrow$ & 4.60~$\uparrow$ & 47.86~$\uparrow$~(\ding{55}) & {10.61} & {4.41} & {36.77}~(\ding{51})\\
\bottomrule
\end{tabular}
}

\end{center}
\end{table*}

\begin{table*}[t]
\begin{center}
\caption{\label{tab:ablation_study_part2}
\emph{Sensitivity with respect to the size of the sliding window and necessity of an IMU (Sec.~\ref{subsubsec:sliding window size and necessity of using IMU}).} 
The up/down arrows indicate the variation trend compared to the reference group. 
[$\mathbf{ARE}$: $^\circ$, $\mathbf{ATE}$: m]}
\setlength{\tabcolsep}{10pt}{
\begin{tabular}{l*{8}{S[table-format=2.2,table-number-alignment=center]}}
\toprule
\multirow{2}{*}{\textbf{Sequence}} & \multicolumn{2}{c}{\textbf{Window size 8}} & \multicolumn{2}{c}{\textbf{Window size 12}} & \multicolumn{2}{c}{\textbf{w/o IMU}} & \multicolumn{2}{c}{\textbf{Reference (size 5)}} \\
\cmidrule(l{1mm}r{1mm}){2-3} 
\cmidrule(l{1mm}r{1mm}){4-5} 
\cmidrule(l{1mm}r{1mm}){6-7}
\cmidrule(l{1mm}r{1mm}){8-9}
{~}& $\mathbf{ARE}$&$\mathbf{ATE}$& $\mathbf{ARE}$&$\mathbf{ATE}$& $\mathbf{ARE}$&$\mathbf{ATE}$&$\mathbf{ARE}$&$\mathbf{ATE}$ \\
\midrule
\emph{dsec\_city04\_a} & 3.447268~$\uparrow$ & 0.652636~$\uparrow$ & 2.893138~$\uparrow$ & 0.600868~$\uparrow$ & 5.734001~$\uparrow$ & 3.196463~$\uparrow$ & {3.33} & {0.56}\\
\emph{dsec\_city04\_c} & 12.231655~$\uparrow$ & 5.492721~$\uparrow$ & 11.706179~$\uparrow$ & 4.771352~$\downarrow$ & 18.024032~$\uparrow$ & 14.775397~$\uparrow$ & {10.26} & {5.09}\\
\emph{dsec\_city04\_d} & 10.807987~$\uparrow$ & 6.195458~$\uparrow$ & 9.521640~$\uparrow$ & 5.943612~$\uparrow$ & 16.382647~$\uparrow$ & 18.961562~$\uparrow$ & {8.87} & {5.04}\\
\emph{dsec\_city04\_e} & 3.446905~$\downarrow$ & 0.634652~$\downarrow$ & 3.361946~$\downarrow$ & 0.603952~$\downarrow$ & 5.057845~$\uparrow$ & 2.946473~$\uparrow$ & {3.54} & {0.76}\\
\emph{dsec\_city04\_f} & 5.705253~$\uparrow$ & 3.403583~$\uparrow$ & 5.434762~$\uparrow$ & 3.199530~$\downarrow$ & 7.085527~$\uparrow$ & 11.281578~$\uparrow$ & {5.00} & {3.28}\\
\emph{dsec\_city11\_a} & 2.466704~$\uparrow$ & 0.655389~$\uparrow$ & 2.418274~$\uparrow$ & 0.609308~$\uparrow$ & 3.845976~$\uparrow$ & 1.113627~$\uparrow$ & {1.85} & {0.48}\\
\emph{dsec\_city11\_b} & 9.650511~$\downarrow$ & 5.523101~$\uparrow$ & 9.872879~$\downarrow$ & 5.336684~$\uparrow$ & 23.773887~$\uparrow$ & 12.314442~$\uparrow$ & {10.61} & {4.41}\\
\bottomrule
\end{tabular}
}

\end{center}
\end{table*}

\subsection{Outdoor Evaluation}
\label{subsec: Outdoor Evaluation}


To further assess the versatility of our system, we collect our own outdoor dataset for evaluation.
The employed sensor suite, shown in Fig.~\ref{fig:experimental setup}, consists of a stereo event camera (DVXplorer, 640 $\times$ 480 pixels) with a baseline distance of 51 cm, an embedded 6-axis IMU, and a u-blox ZED-F9P GNSS receiver. 
The GNSS receiver owns an internal RTK positioning engine that provides ground-truth trajectories for evaluation.
All the sensors are calibrated and synchronized in advance.

\begin{figure}[t]
  \centering
    \includegraphics[trim={0 0 0 3cm},clip,width=.95\linewidth]{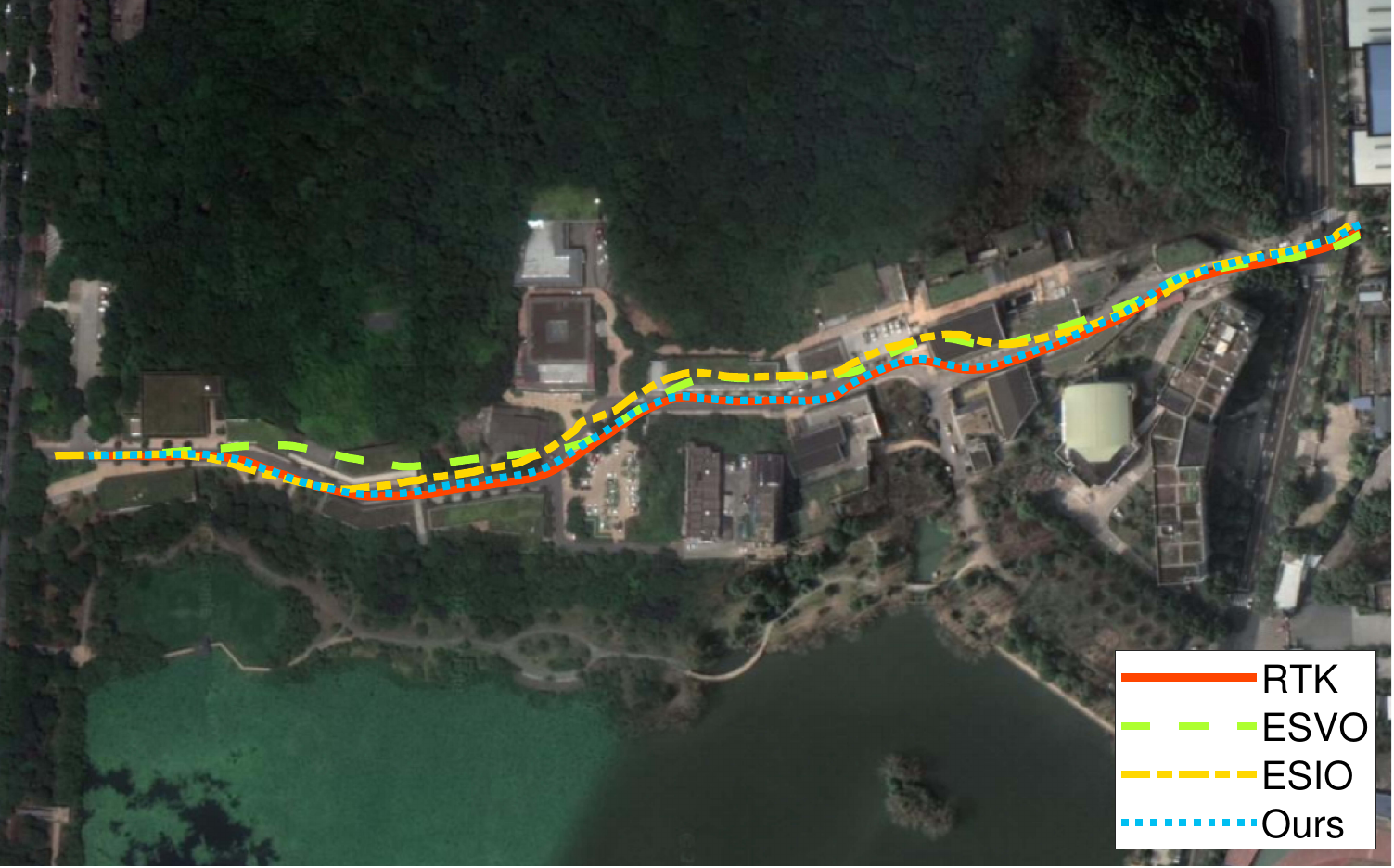}
   \captionof{figure}{
   \emph{Trajectory results on the hnu\_peachlake sequence.}
   The ground-truth trajectory is provided by RTK (red) and aligned with Google Maps. 
   Our method achieves the best performance in terms of global trajectory consistency compared to ESVO~\cite{zhou2021event} (green) and ESIO~\cite{chen2023esvio} (yellow).}
\label{fig:outdoor tracking evaluvation}
\end{figure}

Our outdoor dataset consists of two subsets (called \emph{hnu\_mapping} and \emph{hnu\_tracking}), which are used for assessing the performance of mapping and localization, respectively.
The sequences in the \emph{hnu\_mapping} subset capture scenes that vary in depth range, texture richness, and brightness condition (including HDR scenarios), etc.
Some of the mapping results can be found in Fig.~\ref{fig:outdoor mapping evaluvation}, which demonstrate good mapping performance as has been witnessed on other publicly available datasets.
The \emph{hnu\_tracking} subset consists of two sequences featuring a closed loop (\emph{hnu\_campus}, shown in Fig.~\ref{fig:eye catcher}) and a winding trajectory along a narrow street (\emph{hnu\_peachlake}, shown in Fig.~\ref{fig:outdoor tracking evaluvation}), respectively.
Compared to the other two stereo event-based visual odometry systems (\ie, ESVO and ESIO), the estimated trajectories of our system align better with the ground truth, especially exhibiting smaller drift in absolute orientation.
More importantly, our system is the only one that can operate on these sequences in real time, demonstrating the potential for practical applications.


\subsection{Ablation Studies}
\label{subsec: Ablation Studies}
\def\figWidth{0.186\linewidth} 
\begin{figure*}[t]
    \centering
    {\small
    \setlength\tabcolsep{2pt} 
    \begin{tabular}{
	>{\centering\arraybackslash}m{0.3cm} 
	>{\centering\arraybackslash}m{\figWidth} 
	>{\centering\arraybackslash}m{\figWidth}
	>{\centering\arraybackslash}m{\figWidth}
	>{\centering\arraybackslash}m{\figWidth}
	>{\centering\arraybackslash}m{\figWidth}
    }

    & Scene & Time Surface (left) & Selected points & Inverse depth map & 3D reconstruction \\
    
	\rotatebox{90}{\makecell[c]{\ \emph{HNU\_logo}}} & 
    {\includegraphics[width=\linewidth]{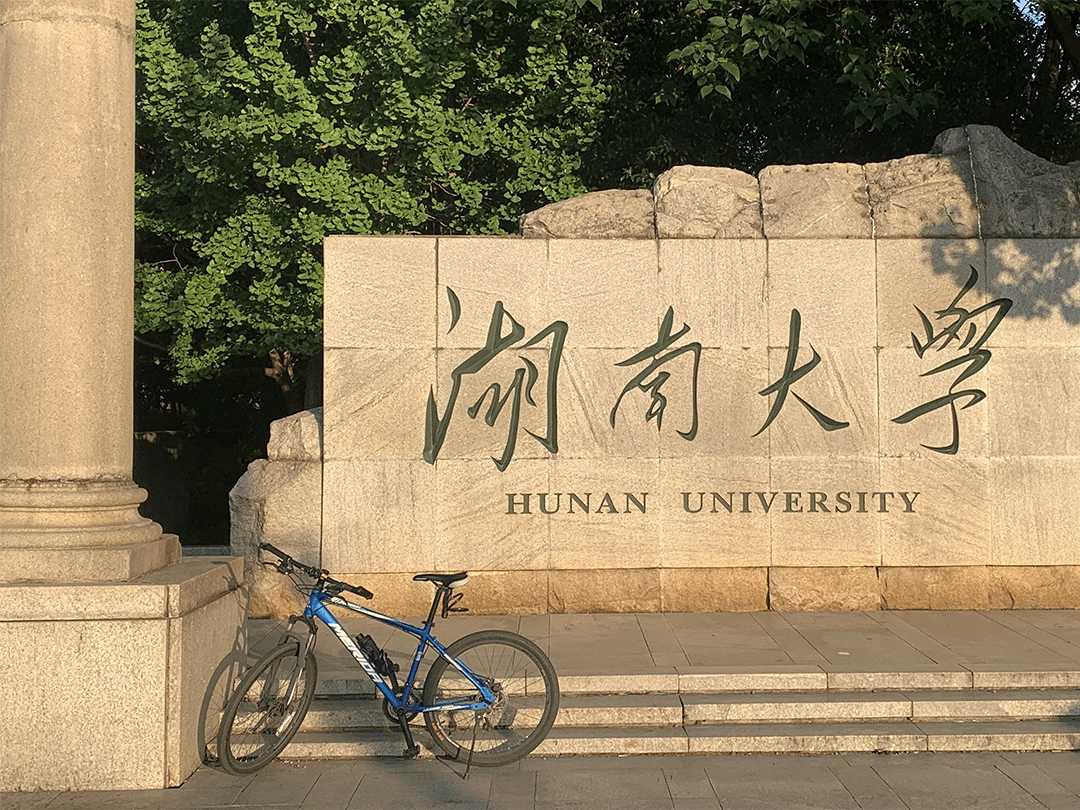}} & 
    \gframe{\includegraphics[trim={4px 4px 4px 4px},clip,width=\linewidth]{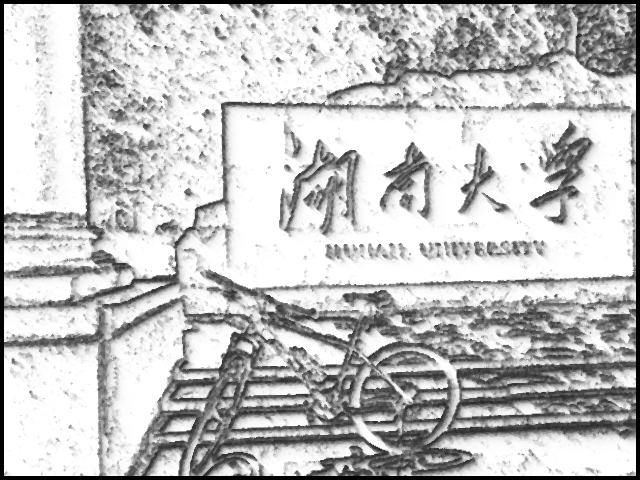}} &
    \gframe{\includegraphics[trim={4px 4px 4px 4px},clip,width=\linewidth]{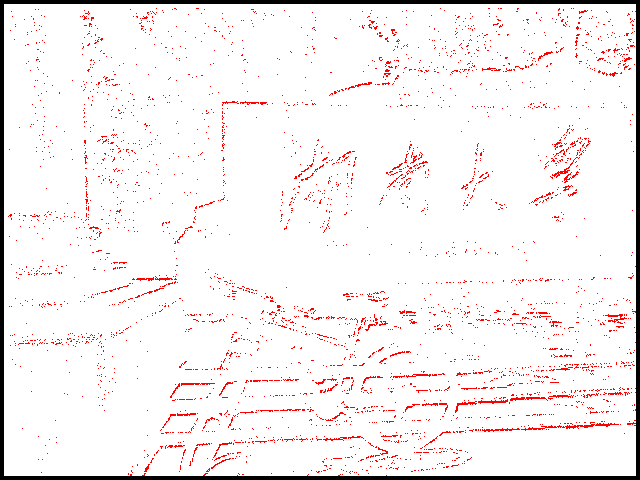}} &
    \gframe{\includegraphics[trim={4px 4px 4px 4px},clip,width=\linewidth]{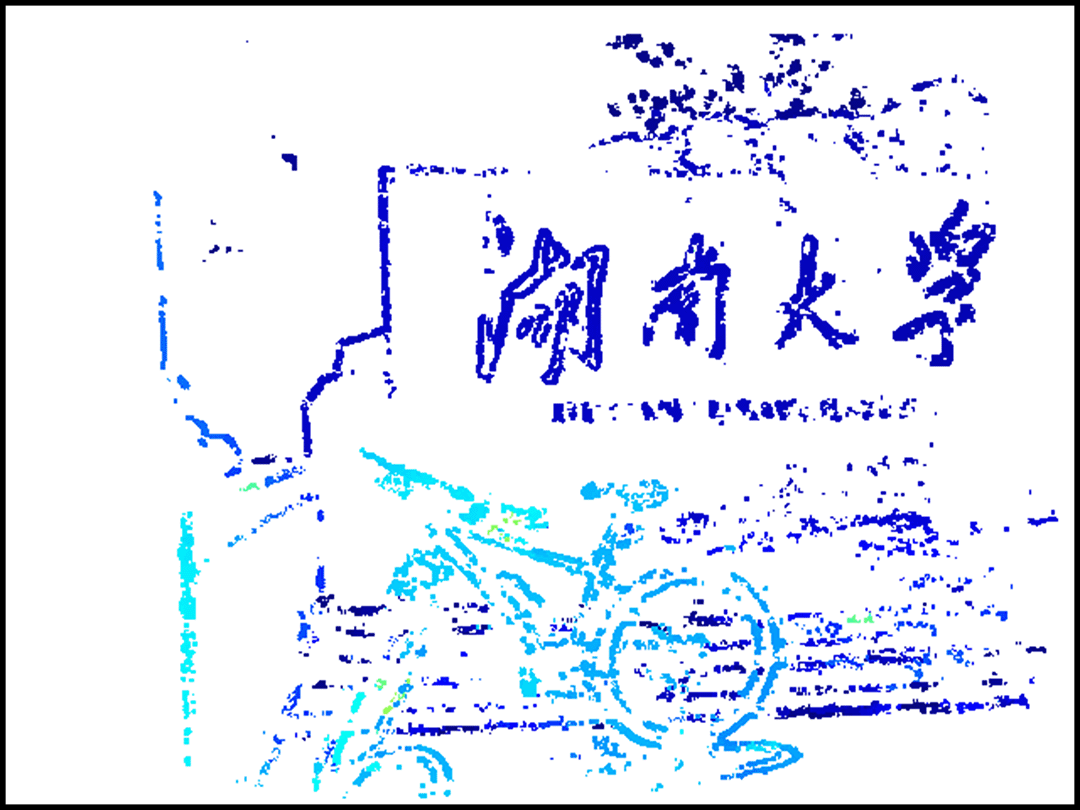}} &
	\gframe{\includegraphics[trim={4px 4px 4px 4px},clip,width=\linewidth]{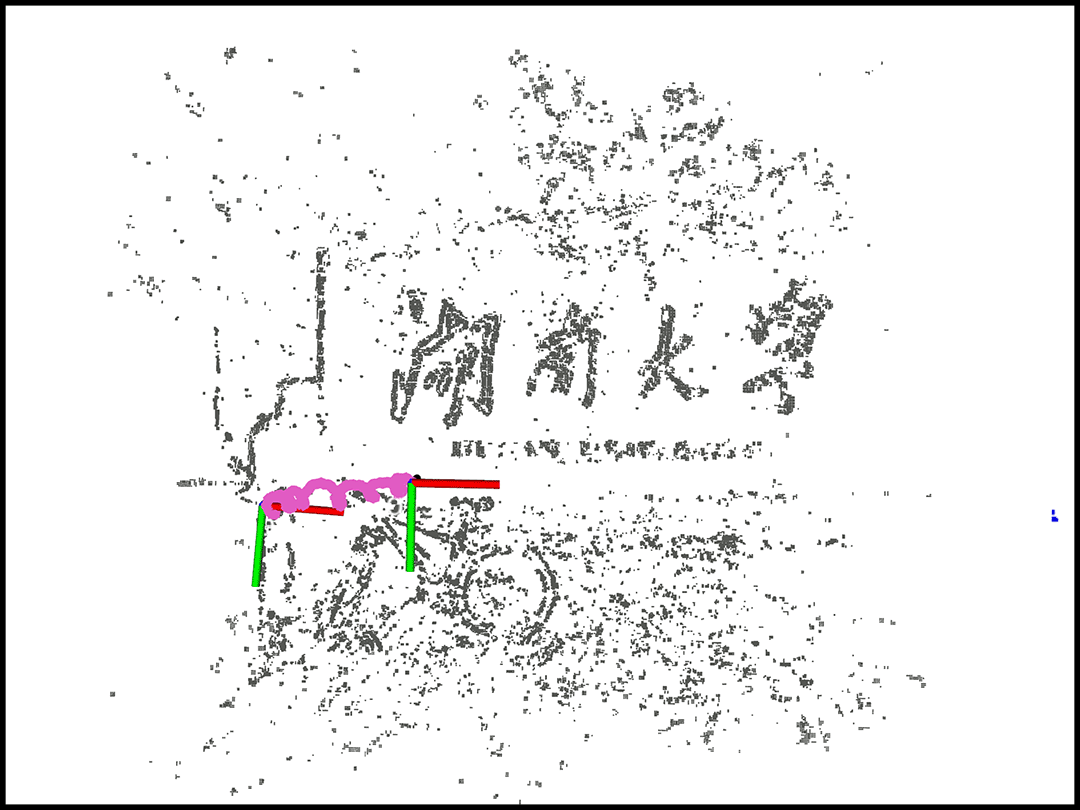}} \\

 	\rotatebox{90}{\makecell[c]{\ \emph{HNNU\_logo}}} & 
    {\includegraphics[width=\linewidth]{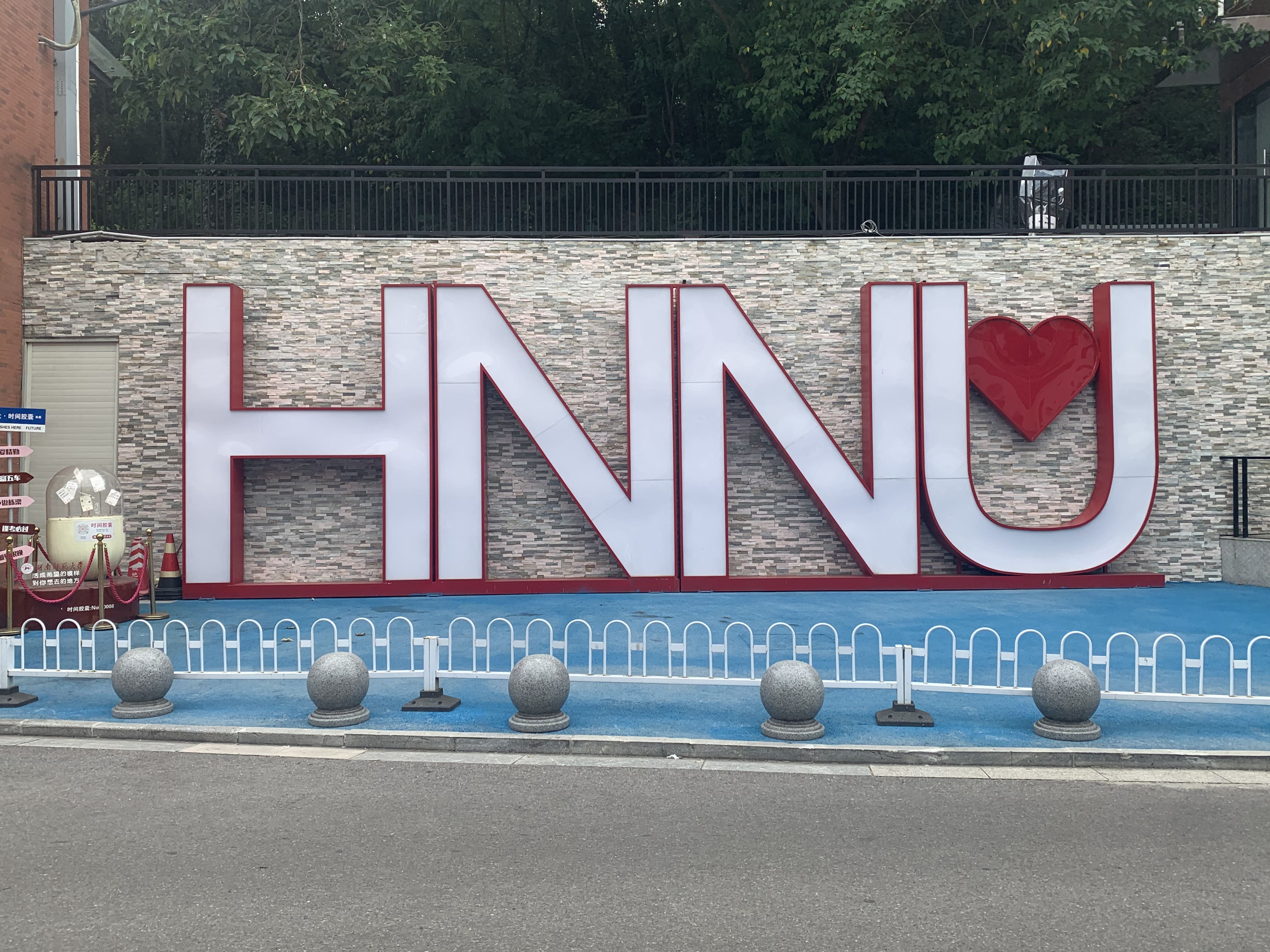}} & 
    \gframe{\includegraphics[trim={4px 4px 4px 4px},clip,width=\linewidth]{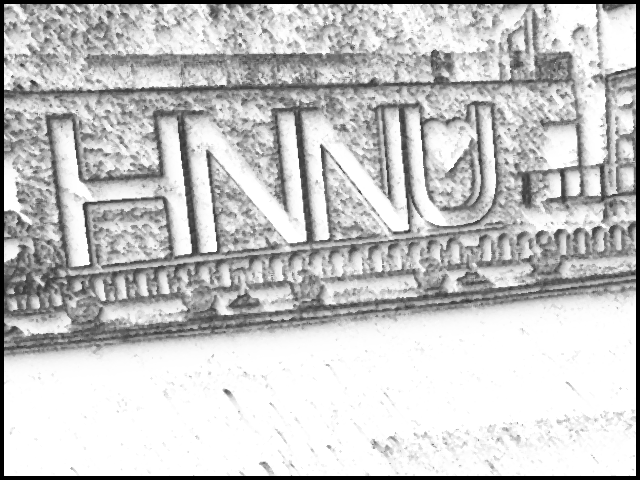}} &
    \gframe{\includegraphics[trim={4px 4px 4px 4px},clip,width=\linewidth]{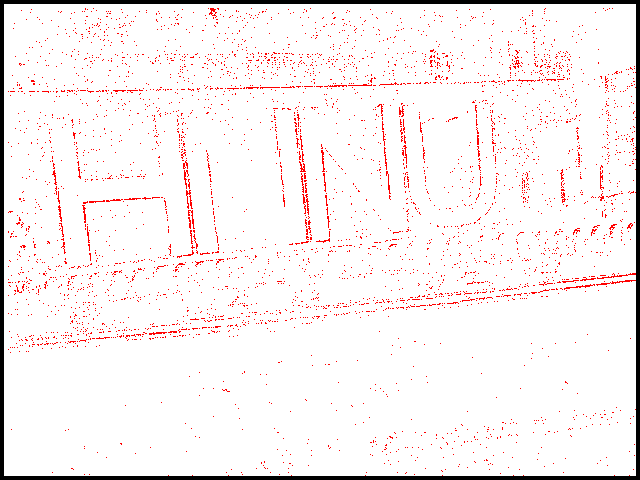}} &
    \gframe{\includegraphics[trim={4px 4px 4px 4px},clip,width=\linewidth]{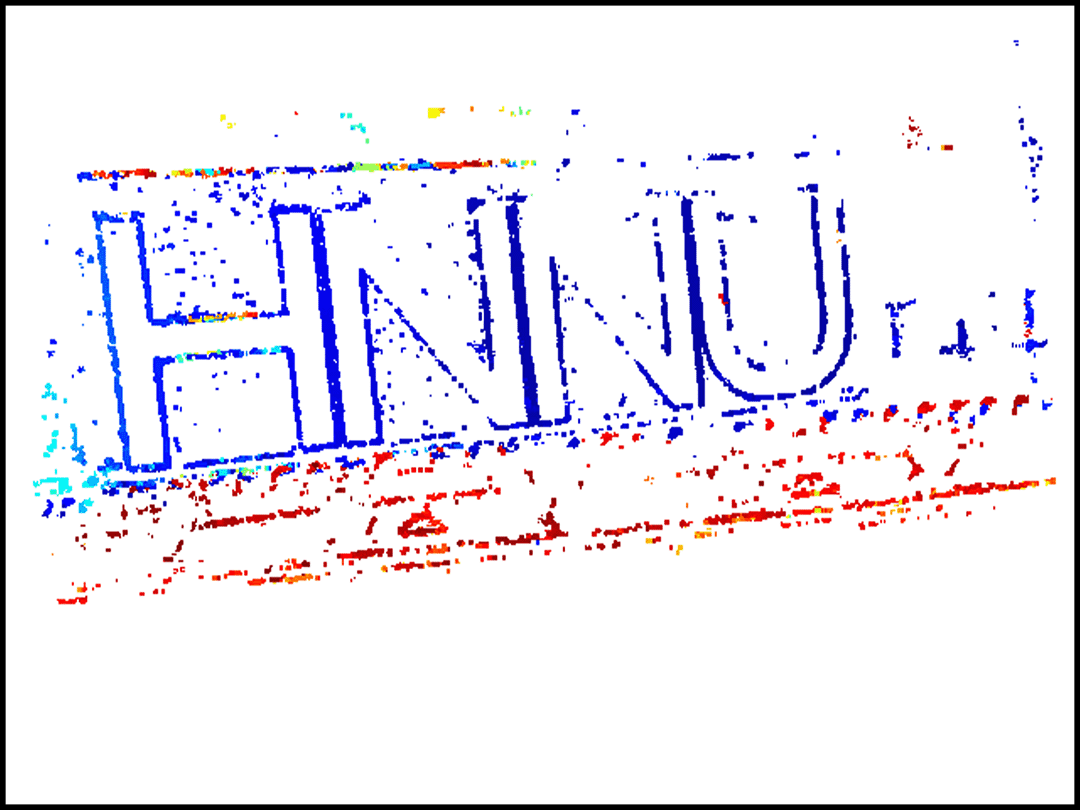}} &
	\gframe{\includegraphics[trim={4px 4px 4px 4px},clip,width=\linewidth]{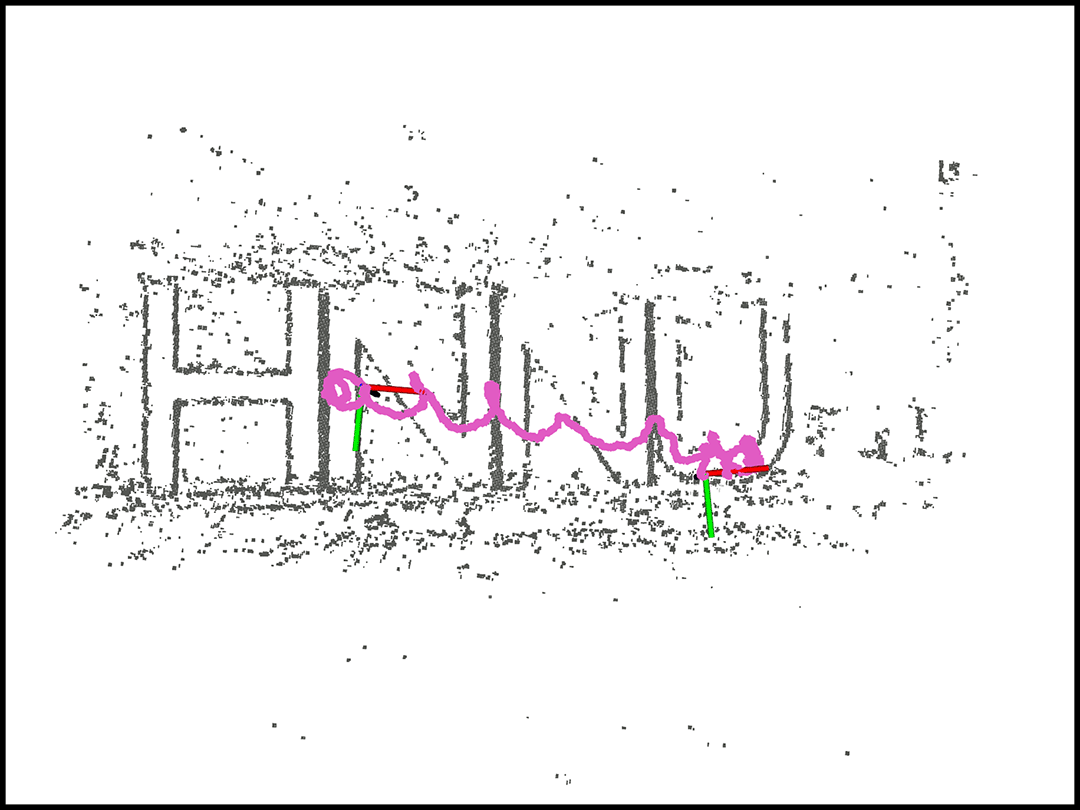}} \\

	\rotatebox{90}{\makecell[c]{\ \emph{HNU\_street}}} & 
    {\includegraphics[width=\linewidth]{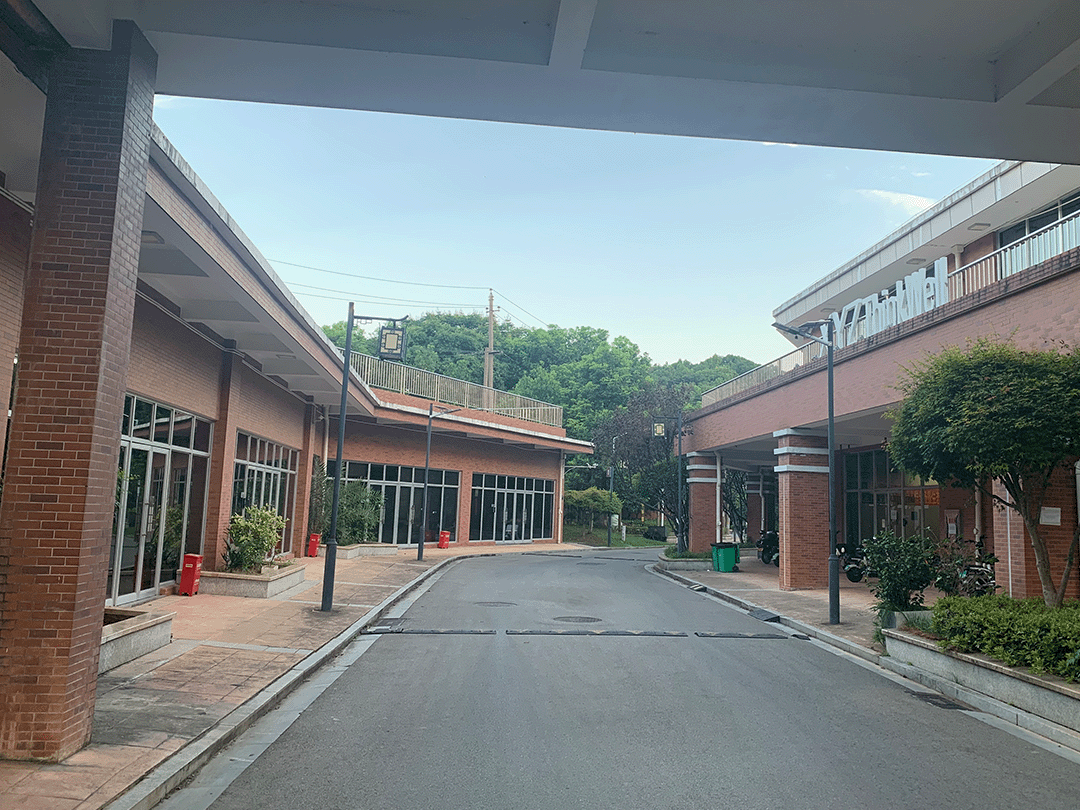}} & 
    \gframe{\includegraphics[trim={4px 4px 4px 4px},clip,width=\linewidth]{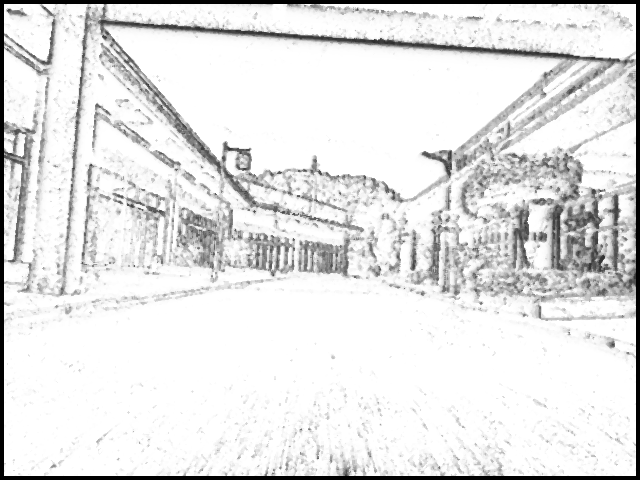}} &
    \gframe{\includegraphics[trim={4px 4px 4px 4px},clip,width=\linewidth]{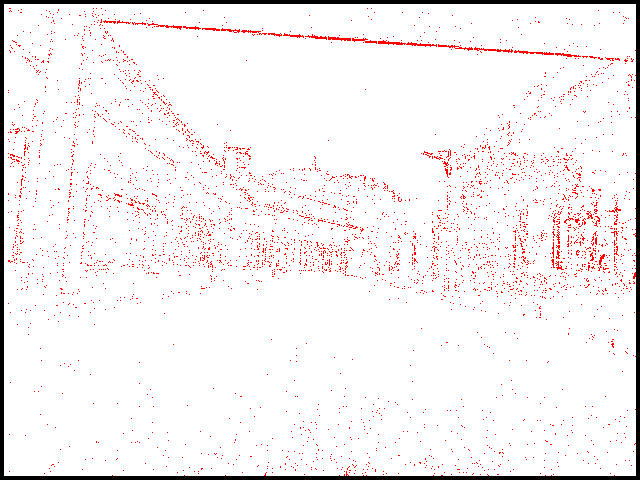}} &
    \gframe{\includegraphics[trim={4px 4px 4px 4px},clip,width=\linewidth]{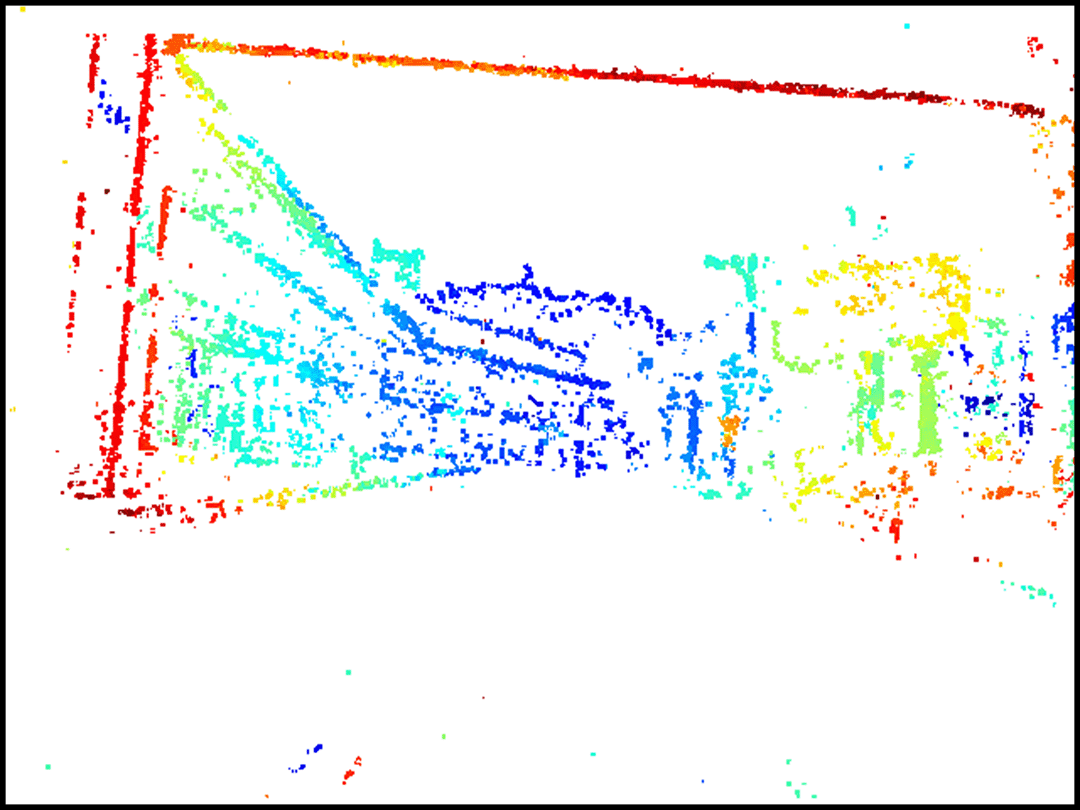}} &
	\gframe{\includegraphics[trim={4px 4px 4px 4px},clip,width=\linewidth]{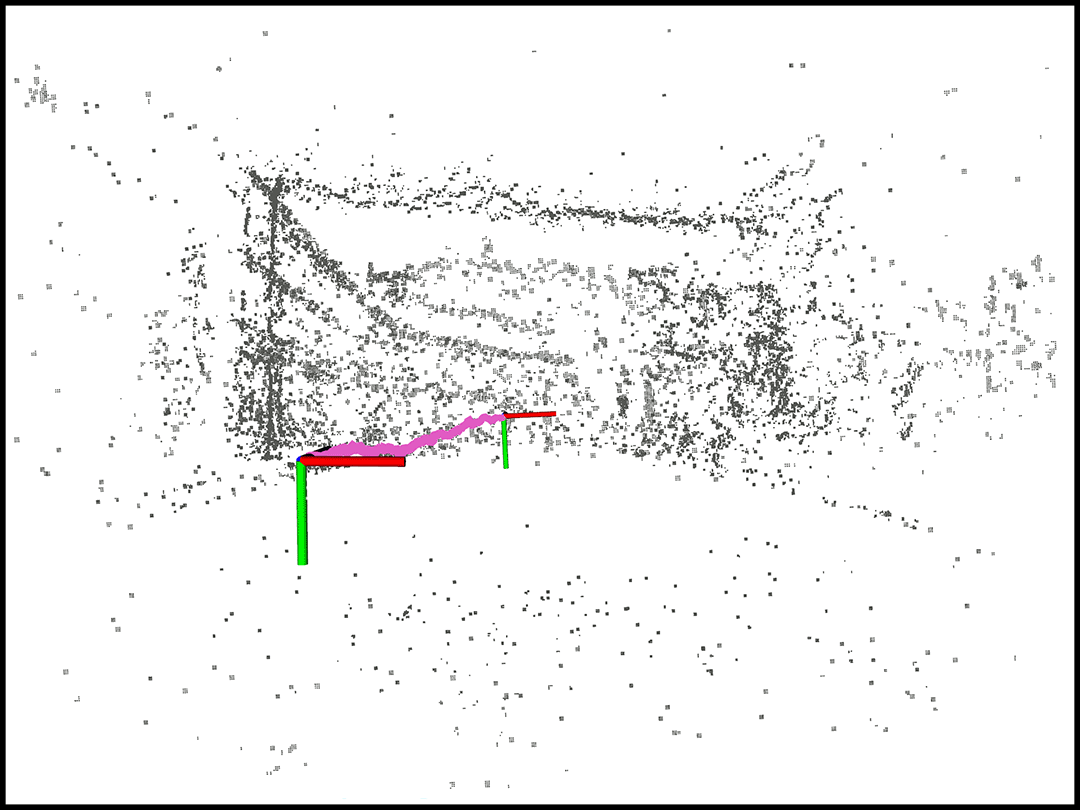}} \\

	\rotatebox{90}{\makecell[c]{\ \emph{HNU\_pavilion }}} & 
    {\includegraphics[width=\linewidth]{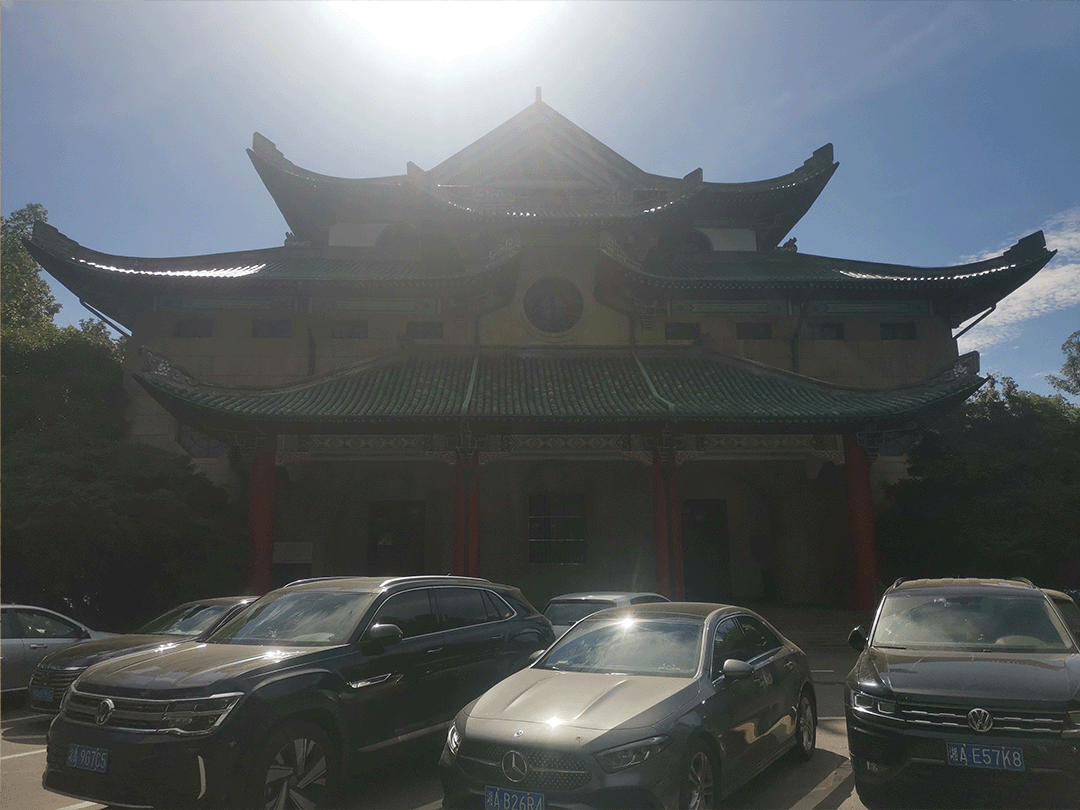}} & 
    \gframe{\includegraphics[trim={4px 4px 4px 4px},clip,width=\linewidth]{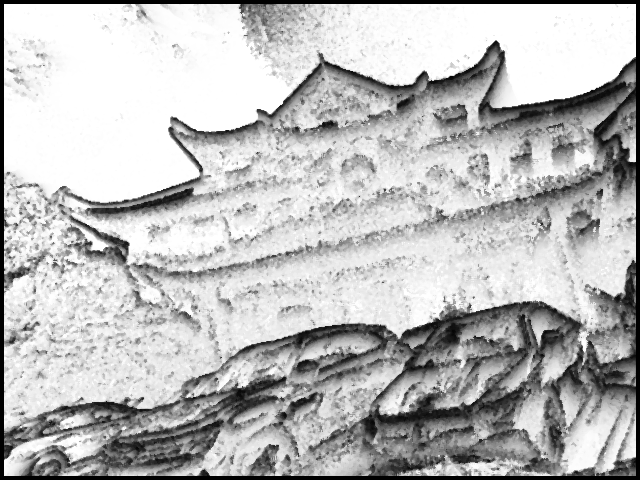}} &
    \gframe{\includegraphics[trim={4px 4px 4px 4px},clip,width=\linewidth]{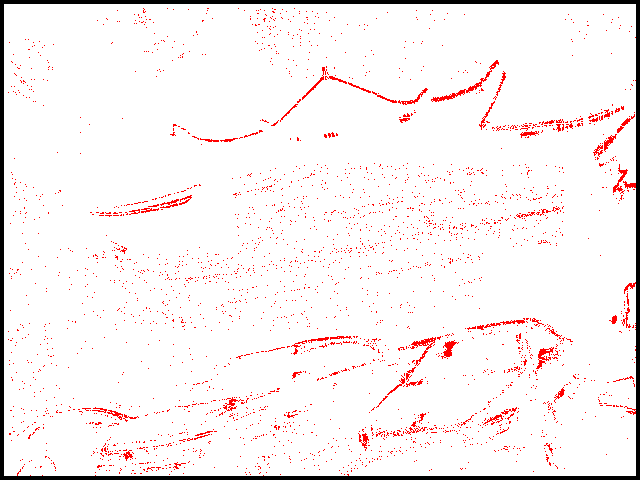}} &
    \gframe{\includegraphics[trim={4px 4px 4px 4px},clip,width=\linewidth]{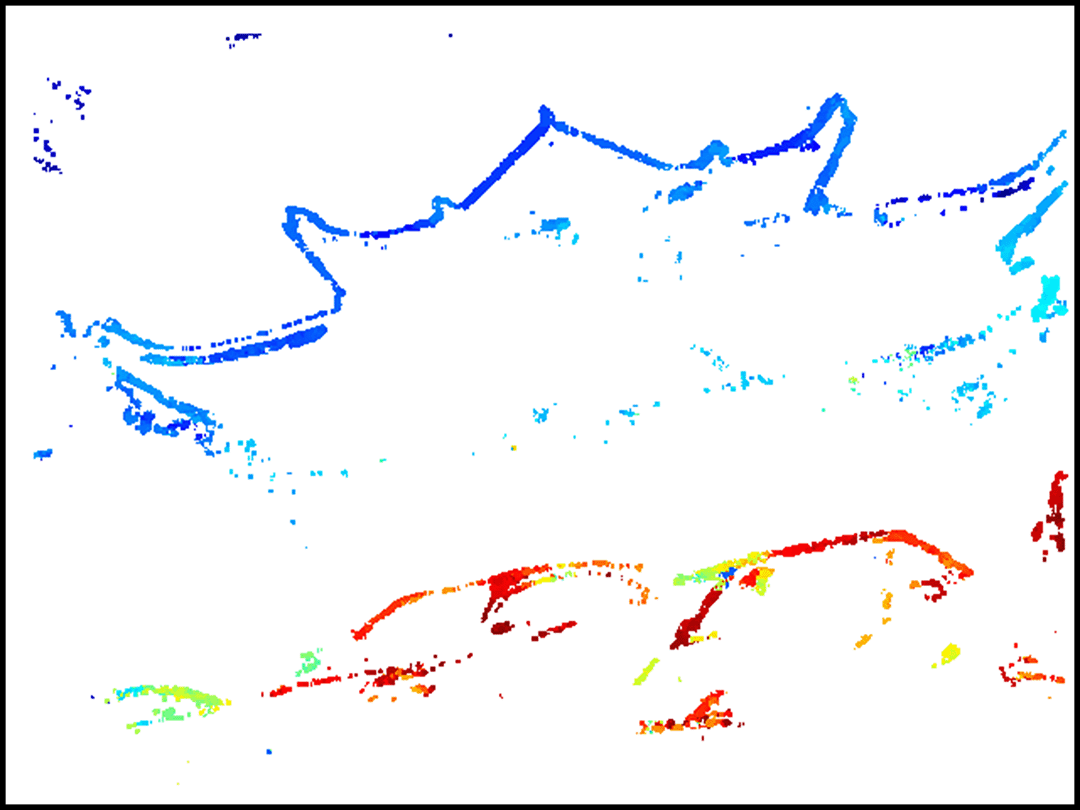}} &
	\gframe{\includegraphics[trim={4px 4px 4px 4px},clip,width=\linewidth]{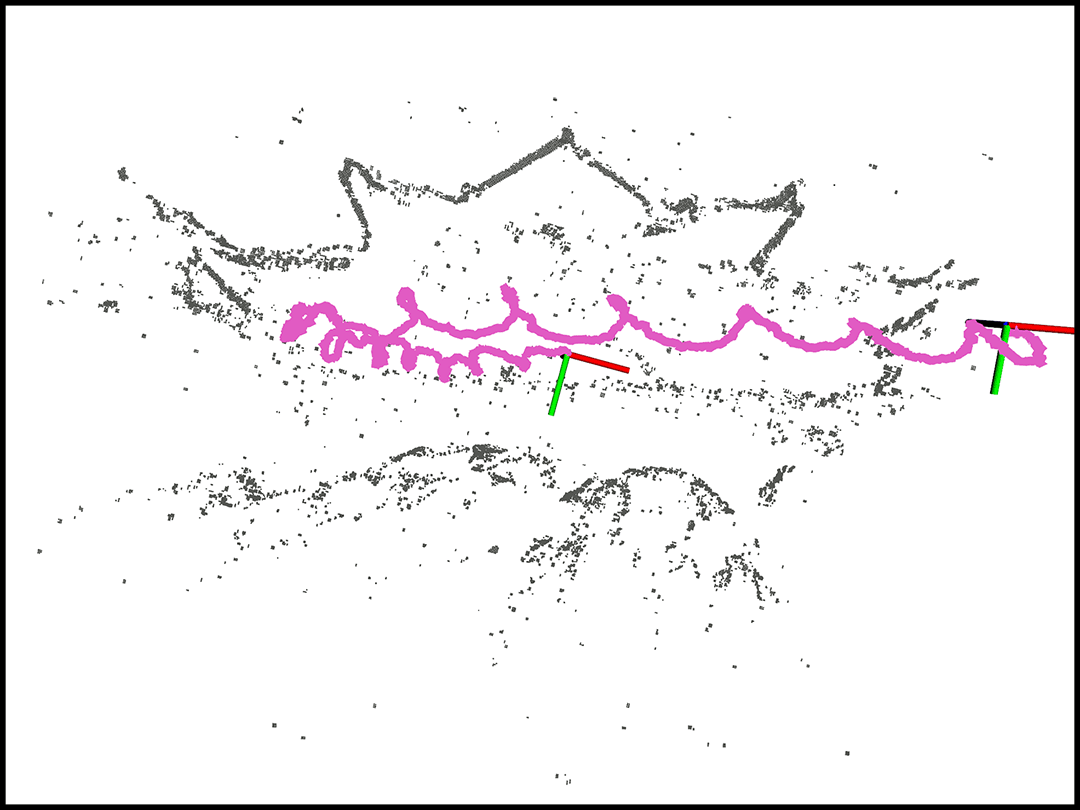}} \\

	\rotatebox{90}{\makecell[c]{\ \emph{HNU\_statue}}} & 
    {\includegraphics[width=\linewidth]{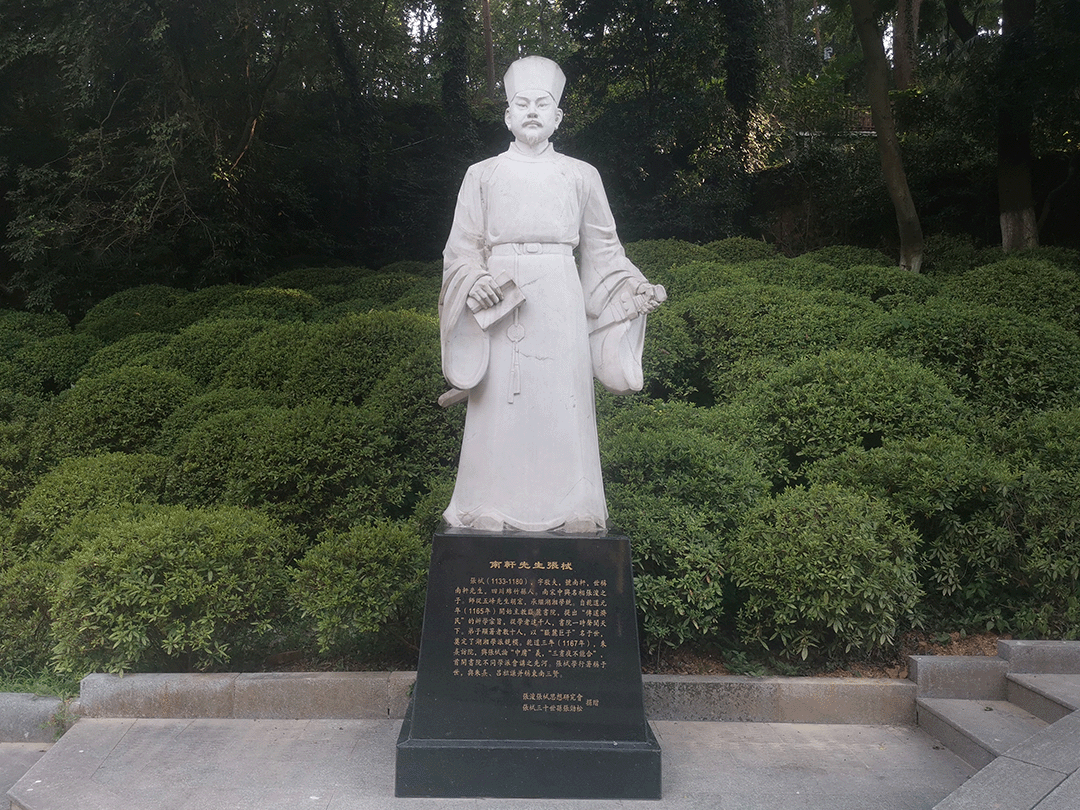}} & 
    \gframe{\includegraphics[trim={4px 4px 4px 4px},clip,width=\linewidth]{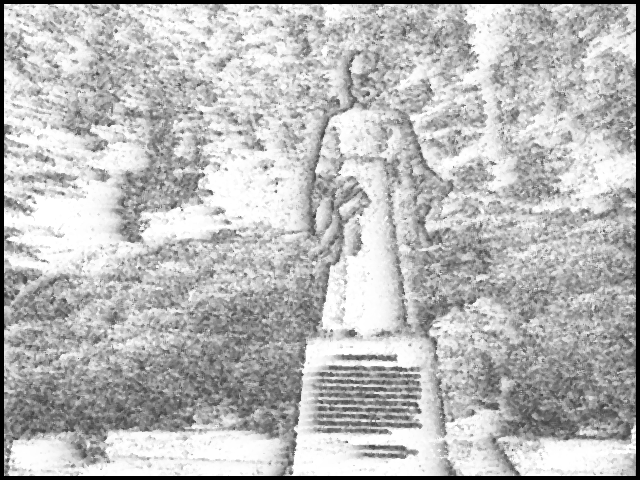}} &
    \gframe{\includegraphics[trim={4px 4px 4px 4px},clip,width=\linewidth]{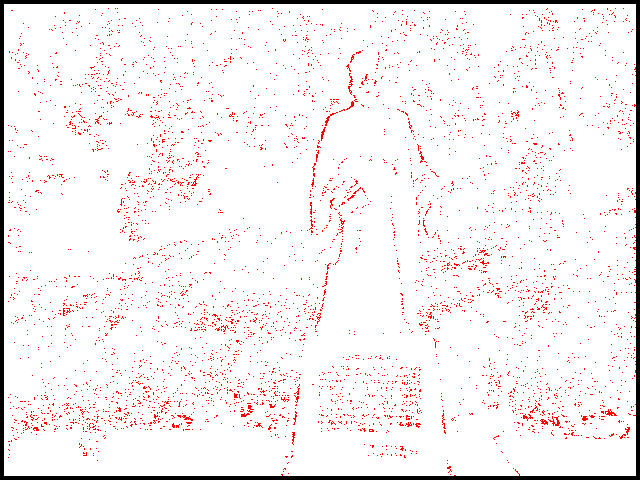}} &
    \gframe{\includegraphics[trim={4px 4px 4px 4px},clip,width=\linewidth]{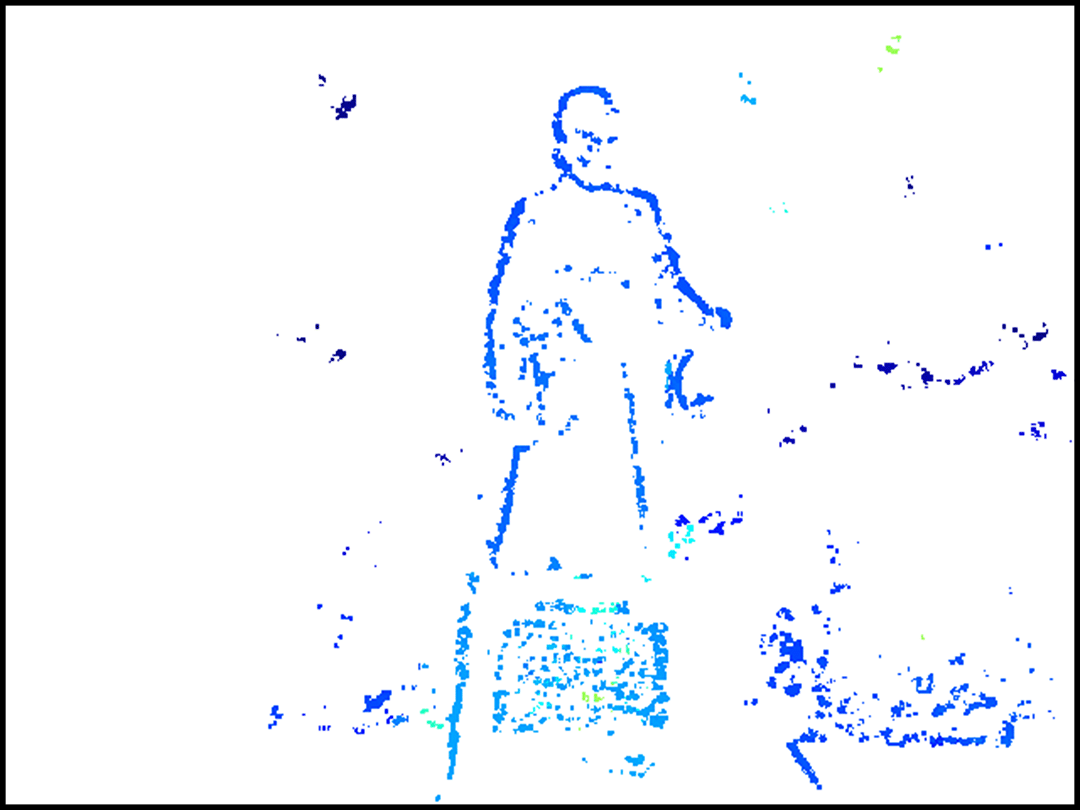}} &
	\gframe{\includegraphics[trim={4px 4px 4px 4px},clip,width=\linewidth]{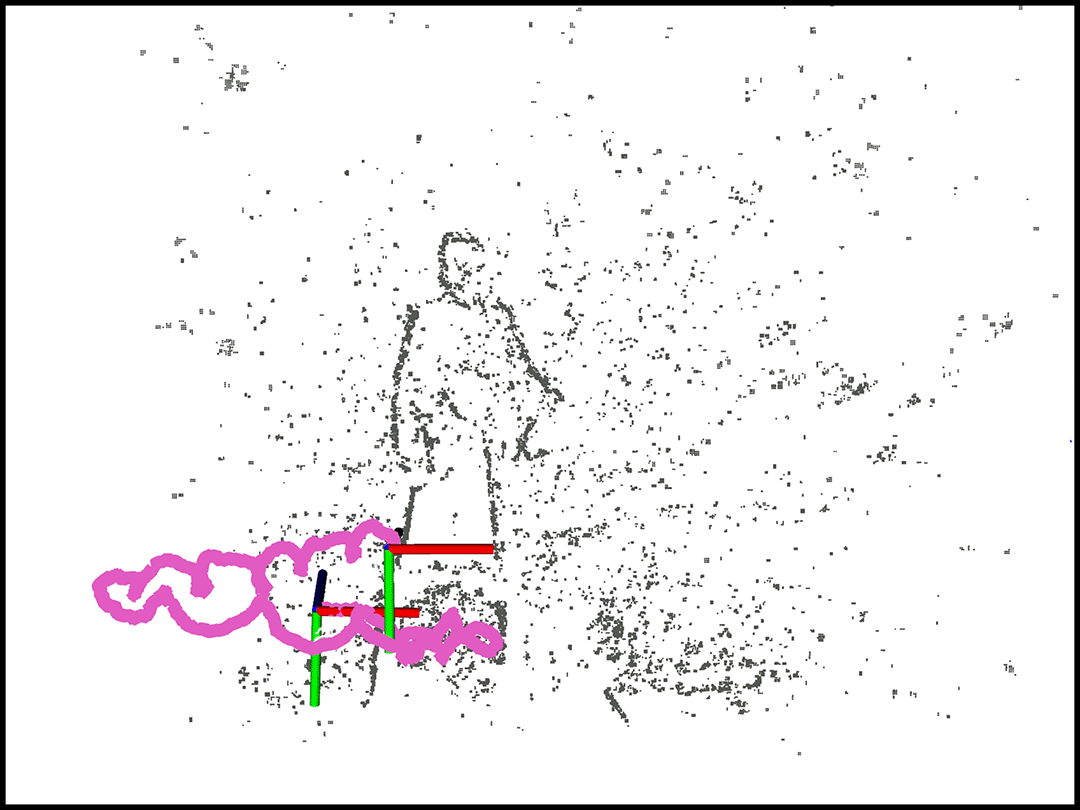}} \\
 
	\rotatebox{90}{\makecell[c]{\ \emph{HNU\_courtyard}}} & 
    {\includegraphics[width=\linewidth]{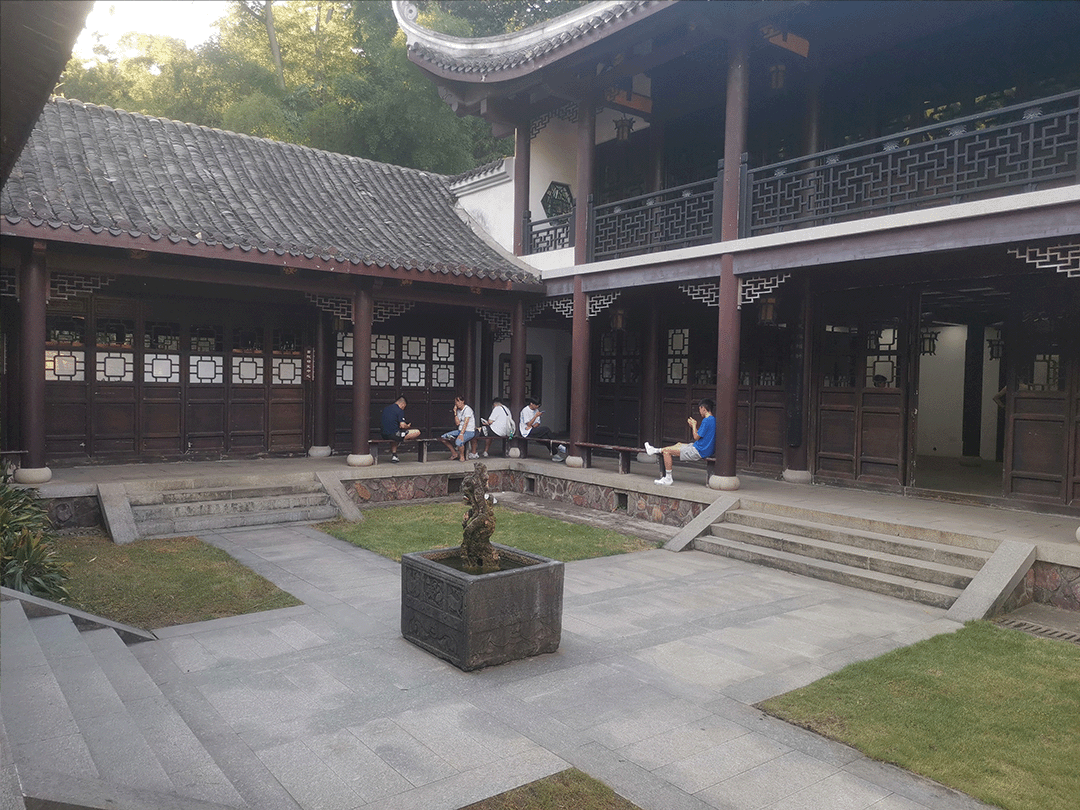}} & 
    \gframe{\includegraphics[trim={4px 4px 4px 4px},clip,width=\linewidth]{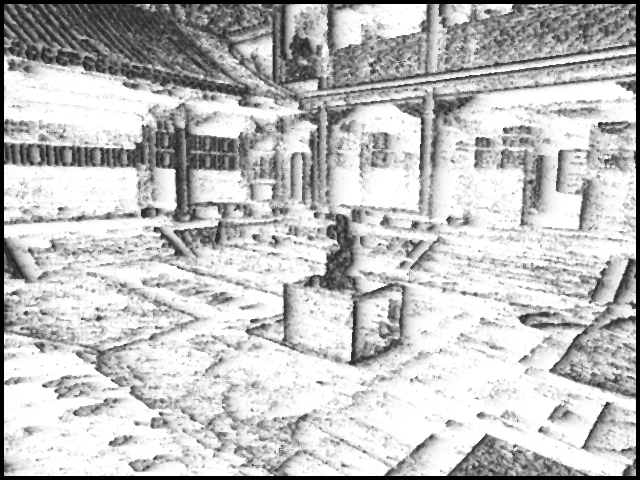}} &
    \gframe{\includegraphics[trim={4px 4px 4px 4px},clip,width=\linewidth]{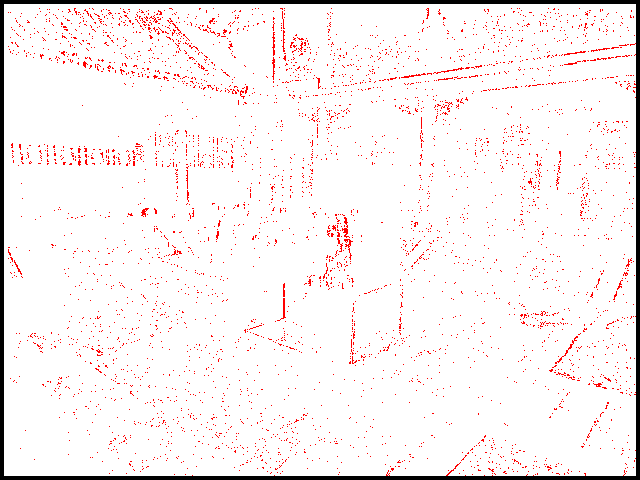}} &
    \gframe{\includegraphics[trim={4px 4px 4px 4px},clip,width=\linewidth]{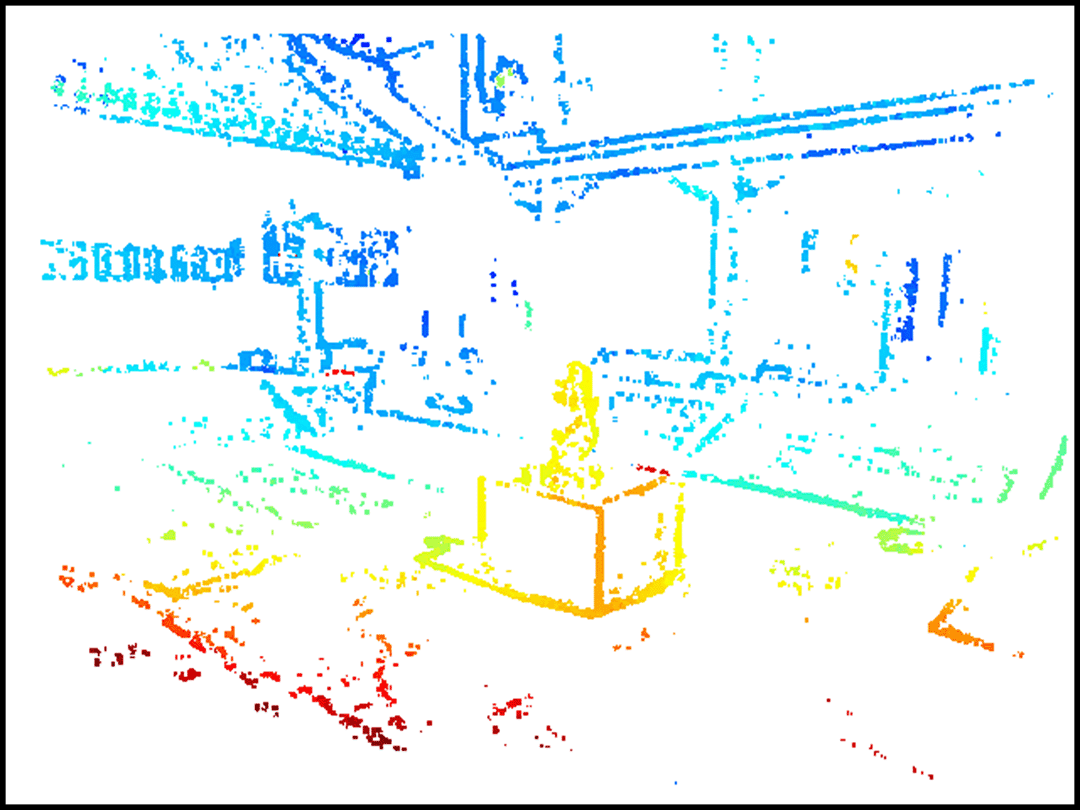}} &
	\gframe{\includegraphics[trim={4px 4px 4px 4px},clip,width=\linewidth]{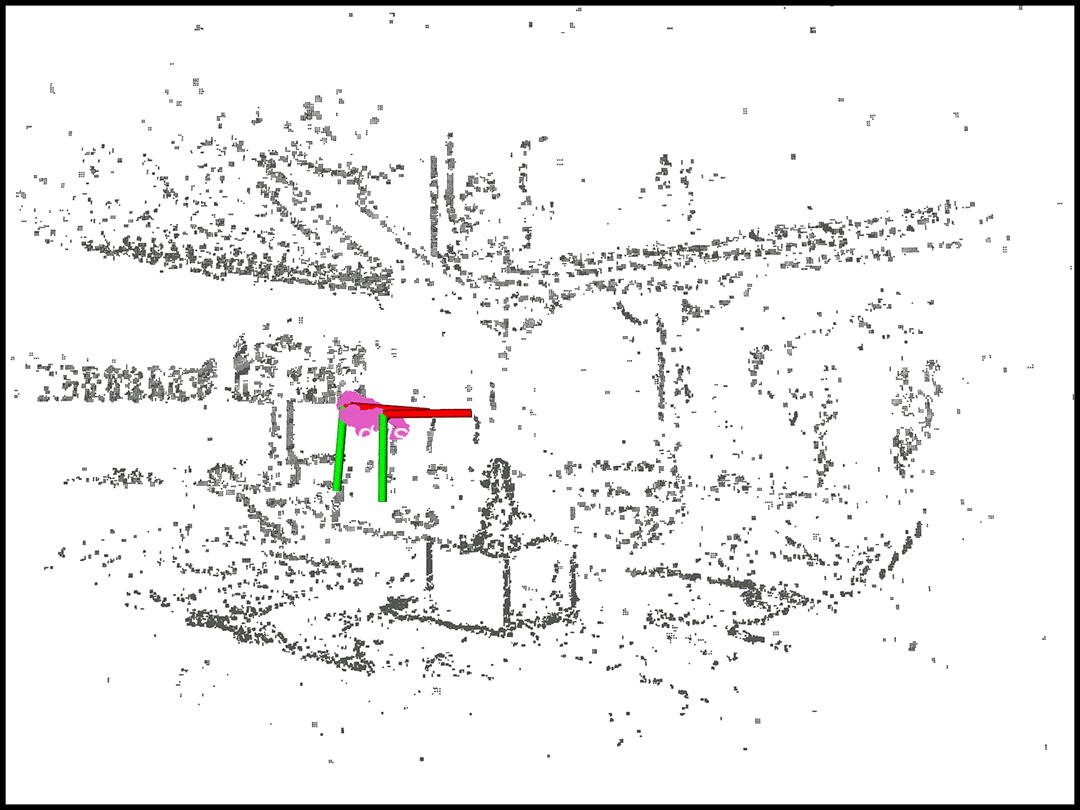}} \\

    \end{tabular}
    }

   \caption{\label{fig:outdoor mapping evaluvation}
   \emph{Mapping results of our system on the hnu\_mapping dataset}. 
   From left to right: scenes, time surfaces, sampled edge pixels, estimated inverse depth maps, and the resulting point clouds with the camera trajectory.}
\end{figure*}

We conduct several ablation studies to assess the benefits brought by the new functions and modules proposed in this work (Secs.~\ref{subsubsec:Ablation study Back-end} 
to \ref{subsubsec:Ablation study refinement}).
Additionally, we investigative the sensitivity with respect to the size parameter in the sliding window and with respect to running our pipeline with/without the IMU input (Sec.~\ref{subsubsec:sliding window size and necessity of using IMU}).
In all studies, we use the \emph{DSEC} dataset~\cite{Gehrig21ral} and compare the performance of our system (denoted as ``reference'' in Tabs.~\ref{tab:ablation_study_part1} and \ref{tab:ablation_study_part2}) with the following features disabled, respectively.

\subsubsection{Effect of the Back-end}
\label{subsubsec:Ablation study Back-end}
As said in Sec.~\ref{sec:backend}, the presence of back-end facilitates suppressing drift in IMU biases and predicting the linear velocity, thus leading to better performance in camera pose tracking.
To justify this, we compare the trajectory results obtained with and without the back-end.
In the configuration without the back-end (denoted by ``w/o back-end'' in Tab.~\ref{tab:ablation_study_part1}), the gyroscope pre-integration results are still used as an initialization for the spatio-temporal registration, but its bias is not updated.
As shown in Tab.~\ref{tab:ablation_study_part1}, ``reference'' exhibits notably fewer drift in the estimated trajectories than ``w/o back-end'', demonstrating the necessity of constantly updating the belief of the IMU biases and the camera's linear velocity.
With these updated parameters, we obtain better initial values for the spatio-temporal registration and, thus, more accurate motion estimation.
Consequently, the improved tracking results lead to more accurate mapping results (especially for the temporal-stereo part) and, in turn, facilitate the tracking module.

It is also worth explaining the difference between the results in Tab.~\ref{tab:ate_eval} (denoted by ``ICRA'24'') and those in Tab.~\ref{tab:ablation_study_part1}.
Note that the results obtained with the back-end disabled are less accurate than those reported in \cite{niu2024imu}.
This is because we not only use the gyroscope's pre-integration results in \cite{niu2024imu} but also predict the next position using a constant-velocity motion model.
In this ablation study, only the rotation pre-integration is performed, which indicates that the constant-velocity motion model can enhance the accuracy of trajectory estimation.

\subsubsection{Effect of OS-TS}
\label{subsubsec:Ablation study OS-TS}
As claimed in Sec.~\ref{subsec:camera pose tracking}, the proposed OS-TS ensures no offset in the location of edges while preserving valid gradient information in the neighboring area.
To justify this, we compare the trajectory results obtained with and without OS-TS.
In the configuration without OS-TS, the blurred TS (Fig.~\ref{fig:TS_discussion_TS_blur}) is used in the 3D-2D spatio-temporal registration.
As shown in Tab.~\ref{tab:ablation_study_part1}, the trajectory errors increase significantly when OS-TS is not employed (denoted by ``w/o OS-TS''). 
This is because the highest pixel values in the blurred TS do not accurately correspond to the positions of the edges in the current scenes.
As a result, even when the registration problem is solved successfully, the obtained pose estimation is consistently sub-optimal.
Our proposed OS-TS addresses this issue by filling in gradients on the original TS, ensuring the successful resolution of the registration problem and reducing the accumulated error caused by inaccurate pose estimations.

\subsubsection{Effect of the Nonlinear Refinement of Depth}
\label{subsubsec:Ablation study refinement}
In the original ESVO pipeline~\cite{zhou2021event}, the disparity estimated by block matching is further refined with a successive nonlinear optimization, aiming at a sub-pixel accuracy.
As shown in Tab.~\ref{tab:ablation_study_part1}, however, we observe that this refinement (denoted by ``w/ refinement'') sometimes offers only a marginal improvement in trajectory accuracy compared to the reference method while significantly increasing the runtime of the mapping operation.
To guarantee a stable real-time performance at VGA resolution, we choose to omit the nonlinear refinement.

\subsubsection{Sensitivity with respect to the Size of the Sliding Window and Necessity of using an IMU}
\label{subsubsec:sliding window size and necessity of using IMU}

To investigate how the performance changes when using different sizes of the sliding window and running the pipeline without any IMU measurements, we conduct another ablation study, as reported in Tab.~\ref{tab:ablation_study_part2}.
The ``reference'' group denotes the results obtained under the default configuration, namely choosing a window size of 5 and using the IMU measurements.
Compared to the ``reference'' group, the pose-estimation accuracy oscillates slightly up and down when using a window size of 8 or 12.
This indicates that the choice of using different window sizes does not have a significant effect on the results.  
Our empirical selected window size aims to strike a balance between efficiency and accuracy.
Second, it is clearly seen that the accuracy of the pose estimation drops notably when the IMU is not used, demonstrating the importance of introducing the IMU into the sensor suite.

\subsection{Computational Efficiency}
\label{subsec:implementation details and computational efficiency}

\begin{figure*}[t]
    \centering
    \begin{minipage}[m]{0.625\textwidth}
        \centering
        \begin{adjustbox}{max width=\linewidth}{
        \begin{tabular}{lllll}
\toprule
\textbf{Node} (\#\textbf{threads}) & \textbf{Function} & \textbf{ESVO}~\cite{zhou2021event} & \textbf{ICRA'24}~\cite{niu2024imu} & \textbf{Ours}\\
\midrule
\multirow{2}{*}{Pre-processing (1)} & Time surface & 27 ($\sim$70k) & \textbf{3.8} ($\sim$70k) & \textbf{3.8}  ($\sim$70k)\\
{~} & Adaptive accumulation  & - & 8 ($\sim$70k) & \textbf{2.1} ($\sim$70k) \\
\midrule
\multirow{1}{*}{Tracking (2)} & Spatio-temporal registration & 8 ($\sim$2k) & 7 ($\sim$2k) & \textbf{5} ($\sim$2k) \\
\midrule
\multirow{7}{*}{Mapping (4)} & Static stereo & 46 ($\sim$10k) & 33 ($\sim$3.6k) & \textbf{8} ($\sim$3.6k) \\

{~} & Depth optimization & 78 ($\sim$4.5k) & 13 ($\sim$1.9k) & - \\
{~} & Static-stereo fusion & 16.4 ($\sim$140k) & 5.4 ($\sim$15k) & \textbf{5.3} ($\sim$14k) \\
{~} & Temporal stereo & - & \textbf{4} ($\sim$0.4k) & \textbf{4} ($\sim$0.4k) \\
{~} & Temporal-stereo fusion & - & \textbf{0.7} ($\sim$2k) & \textbf{0.7} ($\sim$2k) \\
{~} & Regularization [optional] & 253 ($\sim$25k) & - & - \\
{~} & Others & \textbf{4} & 15.7 & 14.5 \\
\cdashline{2-5}\vspace{-3.5mm}\\
{~} & {Subtotal [w/ optional]} & 144.4 [397.4] & 71.8 & \textbf{32.5}  \\
\midrule
{Back end (1)} & Optimization & - & - & 5 \\
\bottomrule
\end{tabular}
}
        \end{adjustbox}
        \captionof{table}{\emph{Computational performance.} The numbers in parentheses indicate the approximate number of event points used by each function. [$\mathbf{Time}$: ms].}
        \label{tab:6}
    \end{minipage}
    \hspace{0.4cm}
    \begin{minipage}[m]{0.32\textwidth}
        \centering
        \includegraphics[width=5.6cm]{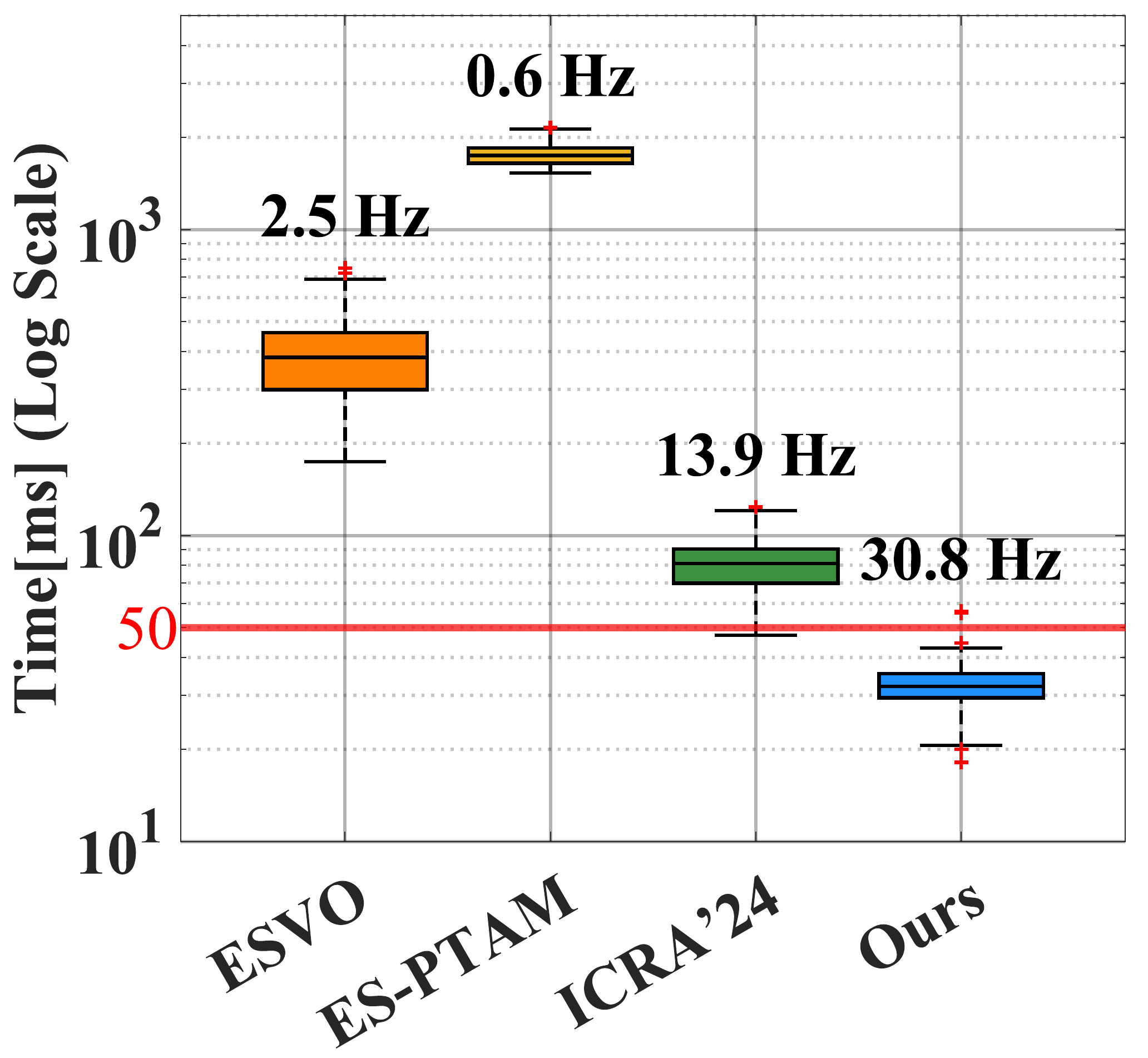}
        \captionof{figure}{\emph{Comparison of mapping runtime.} The mean frequency of running each method's mapping module is shown above the corresponding box.}
        \label{fig:mapping cost}
    \end{minipage}%
    
\end{figure*}

To assess the computational efficiency, we compare the runtime of ESVO~\cite{zhou2021event}, ICRA'24~\cite{niu2024imu}, and the proposed system using a desktop with an Intel Core i7-14700k CPU, as shown in Tab.~\ref{tab:6}.
The tests are performed on the \emph{DSEC} dataset.
All three systems are implemented using hyper-threading technology, and the number of threads occupied by each node is declared inside the parentheses following the node name.
The average runtime of each function is profiled, and the numbers in parenthesis (hyper-parameters) denote the amount of data processed by each function.
It should be noted that the runtime of our system is primarily influenced by the spatial resolution of the event cameras, proportionally to which the number of points sampled from AA is set.
All other hyper-parameters (\eg, the number of points used by each function) are set accordingly, aiming at best performance.
As a result, the event streaming rate has minimal influence on our system's runtime.

\begin{figure*}[t]
    \centering
    \subfigure[\label{fig:discussion_indoor_a}\emph{corridors\_dolly}\protect~\cite{gao2022vector}]{
        \includegraphics[width=0.192\linewidth]{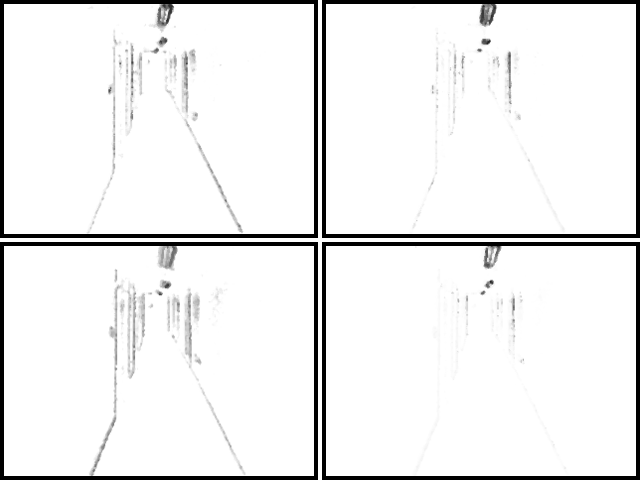}}%
    \subfigure[\label{fig:discussion_indoor_b}\emph{school\_scooter}\protect~\cite{gao2022vector}]{
        \includegraphics[width=0.192\linewidth]{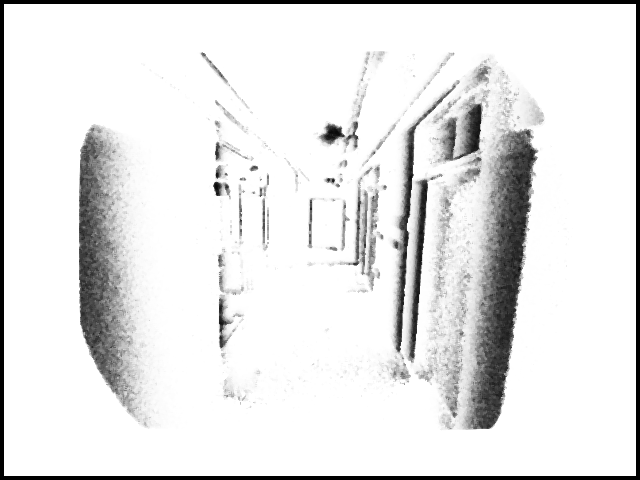} 
        \includegraphics[width=0.192\linewidth]{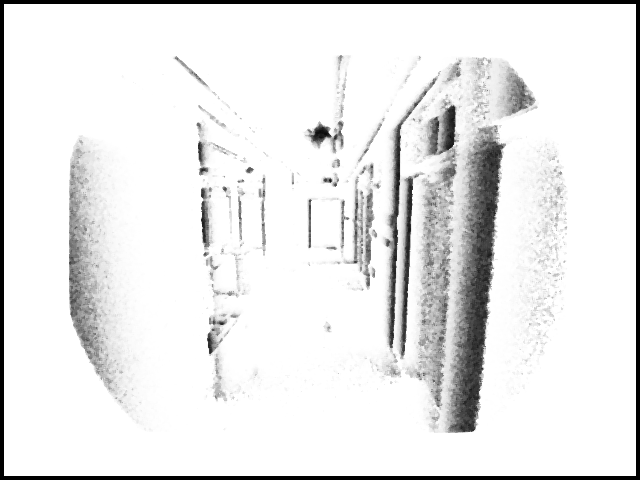}}%
    \subfigure[\label{fig:discussion_indoor_c}\emph{rpg\_bin}~\cite{Zhou18eccv}]{
        \includegraphics[width=0.192\linewidth]{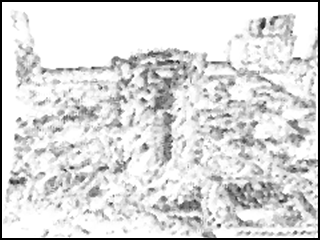}\hspace{2pt}
        \includegraphics[width=0.192\linewidth]{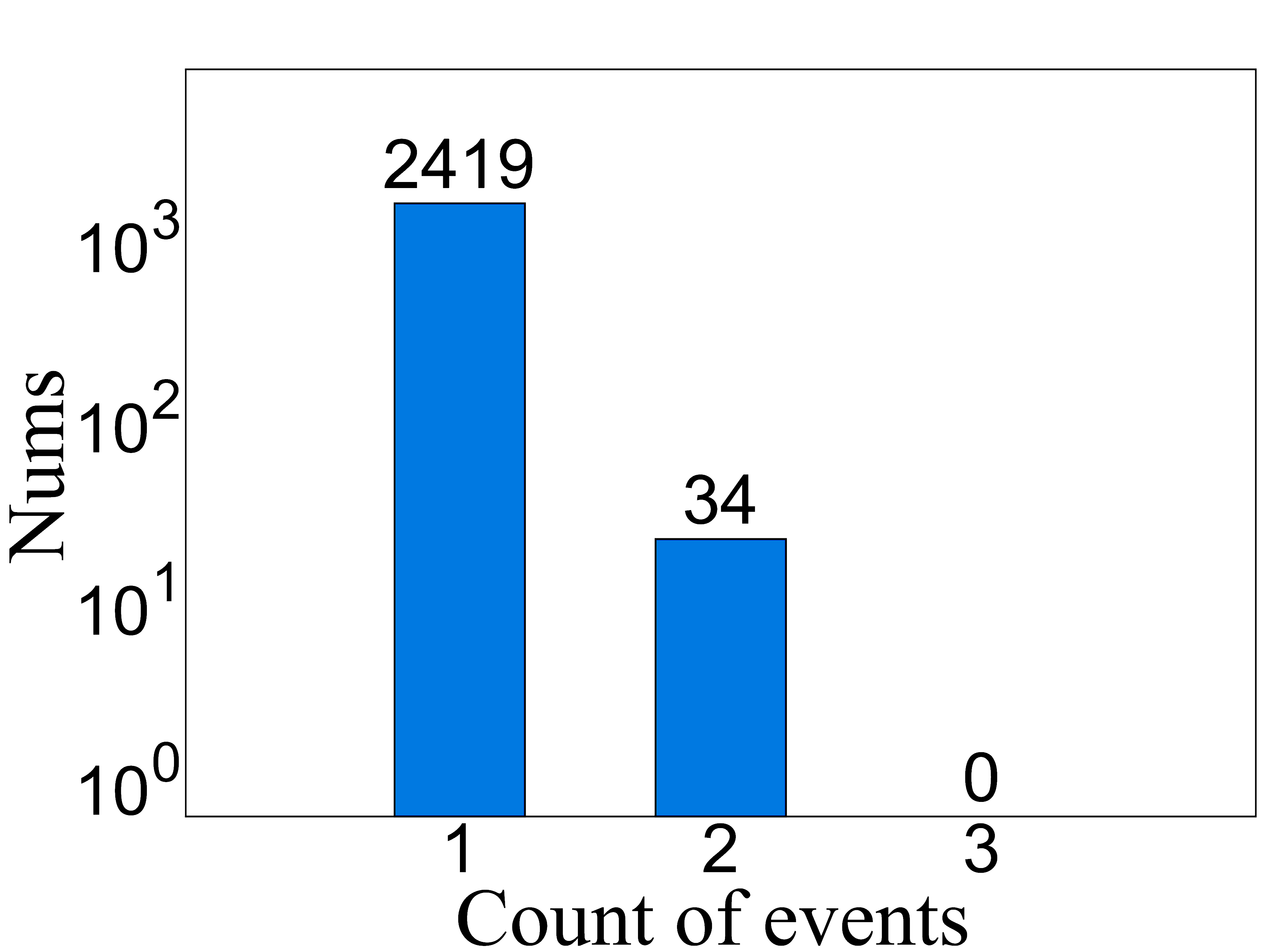}\vspace{1pt}}
    
    \caption{\emph{Degenerate examples and justifications.}
    (a) Consecutive TSs (from the left camera) selected from the \emph{VECtor}~\cite{gao2022vector} dataset,
    where the event edges are unstable (flickering).
    (b) TSs (from the left and right camera, respectively) from the \emph{VECtor}~\cite{gao2022vector} dataset, exhibiting prominent ``long ramps'' due to the rapid motion of edges.
    (c) Left: TS from the \emph{rpg}~\cite{Zhou18eccv} dataset.
    Right: the corresponding histogram made on the number of events occurred per pixel every 10 ms.
    }
    \label{fig:disucssion_indoor}
    \vspace{-0em}
\end{figure*}

    

We improve the rendering efficiency of TS and AA by code refactoring, resulting in TS generation being $8\times$ faster than in ESVO and AA generation being $4\times$ faster than in \cite{niu2024imu}.
The efficiency of the front-end is also slightly improved due to better initial values provided by the back-end. 

Most of the modifications are made in the mapping module because its performance is mostly relevant to the accuracy and efficiency of the entire system.
The improvement in terms of mapping efficiency is due to the following:
($i$) The number of points required by the proposed mapping module is significantly reduced compared to ESVO because the input points are no longer raw events but precise contour points extracted by the applied AA and contour-point sampling method;
($ii$) The proposed fast block matching speeds up the static-stereo operation;
($iii$) We get rid of the depth refinement because we find that the nonlinear optimization does not bring notable improvements in mapping accuracy when the accurate contour points are fed as input;
($iv$) The newly added temporal-stereo operation causes a runtime increment ($\approx$5 ms).
Nevertheless, it is worthwhile as it is the key to the improvement of the overall mapping quality.
Moreover, the optional regularization operation is no longer used because it is time-consuming and has little effect on the front-end's performance.
The ``others'' entry in Tab.~\ref{tab:6} includes general operations in the code that cannot be named as functions, such as data transfer and recycle.
The runtime of the mapping module demonstrates that our mapping module can stably operate in real time at 20~Hz on input stereo events of VGA resolution.
Last but not the least, our back-end runs in another independent thread, taking 5 ms for each update of the optimization variables in the sliding window.


Considering that the mapping operation typically consumes the biggest portion of the computing resources in an event-based VO pipeline, we compare the mapping module's runtime performance of four open-source direct methods.
To do this, we record the mapping module's runtime of every method on all VGA-resolution datasets and visualize their statistics using a box plot (Fig.~\ref{fig:mapping cost}).
It is clearly seen that only the median runtime of our method is below the $50$-ms bar, ensuring real-time performance on data of VGA resolution.
Additionally, we assess the runtime of a feature-based method (ESIO~\cite{chen2023esvio}) using the same hardware. 
Using the recommended configuration by the authors, the average cost of the most time-consuming back-end module in ESIO is 37.34 ms.
This indicates that the proposed direct method is on par with state-of-the-art feature-based methods in terms of computational efficiency; meanwhile, as shown in Secs.~\ref{subsec:system evaluation} and \ref{subsec: Outdoor Evaluation}, our method outperforms others in accuracy.


\subsection{Limitations}
\label{subsec:limitations}
We observe that our system does not perform well on a small number of sequences, and we attribute this to the inherent limitations of direct methods and the specific characteristics of some datasets.
The first category of degenerate cases is typically witnessed in small-scale and narrow spaces, \eg, a textureless corridor (shown in Fig.~\ref{fig:discussion_indoor_a} and Fig.~\ref{fig:discussion_indoor_b}).
In this case, edge patterns either appear parallel to the moving direction of the event camera, resulting in unstable visual observations, or are so close to the event camera that the spatio-temporal profile of event data becomes significantly less distinctive, leading to data association failure in mapping and tracking.
This situation exposes more the limitations of the sensor than the limitations of the proposed system.
A simple cure is to introduce an additional complementary sensor, \eg, a standard camera, which easily enriches visual information. 
The second type of degeneration occurs when using low-resolution (\eg, 240 $\times$ 180 pixel) event cameras in cluttered environments of repeated textures.
In this case, the streaming rate of events becomes lower than that using a high-resolution event camera under the same dynamics, as justified in Fig.~\ref{fig:discussion_indoor_c}, thus hindering the determination of contour points via the proposed AA.
Furthermore, the presence of widely cluttered, repetitive textures in the scene hinders block matching, thereby worsening depth estimation. 

Besides, the employed image-like representations have limitations on further (fully) exploiting the high-speed property of event cameras.
We notice the existence of several alternative representations of event data \cite{li2021graph}. 
For example, an early work interprets an event stream as 3D point clouds \cite{Benosman14tnnls} in the spatio-temporal domain, from which normal flow can be computed by fitting local planes. 
The computation is asynchronous and efficient (closed-form solution), and the resulting normal flow can be used for state estimation \cite{lu2023event}, leading to ultra-frame-rate update rate. 
However, this map-free design \cite{lu2023event} is an open-loop scheme, outperforming in speed by sacrificing accuracy. 
As for other representations, such as the spatio-temporal graph of events, they are typically discussed in the task of object recognition \cite{li2021graph} but not in the context of event-based SLAM.
\section{Conclusion}
\label{sec:conclusion}

We have presented an event-based stereo visual-inertial odometry system on top of our previous work ESVO~\cite{zhou2021event}.
It is a direct method that solves the tracking and mapping problems in parallel by leveraging the spatio-temporal coherence in the stereo event data.
The goal has been to alleviate ESVO's high computational complexity in mapping and address its degeneracy in camera pose tracking.
To this end, we have additionally introduced an IMU and added several modules to the new pipeline.
The newly proposed contour-point sampling method reduces the number of points required by the mapping module, significantly reducing runtime.
The temporal-stereo operation acts as a complement to the static stereo, enhancing the mapping quality in terms of structure completeness and local smoothness.
The tracking method is still formulated as a 3D-2D registration problem, and its degeneracy is addressed via the usage of IMU pre-integration as motion priors and special considerations for better convergence.
To maintain an accurate pre-integration, a compact back-end has been proposed to suppress drift in the estimated IMU biases.
Extensive experimental evaluation on five publicly available datasets of different resolutions and our own has demonstrated that our system represents the state of the art and fulfills all the claimed contributions. 
To the best of our knowledge, the system is the first published work that achieves real-time performance using a standard CPU on event cameras of VGA pixel resolution.
The software and datasets used for evaluation are open-sourced.
\section*{Acknowledgment}
\label{sec:ack}
The author would like to thank Mr. Peiyu Chen, Dr. Zhe Liu, and Mr. Suman Ghosh for providing the raw results of \cite{Ghosh24eccvw, chen2023esvio, liu2023esvio}, respectively, as baselines used in the evaluation.
We thank Mr. Jinghang Li and Mr. Kaizhen Sun for the help in the data collection.
We also thank Dr. Yi Yu for proofreading.
This work was supported by the National Key Research and Development Project of China under Grant 2023YFB4706600.


\bibliographystyle{IEEEtran} 

\bibliography{myBib}


\begin{IEEEbiography}[{\includegraphics[width=1in,height=1.25in,clip,keepaspectratio]{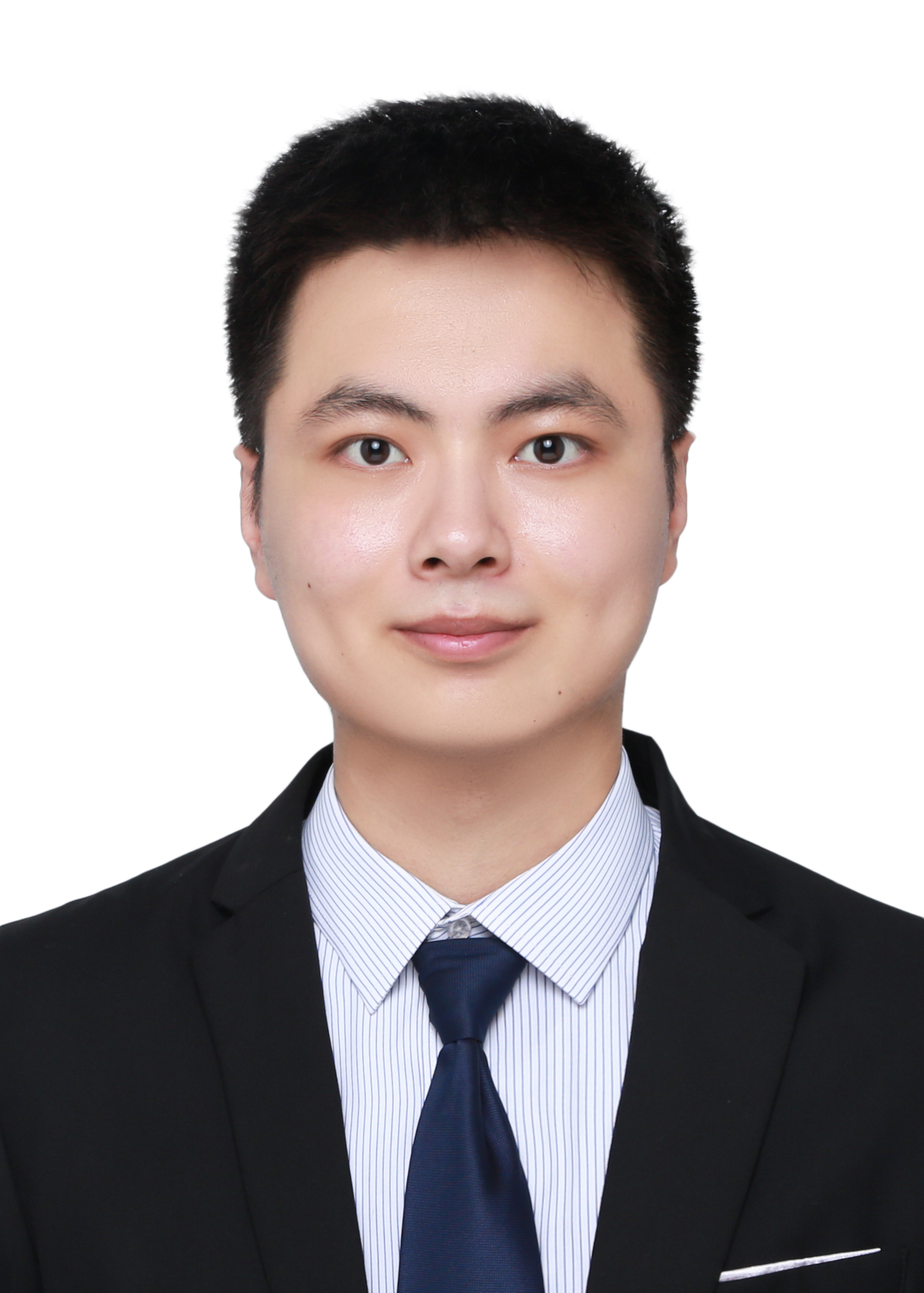}}]{Junkai Niu}
received the Master's degree in Aerospace Science and Technology from the University of Electronic Science and Technology of China, Chengdu, China in 2023.
He is currently working toward the Ph.D. degree in the Neuromorphic Automation and Intelligence Lab (NAIL) with School of Robotics at Hunan University.
His research interests primarily focus on event-based visual odometry/simultaneous localization and mapping.
\end{IEEEbiography}

\begin{IEEEbiography}[{\includegraphics[width=1in,height=1.25in,clip,keepaspectratio]{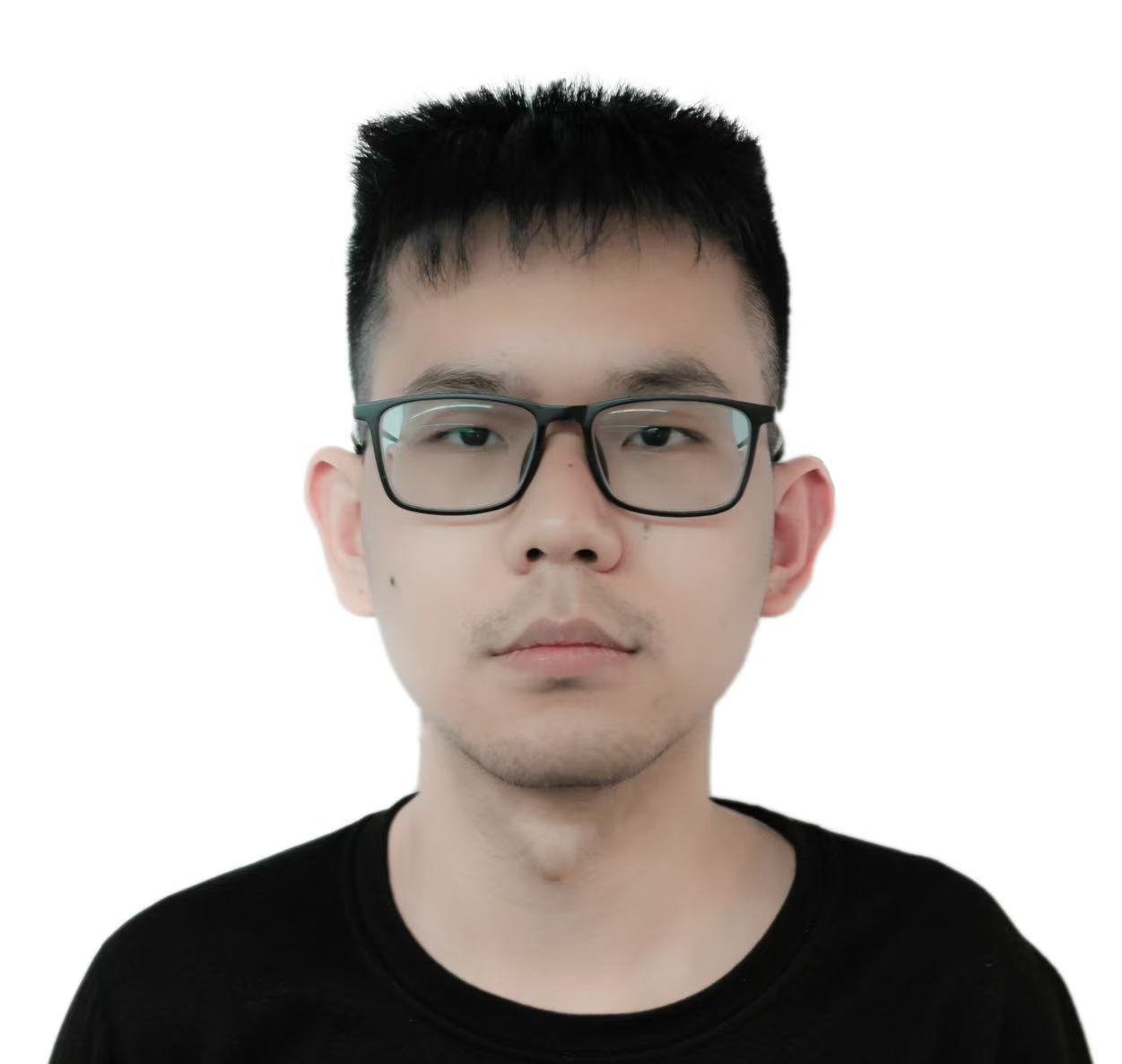}}]{Sheng Zhong}
received the B.Eng. degree in Robotics Engineering from Zhejiang University, Hangzhou, China, in 2023. 
He is currently working toward the Ph.D. degree in the Neuromorphic Automation and Intelligence Lab (NAIL) with School of Robotics at Hunan University.
His research interests primarily focus on event-based visual odometry/simultaneous localization and mapping.
\end{IEEEbiography}

\begin{IEEEbiography}[{\includegraphics[width=1in,height=1.25in,clip,keepaspectratio]{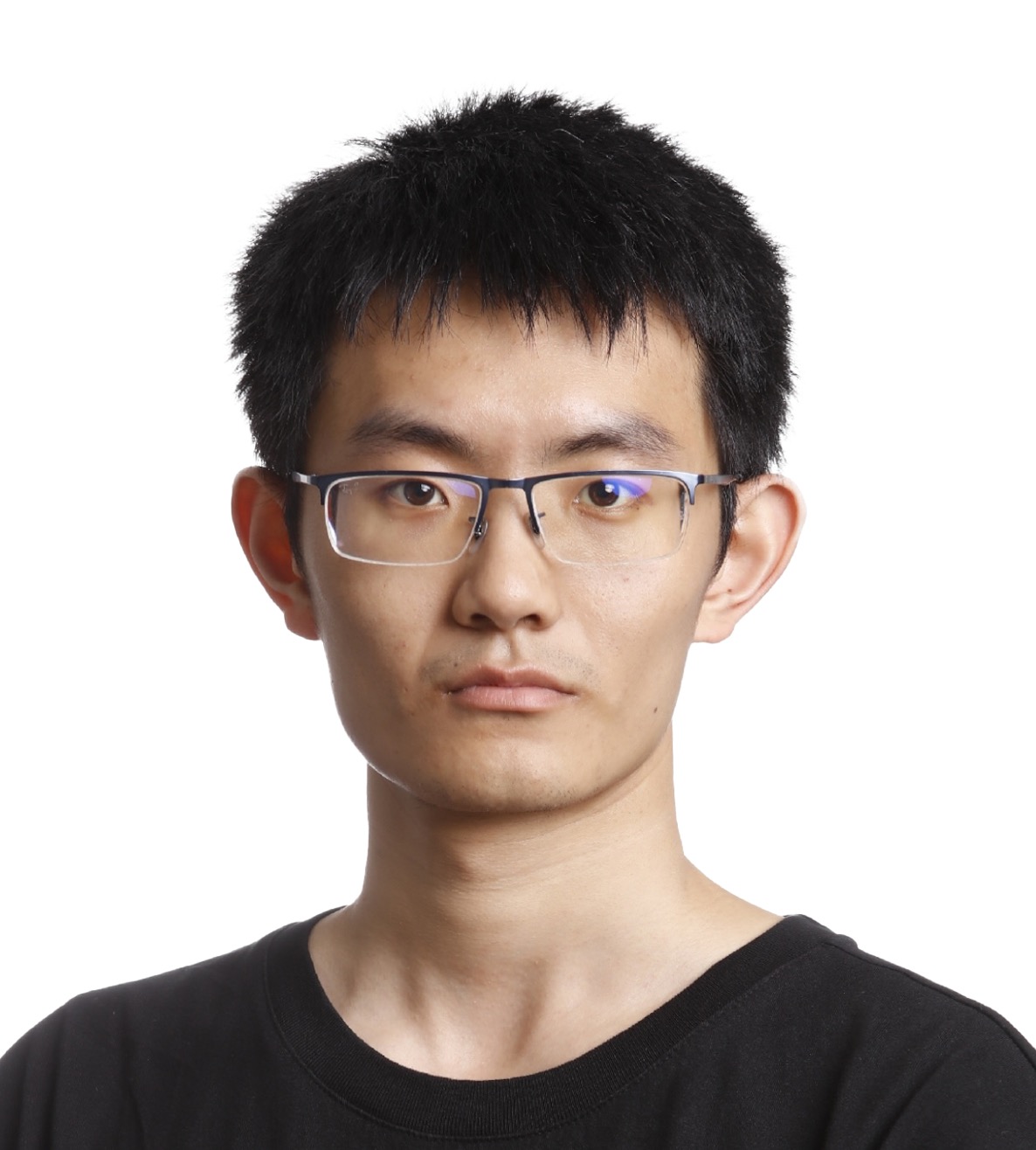}}]{Xiuyuan Lu}
received the B.Eng. degree in computer science from the Hong Kong University of Science and Technology (HKUST), Hong Kong, in 2020. 
He is currently a PhD candidate in electronic and computer engineering at the Hong Kong University of Science and Technology, Hong Kong.
His research interests include event-based vision and visual odometry/simultaneous localization and mapping. 
\end{IEEEbiography}

\begin{IEEEbiography}[{\includegraphics[width=1in,height=1.25in,clip,keepaspectratio]{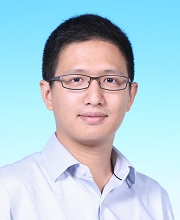}}]{Shaojie Shen} (Member, IEEE) received his B.Eng. degree in Electronic Engineering from the Hong Kong University of Science and Technology (HKUST) in 2009. 
He received his M.S. in Robotics and Ph.D. in Electrical and Systems Engineering in 2011 and 2014, respectively, from the University of Pennsylvania. 
He is currently an Associate Professor in the Department of Electronic and Computer Engineering and the founding director of the HKUST-DJI Joint Innovation Laboratory (HDJI Lab) at HKUST. 
His research interests are in the areas of robotics and unmanned aerial vehicles, with focus on state estimation, sensor fusion, localization and mapping, and autonomous navigation in complex environments. 
He is currently serving as a senior editor for ICRA 2024-2026, and as an associate editor for IJRR 2023-2024. 
He and his research team received the 2023 IEEE T-RO King-Sun Fu Memorial
Best Paper Award and the 2023 IEEE RA-L Best Paper Award, and also achieved the Honorable Mention status for the IEEE T-RO Best Paper Award in 2018 and 2020, and won the Best Student Paper Award in IROS 2018. 
Additionally, Prof. Shen was recognized as the AI 2000 Most Influential Scholar Award Honorable Mention in 2020 and consecutively from 2021-2024.
\end{IEEEbiography}

\begin{IEEEbiography}[{\includegraphics[width=1in,height=1.25in,clip,keepaspectratio]{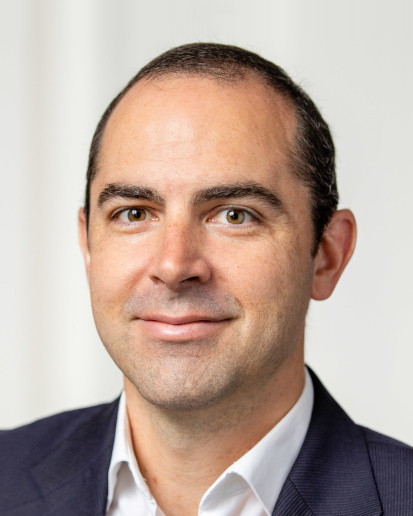}}]{Guillermo Gallego} (SM'19) is Full Professor at Technische Universit\"at Berlin, Berlin, Germany, in the Dept. of Electrical Engineering and Computer Science, and at the Einstein Center Digital Future, Berlin, Germany.
He is also a Principal Investigator at the Science of Intelligence Excellence Cluster and the Robotics Institute Germany, Berlin, Germany.
He received the PhD degree in Electrical and Computer Engineering from the Georgia Institute of Technology, USA, in 2011, supported by a Fulbright Scholarship.
From 2011 to 2014 he was a Marie Curie researcher with Universidad Politecnica de Madrid, Spain, and from 2014 to 2019 he was a postdoctoral researcher at the Robotics and Perception Group, University of Zurich and ETH Zurich, Switzerland.
He serves as Associate Editor for IEEE Transactions on Pattern Analysis and Machine Intelligence, and for IEEE Robotics and Automation Letters, the International Journal of Robotics Research, and as Guest Editor of the IEEE T-RO Special Collection on Event-based Vision for Robotics.
His research interests include robotics, computer vision, signal processing, optimization and geometry. 
\end{IEEEbiography}

\begin{IEEEbiography}[{\includegraphics[width=1in,height=1.25in,clip,keepaspectratio]{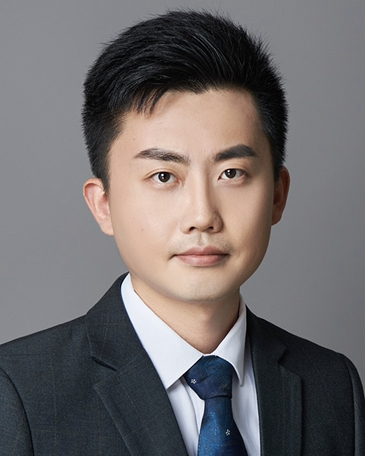}}]{Yi Zhou} (Member, IEEE) is a Professor at Hunan University, where he directed the Neuromorphic Automation and Intelligence Lab (NAIL).
He obtained his Ph.D. degree in engineering and computer science from the Australian National University, Canberra, Australia in 2018.
He was a visiting scholar at ETH Zurich (2017-2018) and was awarded the NCCR Fellowship Award by the Swiss National Science Foundation for his research on neuromorphic event-based 3D vision. 
From 2019 to 2021, he was a postdoc research fellow at the HKUST\&DJI Innovation Joint Lab, where he proposed the world's first open-source event-based stereo visual odometry (ESVO) system.
His research interests include visual odometry / simultaneous localization and mapping, geometry problems in computer vision, and dynamic vision sensors.
\end{IEEEbiography}

\end{document}